\documentclass[journal]{IEEEtran}
\usepackage{cite}
\ifCLASSINFOpdf
\else
\fi
\usepackage{url}
\makeatletter
\g@addto@macro{\UrlBreaks}{\UrlOrds}
\makeatother

\usepackage{float}
\usepackage{enumerate}
\usepackage{diagbox}
\usepackage{subcaption,multirow,color,amsmath}
\usepackage{array} % DO NOT CHANGE THIS
\usepackage{tikz,balance}
\usepackage{ulem}
\usepackage{enumitem,xcolor,booktabs,colortbl}
\usepackage[switch]{lineno}
\usetikzlibrary{positioning}
\begin{document}

  \title{Reinforcement Learning with Dual-Observation for General Video Game Playing}

\author{Chengpeng~Hu, Ziqi~Wang, Tianye~Shu, Hao~Tong, 
Julian~Togelius,~\IEEEmembership{Senior Member,~IEEE,} 
Xin~Yao,~\IEEEmembership{Fellow,~IEEE,}
Jialin~Liu,~\IEEEmembership{Senior Member,~IEEE}%<-this % stops a space
\thanks{C. Hu and Z. Wang are with the Research Institute of Trustworthy Autonomous System (RITAS), Southern University of Science and Technology, Shenzhen 518055, China. 
C. Hu, Z. Wang, T. Shu, H. Tong, X. Yao and J. Liu are with the Guangdong Provincial Key Laboratory of Brain-inspired Intelligent Computation, Department of Computer Science and Engineering, Southern University of Science and Technology, Shenzhen 518055, China.}% stops a space
\thanks{H. Tong and X. Yao are also with the University of Birmingham, Birmingham B15 2TT, United Kingdom.}% <-this % stops a space
\thanks{J. Togelius is with the New York University, New York 11201, USA.}% <-this % stops a space
%\thanks{This work was supported by the Research Institute of Trustworthy Autonomous Systems (RITAS), the Guangdong Provincial Key Laboratory (Grant No. 2020B121201001), the Program for Guangdong Introducing Innovative and Enterpreneurial Teams (Grant No. 2017ZT07X386), the Shenzhen Science and Technology Program (Grant No. KQTD2016112514355531) and the National Natural Science Foundation of China (Grant No. 61906083).}
\thanks{Corresponding author: Jialin Liu (liujl@sustech.edu.cn).}
\thanks{This work has been accepted by the IEEE Transactions on Games on March 21, 2022.}
}

\markboth{Journal of \LaTeX\ Class Files,~Vol.~14, No.~8, August~2015}%
{Shell \MakeLowercase{\textit{et al.}}: Bare Demo of IEEEtran.cls for IEEE Journals}

\maketitle

\begin{abstract}

Reinforcement learning algorithms have performed well in playing challenging board and video games. More and more studies focus on improving the generalisation ability of reinforcement learning algorithms. The General Video Game AI Learning Competition aims to develop agents capable of learning to play different game levels that were unseen during training. This paper summarises the five years' General Video Game AI Learning Competition editions. At each edition, three new games were designed. The training and test levels were designed separately in the first three editions. Since 2020, three test levels of each game were generated by perturbing or combining two training levels. Then, we present a novel reinforcement learning technique with dual-observation for general video game playing, assuming that it is more likely to observe similar local information in different levels rather than global information. Instead of directly inputting a single, raw pixel-based screenshot of the current game screen, our proposed general technique takes the encoded, transformed global and local observations of the game screen as two simultaneous inputs, aiming at learning local information for playing new levels. Our proposed technique is implemented with three state-of-the-art reinforcement learning algorithms and tested on the game set of the 2020 General Video Game AI Learning Competition. Ablation studies show the outstanding performance of using encoded, transformed global and local observations as input.

\end{abstract}

\IEEEpeerreviewmaketitle
\section{Introduction}

\IEEEPARstart{G}{ames} have always been popular benchmarks for testing artificial intelligence methods. In recent years, video games have been widely used due to multiple favourable characteristics, such as continuous dynamics, large action/state space, partial observability, and speed, which make them useful proxies of real-world problems. Although reinforcement learning (RL) has shown considerable success in learning to play the game of Go~\cite{Silver2016Mastering,silver2017mastering}, and various video games~\cite{Schrittwieser2019MasteringAG,kaiser2019model,kempka2016vizdoom}, the trained models were only used to play the games and levels on which they have been trained. Therefore, researchers have more recently also addressed the challenge of making RL \emph{generalise}~\cite{torrado2018drlgvgai,packer2018assessing,Oh2020Discovering}. 

The word \emph{generalise} in the context of general video game playing (GVGP) has several meanings: An agent can perform well on unseen levels after training on given levels of an identical game; or an agent trained on some games can work well on new games. Many studies on GVGP focus on the former aspect. Braylan \emph{et al.}~\cite{braylan2015reuse} used transfer learning to reuse the neural modules of agents in new agents for playing new games. 
Similarly, Tutum et al.~\cite{tutum2021generalization} evolved a context module to recognise temporal variations in games and autonomous driving simulations.
Techniques widely used in deep learning, such as regularisation, data augmentation and batch normalisation, have also been used to enhance the performance of RL algorithms or to increase their generalisation performance~\cite{kostrikov2020image,cobbe2019quantifying,laskin2020reinforcement}.
Justesen \emph{et al.}~\cite{justesen2018illuminating} enlarged the training set with procedural content generation methods to improve the generality of RL agents. It was shown that for some games, with a relatively large amount of generated levels and training steps, the agent can work well on a variety of levels within the same distribution~\cite{justesen2018illuminating}. Partly inspired by this, the Procgen benchmark~\cite{cobbe2020leveraging} includes 16 game-like environments with procedurally generated levels to evaluate the generalisation of RL agents.

The General Video Game AI (GVGAI) Learning Competition~\cite{gvgaibook2019,perez2019general}, first organised in 2017, requires agents to learn to play games without game simulators and achieve good performance on levels that are unseen during training. These unseen levels follow the original game rules but have a different layout and even new elements. The submitted agents are ranked according to their average win rate and the score they achieved while playing the unseen levels.

Despite the facilitation of using the GVGAI learning platform, only a few papers describing research using the platform have been published~\cite{kunanusont2017general,Apeldoorn2017AnAL,Dockhorn2018,torrado2018drlgvgai,justesen2018illuminating,ye2020rotation} and no significantly well-performed
agents were received in the 2017, 2018 or 2019 editions of the competition~\cite{perez2019general}. After observing the behaviours of these entries, we assume that their poor performance can probably be blamed on (i) the significant differences between the training and test level maps (including the map size), even when the game rules remain unchanged, and (ii) the direct use of game screenshots as inputs.

Motivated by verifying the above guess and developing a better GVGAI learning agent, we investigate the following questions in this paper:
(i) Can the specification of game states be reduced for better generality? 
(ii) Can generating more game states as training data improve the performance of an RL agent when playing unseen levels?

In this paper, we seek to answer the above research questions in the context of RL for GVGP. The main contributions of our work are as follows. (i) We propose a novel RL technique that simultaneously takes a transformed global observation and a transformed local observation (referred to as \emph{dual-observation} from now on) as input. Moreover, we use a novel \emph{tile-vector encoding} method for rapidly encoding observations and handling new tiles that may appear in unseen levels. (ii) The avatar is randomly placed at the beginning of each game during training as a simple technique to increase variety in training data.

The aforementioned dual-observation, tile-vector encoding and training with random initial positions are applied to three state-of-the-art RL algorithms from stable-baselines3~\cite{stable-baselines3}. Ablation studies on the 2020 General Video Game AI Competition games with dense, periodic and sparse rewards show that dual-observation and tile-vector encoding significantly improves the tested algorithms' performance in playing most of the unseen levels\footnote{Code available on GitHub: \url{https://github.com/SUSTechGameAI/DORL}}. Moreover, the results of five years' GVGAI Learning Competition editions, particularly the 2019 -- 2021 editions, are reported and summarised in this paper. The methodology of designing the 2020 and 2021 competition games are also described.

In the remainder of this paper, Section~\ref{sec:gvgai} presents the GVGAI Learning Competition and the competition results of all its five editions. The details of our proposed RL technique with dual-observation and tile-vector encoding, and corresponding experimental studies are described in Section~\ref{sec:arcane}. Section~\ref{sec:xp} compares the proposed RL technique with dual-observation implemented with three state-of-the-art RL algorithms on GVGAI Learning Competition games and discusses them. Finally, Section~\ref{sec:conclusion} concludes the paper.

\section{GVGAI Learning Competition}\label{sec:gvgai}

\subsection{GVGAI Learning Platform}\label{sec:framework} 
The GVGAI competition~\cite{perez2019general,gvgaibook2019} has offered an easy-to-use, open-sourced, common platform for researchers and practitioners to test and compare AI methods since 2014. During the last few years, the framework has been expanded into several tracks to meet the demand of different research directions~\cite{perez2019general,gvgaibook2019}. The GVGAI learning platform~\cite{torrado2018drlgvgai} particularly focuses on GVGP using RL methods. The GVGAI learning platform not only contains more than one hundred arcade-style single-player games and dozens of two-player games, but also offers the potential of adding an unlimited number of games and levels thanks to the video game definition language (VGDL)~\cite{schaul2013video,schaul2014extensible}. This functionality facilitates the enlargement of training and test data sets by allowing for games or levels to be designed either manually or with procedural content generation methods~\cite{togelius2011search,shaker2016procedural,summerville2018procedural,risi2020increasing,liu2020deep}.

\subsection{Competition Tasks and Rules}\label{sec:rules} 
Three different games, each with five levels, are provided in the learning competition. For each game, two levels are released as training levels, the other three remain unknown to participants for testing. At every game tick, an agent receives a screenshot of the current game screen, a list of legal actions, its game score as well as the game termination state (\texttt{PLAYER\_WINS} or \texttt{PLAYER\_LOSES} if the game is terminated, otherwise \texttt{NO\_WINNER}), and is expected to return an action to play within a certain time. The agent is expected to be able to play levels that were unseen during training. Readers are referred to \cite{perez2019general,gvgaibook2019} and the competition website\footnote{\label{cog2020}\url{http://aingames.cn/gvgai/ppsncog2020}} for more detailed competition task and rules.

The rules of the GVGAI Learning Competition basically follow the ones of the GVGAI Single-player Planning Competition~\cite{perez2014gvgpc}. In the training phase, agent entries can make the best use of the two given training levels of each game. In the testing phase, all submitted entries are executed using an identical Python script. Each entry is evaluated by its win rate and average score over 10 independent runs on each test level of each game, where each game has a maximum length of $2,000$ game ticks unless a different maximum length is explicitly defined in the game rules. For each level of each game, the entries are ranked first by their average scores, and then their win rates in the decreasing order.
For entries that receive the same number of wins and the same average score, the game length is considered to break the tie. Finally,  points are assigned to entries according to their rankings per level per game (25 for $1^{st}$, 18 for $2^{nd}$, 15 for $3^{rd}$, 12 for $4^{th}$, 10 for $5^{th}$, 0 for the others). The final scores of the individual entries are accumulated across the game levels. The one with the most final points wins the first prize in the competition. 

\subsection{Five Years' Experience of the GVGAI Learning Competition}
The GVGAI Learning Competition has been organised every year since 2017. Disappointingly, no particularly high performing agent has been received between 2017 and 2019~\cite{perez2019general,gvgaibook2019}. First, in the past editions of competition, some of the test level maps were of a different size to the training maps, resulting in different sizes of screenshots and failure of some implementations of RL algorithms~\cite{perez2019general}. Second, the training and testing level maps were designed separately and dissimilar. It's not surprising that an RL agent plays randomly when it encounters a game state that is very different from the one used in training. The former reason poses technical challenges, while the latter makes it necessary to reflect the design of general learning agents. Therefore, the methodology for designing competition games and levels was changed from 2020.

\subsection{Design of Competition Game Sets in 2020 and 2021}
The games used in the 2020 and 2021 competition editions have been carefully designed with the same screen size for all levels of a given game and different goals from each other.

\subsubsection{Game Sets}\label{sec:gameset}
The variety in reward frequency is a challenge to the generality of learning algorithms. The goal of the games varied from resource-collection to maze navigation to survival. This followed the tradition of designing GVGAI competition games since 2017, where designers have aimed for diversity.
Meanwhile, each game has a unique time limit corresponding to the task difficulty.

\paragraph{Game Set in 2020}
The 2020 competition edition\textsuperscript{\ref{cog2020}} provided three games, namely \emph{GoldDigger}, \emph{TreasureKeeper} and \emph{WaterPuzzle}, with dense, periodic and sparse rewards, respectively.
\begin{itemize}
    \item \emph{GoldDigger} is a resource-collection game. The avatar, thus the game character controlled by a player/agent, is expected to avoid monsters and collect all the jewels in a level to win the game. When a collision of the avatar with any monster occurs, it suddenly loses the game. It is also considered as a failure if a maximum of $2,000$ game ticks is passed. Once a jewel is collected, a score will be rewarded to the avatar. The avatar can also kill monsters to gain a score. Hence, \emph{GoldDigger} is a game with dense rewards because of the large number of jewels and monsters in levels. 
\item \emph{TreasureKeeper} is a Sokoban-like game aiming at keeping the avatar and treasure chests away from monsters. The game is designed with a periodic reward given every $100$ game ticks. Either a collision of monster and avatar or a monster and treasure chest will terminate the game with failure. If the avatar and all treasure chests survive for more than $600$ game ticks, it wins.
\item \emph{WaterPuzzle} is a maze game with sparse rewards. There are only three sprites in each level: the avatar, an immobile key and an immobile door. The agent is rewarded for collecting the key and for touching the door after collecting the key. The agent wins the game only if it collects the key and then touches the door in $1,500$ game ticks.
\end{itemize}

\begin{figure}[htbp]  % GoldDigger
\centering
    \begin{subfigure}[t]{.49\linewidth}
		\centering
		\includegraphics[width=.9\columnwidth]{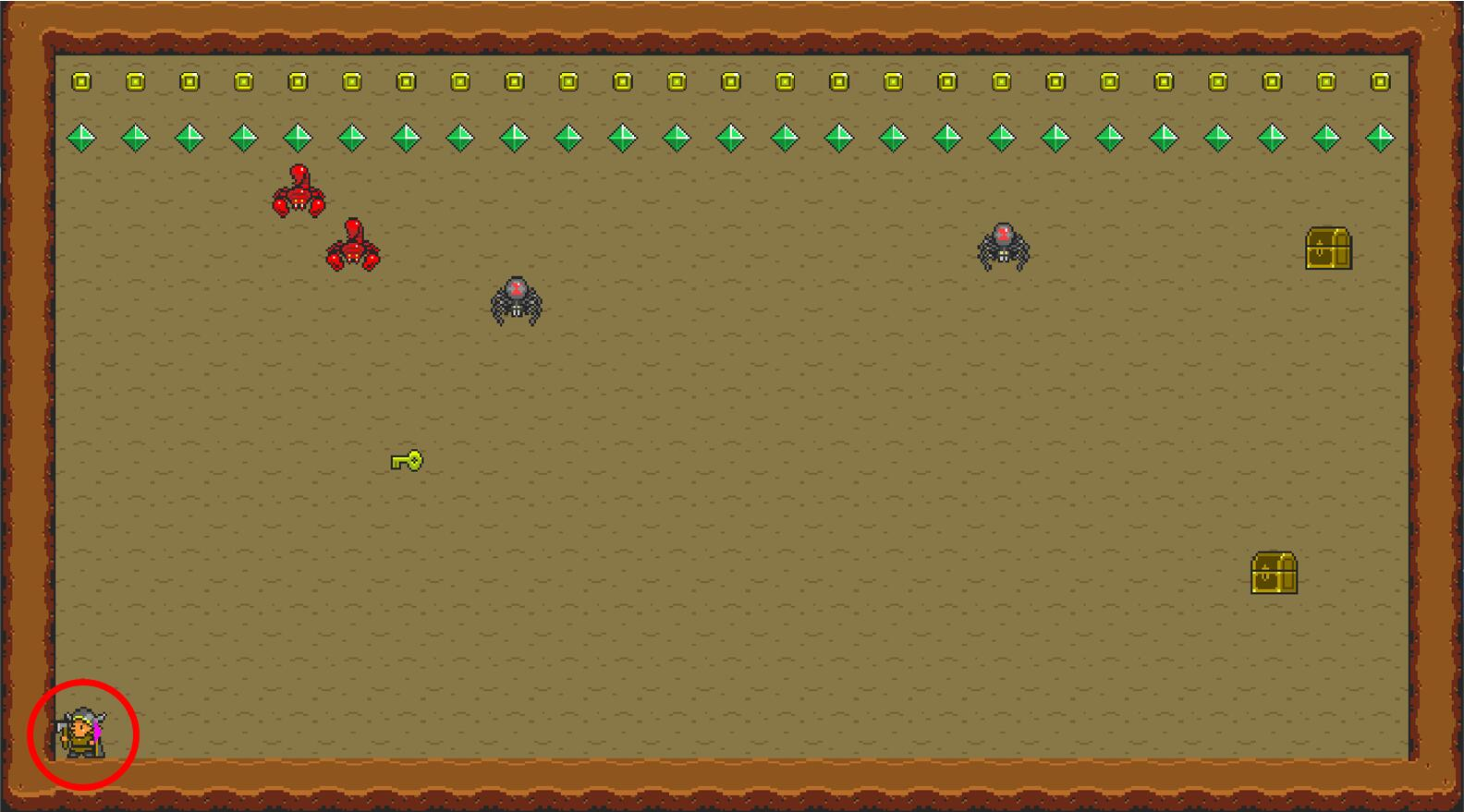}
		\caption{\emph{GoldDigger-0}.}\label{fig:GoldDigger0}
	\end{subfigure}
	%\quad
    \begin{subfigure}[t]{.49\linewidth}
		\centering
		\includegraphics[width=.9\columnwidth]{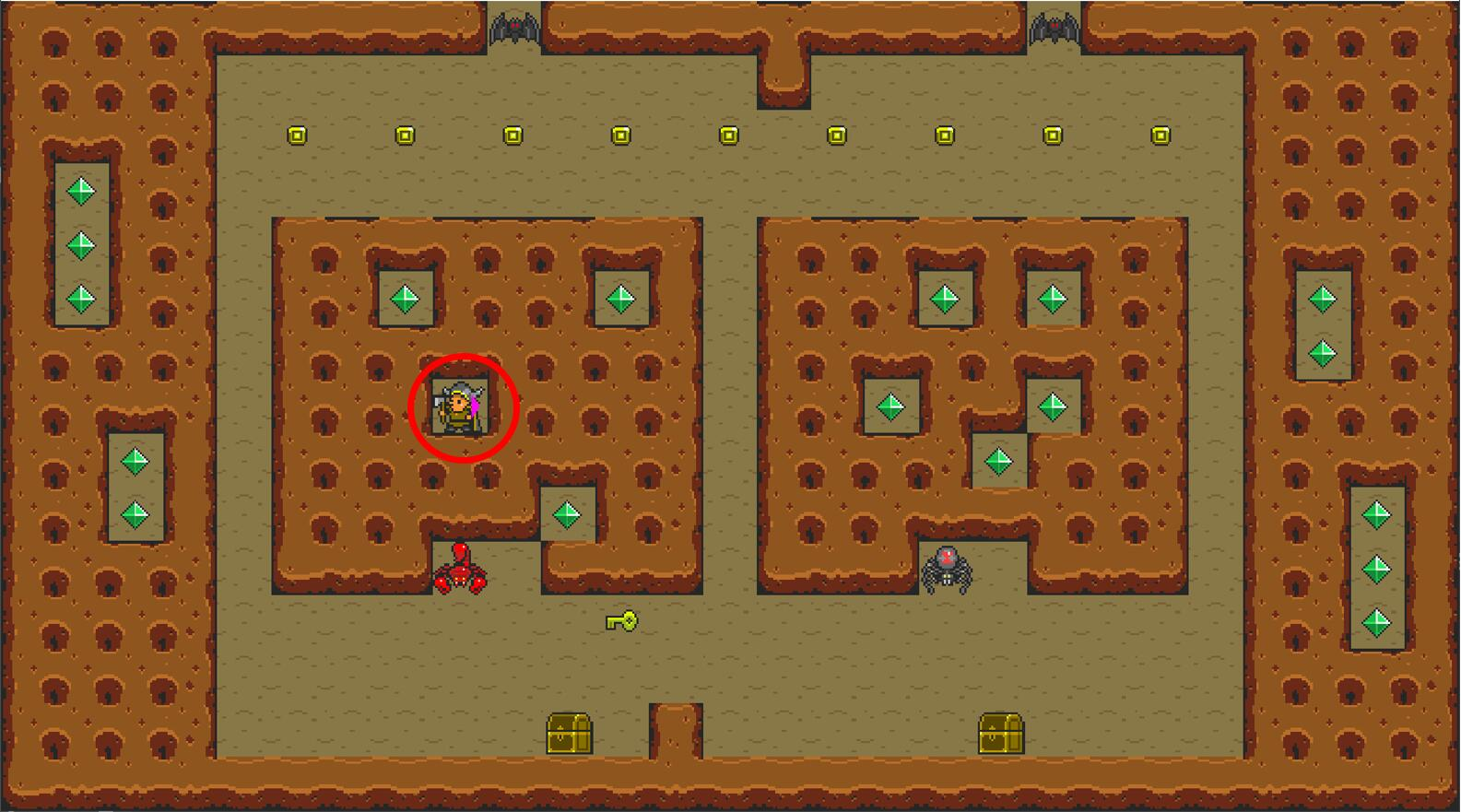}
		\caption{\emph{GoldDigger-1}.}\label{fig:GoldDigger1}		
	\end{subfigure}
	\begin{subfigure}[t]{.49\linewidth}
		\centering
		\includegraphics[width=.9\columnwidth]{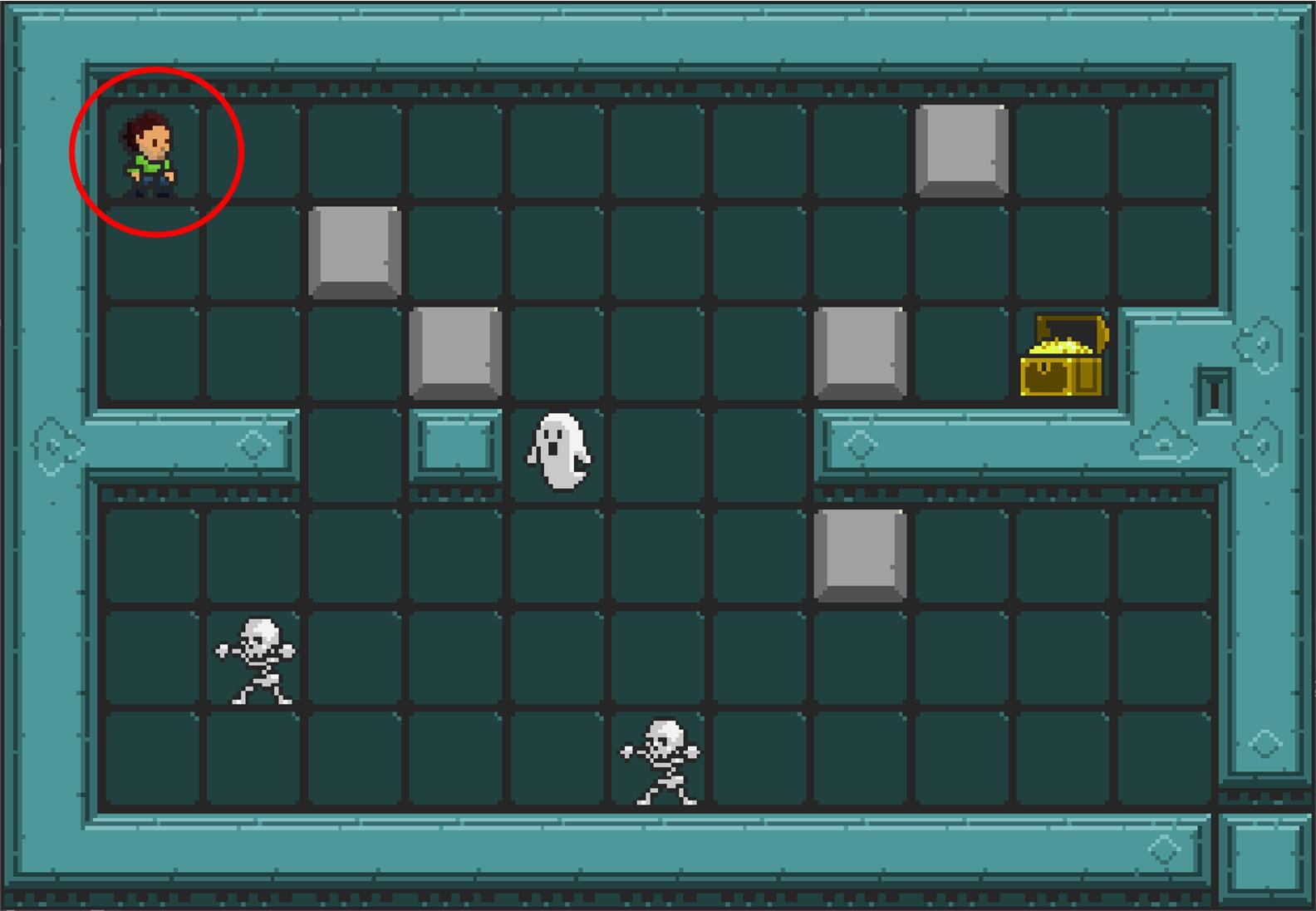}
		\caption{\emph{TreasureKeeper-0}.}\label{fig:TreasureKeeper0}		
	\end{subfigure}
	%\quad
    \begin{subfigure}[t]{.49\linewidth}
		\centering
		\includegraphics[width=.9\columnwidth]{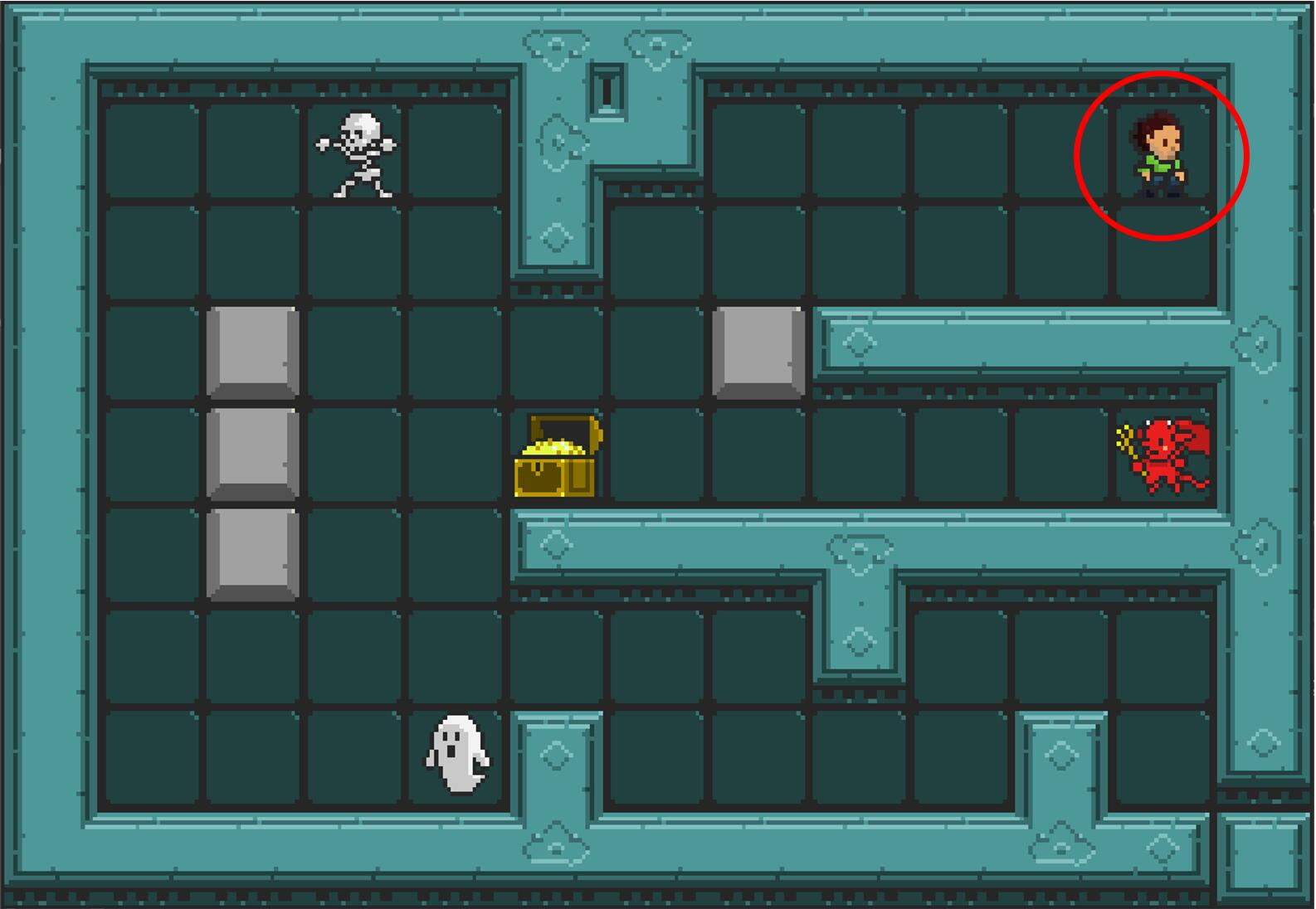}
		\caption{\emph{TreasureKeeper-1}.}\label{fig:TreasureKeeper1}		
	\end{subfigure}
	 \begin{subfigure}[t]{.49\linewidth}
		\centering
		\includegraphics[width=.9\columnwidth]{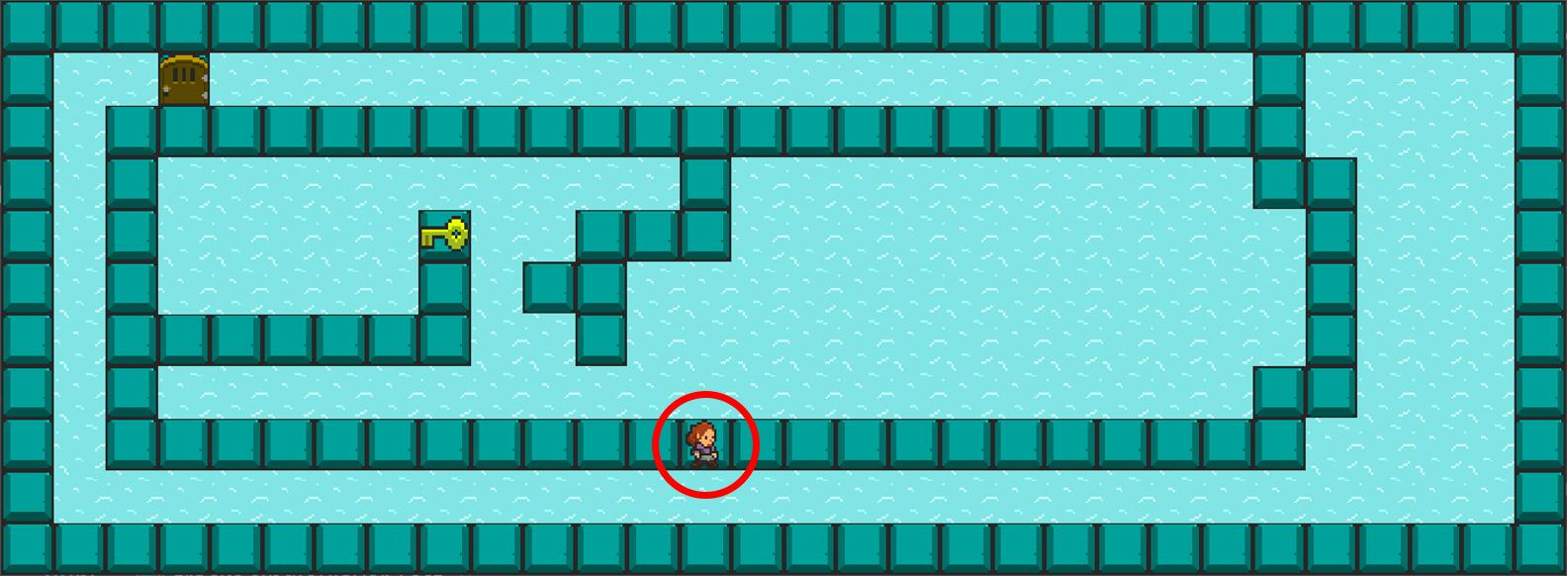}
		\caption{\emph{WaterPuzzle-0}.}\label{fig:WaterPuzzle0}		
	\end{subfigure}
    \begin{subfigure}[t]{.49\linewidth}
		\centering
		\includegraphics[width=1\columnwidth]{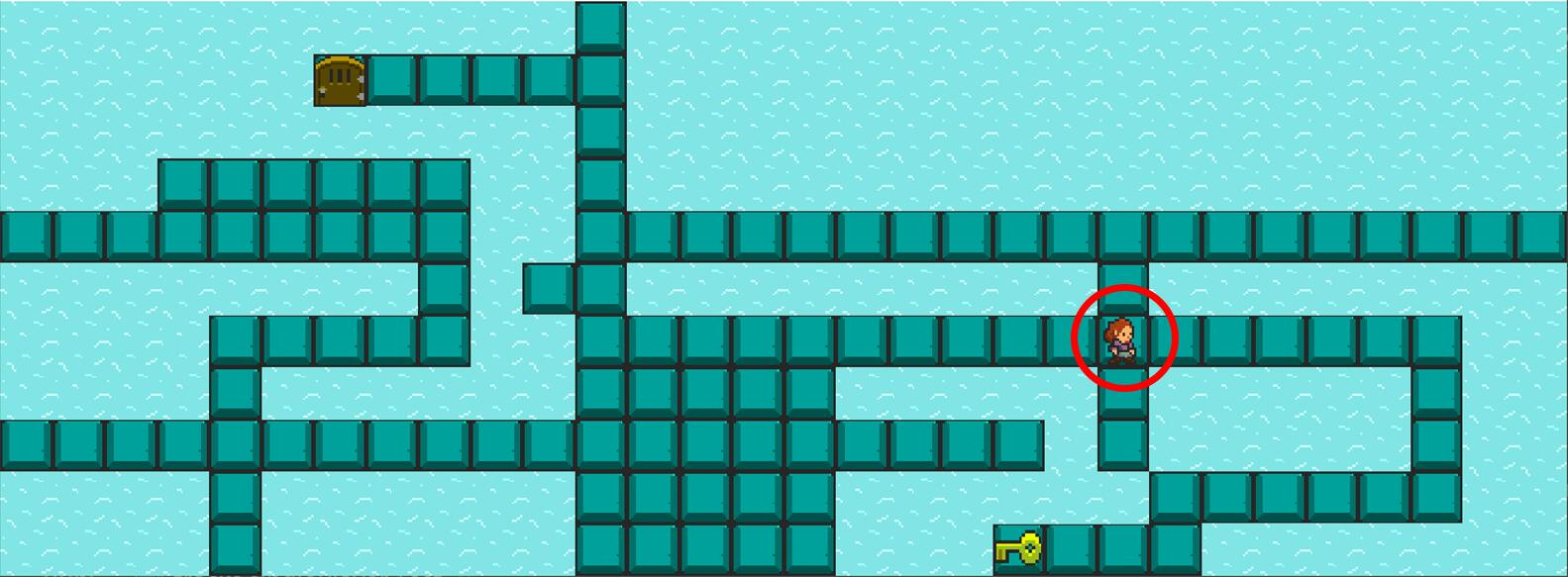}
		\caption{\emph{WaterPuzzle-1}.}\label{fig:WaterPuzzle1}		
	\end{subfigure}
\caption{Training levels in 2020. Red circle indicates the avatar.}
\label{fig:traininglevels}
\end{figure}

\paragraph{Game Set in 2021}
Similarly, the 2021 competition edition\footnote{\url{http://aingames.cn/gvgai/cog2021/}} provides three games, namely \emph{TrappedHero}, \emph{BraveKeeper} and \emph{GreedyMouse}, with dense, periodic and sparse rewards, respectively. Description of the games can be found on the competition website.

\subsubsection{Creation of Competition Level Sets}\label{sec:levelset}

The competition level set for each game is composed of two training levels and three test levels. Instead of using very distinct test levels that were dissimilar to the ones for training as in the 2017--2019 competition editions, all the test levels in the 2020 and 2021 editions were generated from training levels by making \emph{changing one or several tiles} or performing \emph{map combination}, described later in Section \ref{sec:testlevel}. 

\begin{figure}[h!]
\centering
\subfloat{\tikz[remember picture]{\node(C){\includegraphics[width=3cm]{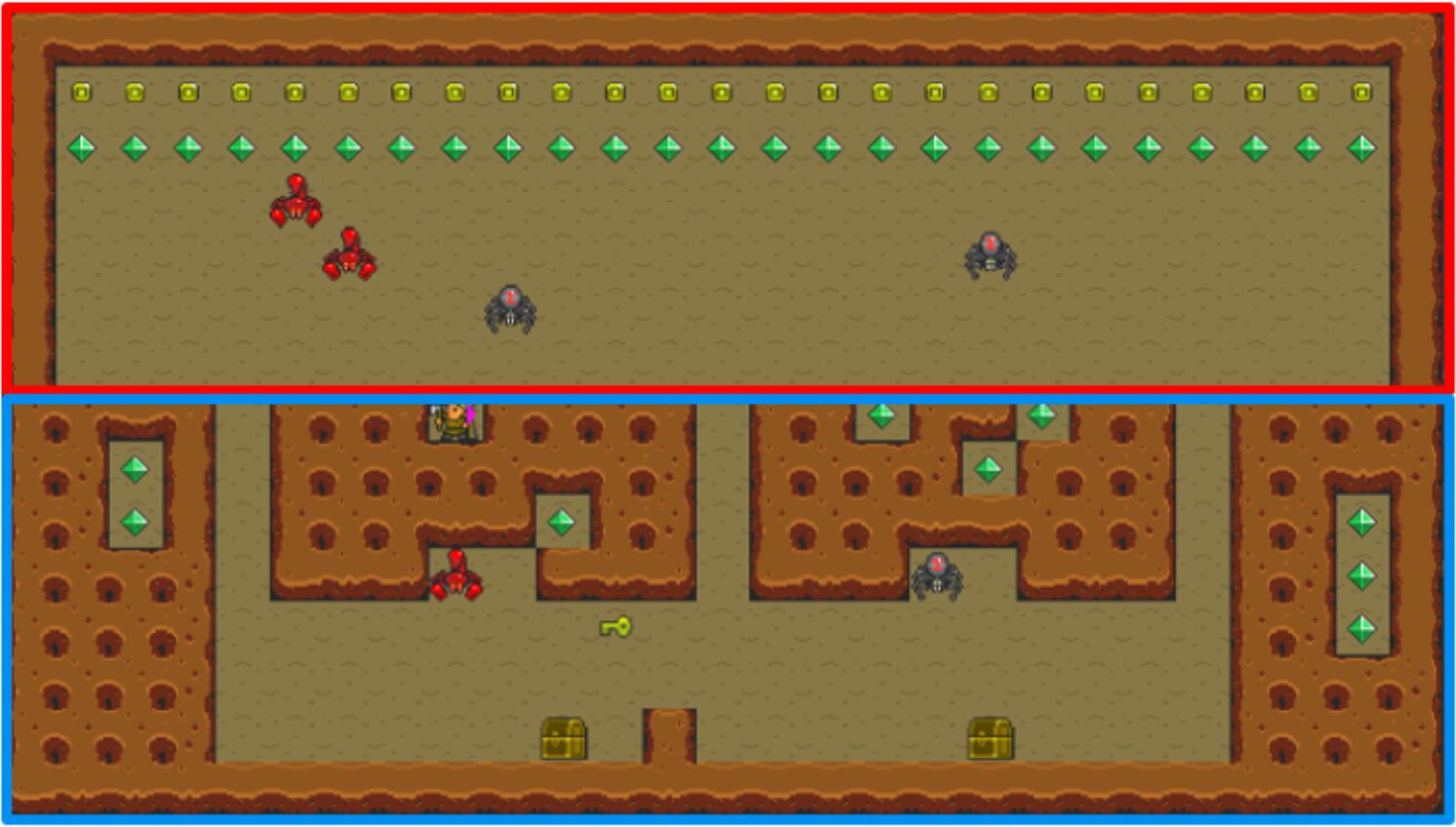}};}}\\
\vspace*{.7cm}
\subfloat{\tikz[remember picture]{\node(1AL){\includegraphics[width=3cm]{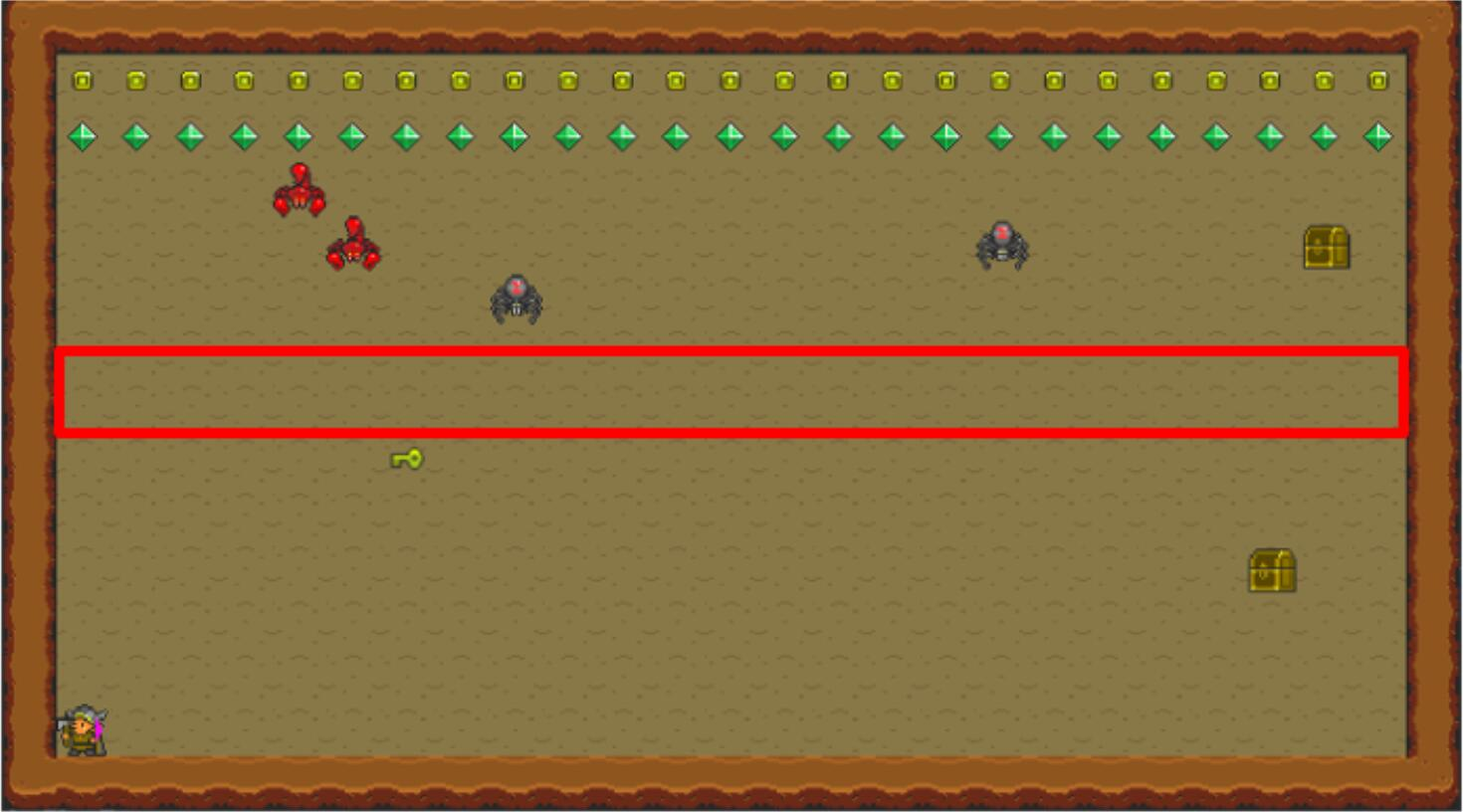}};}}\hspace{1em}
\subfloat{\tikz[remember
picture]{\node(1AR){\includegraphics[width=3cm]{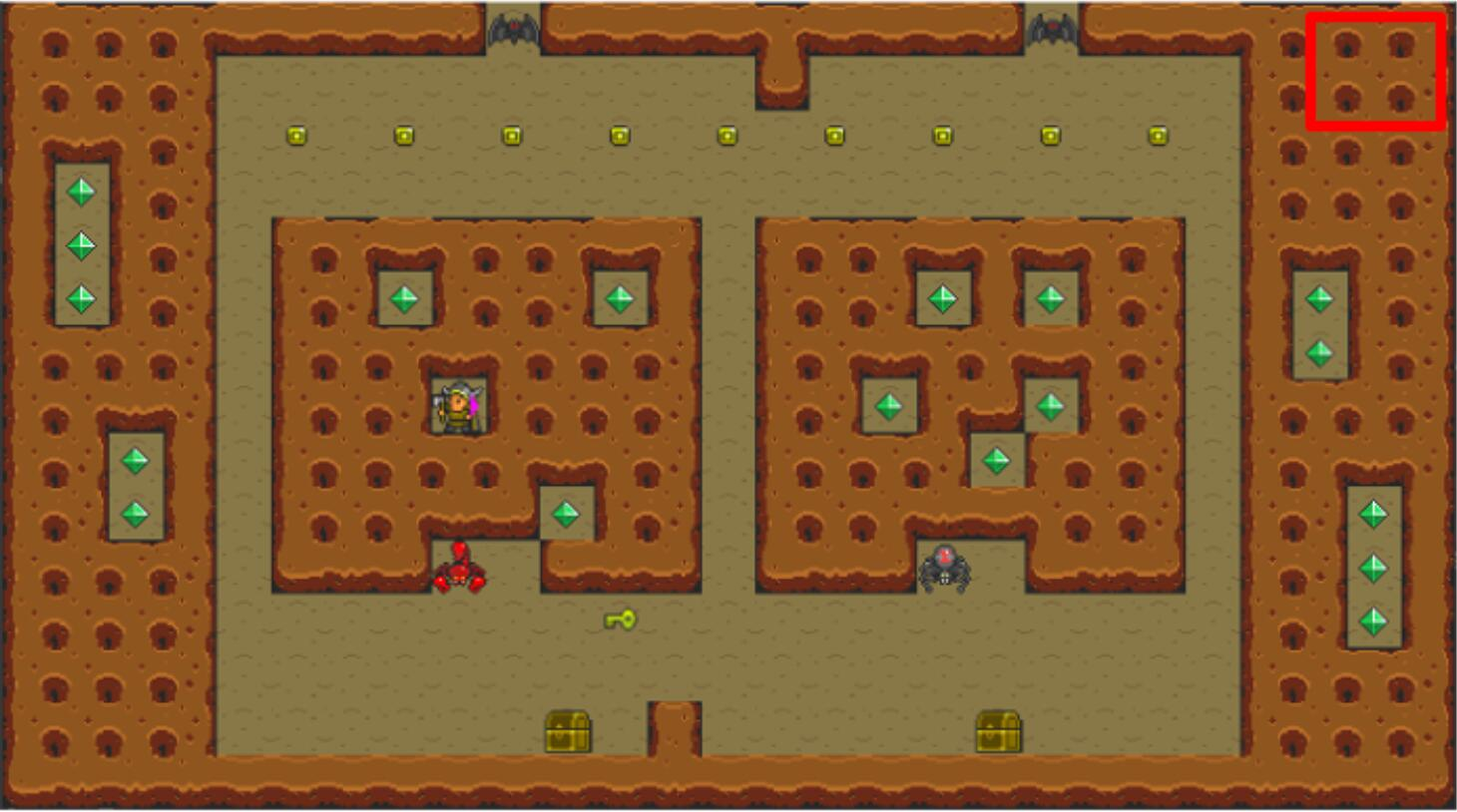}};}}\\
\vspace*{.7cm}
\subfloat{\tikz[remember picture]{\node(2AL){\includegraphics[width=3cm]{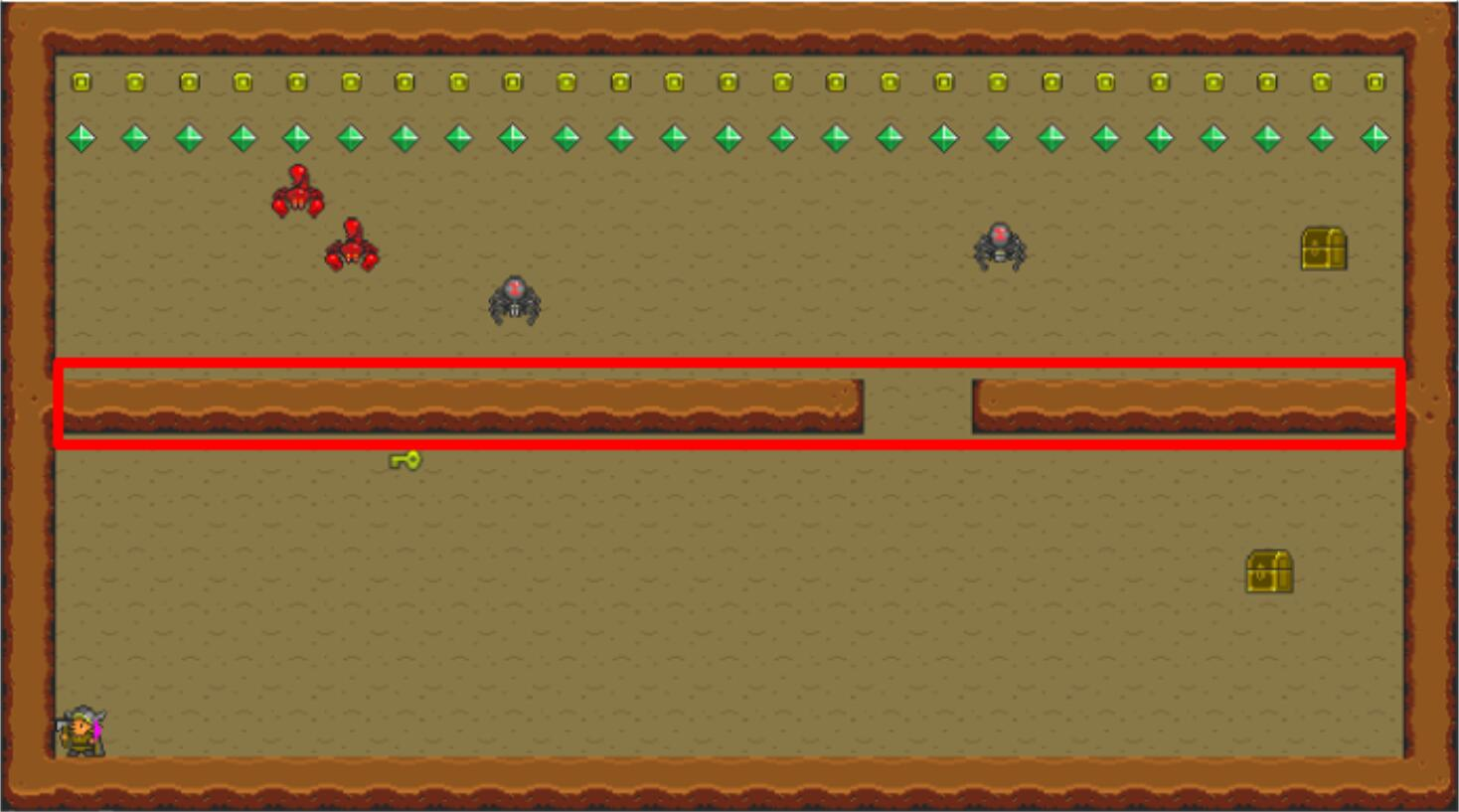}};}}\hspace{1em}
\subfloat{\tikz[remember picture]{\node(2AR){\includegraphics[width=3cm]{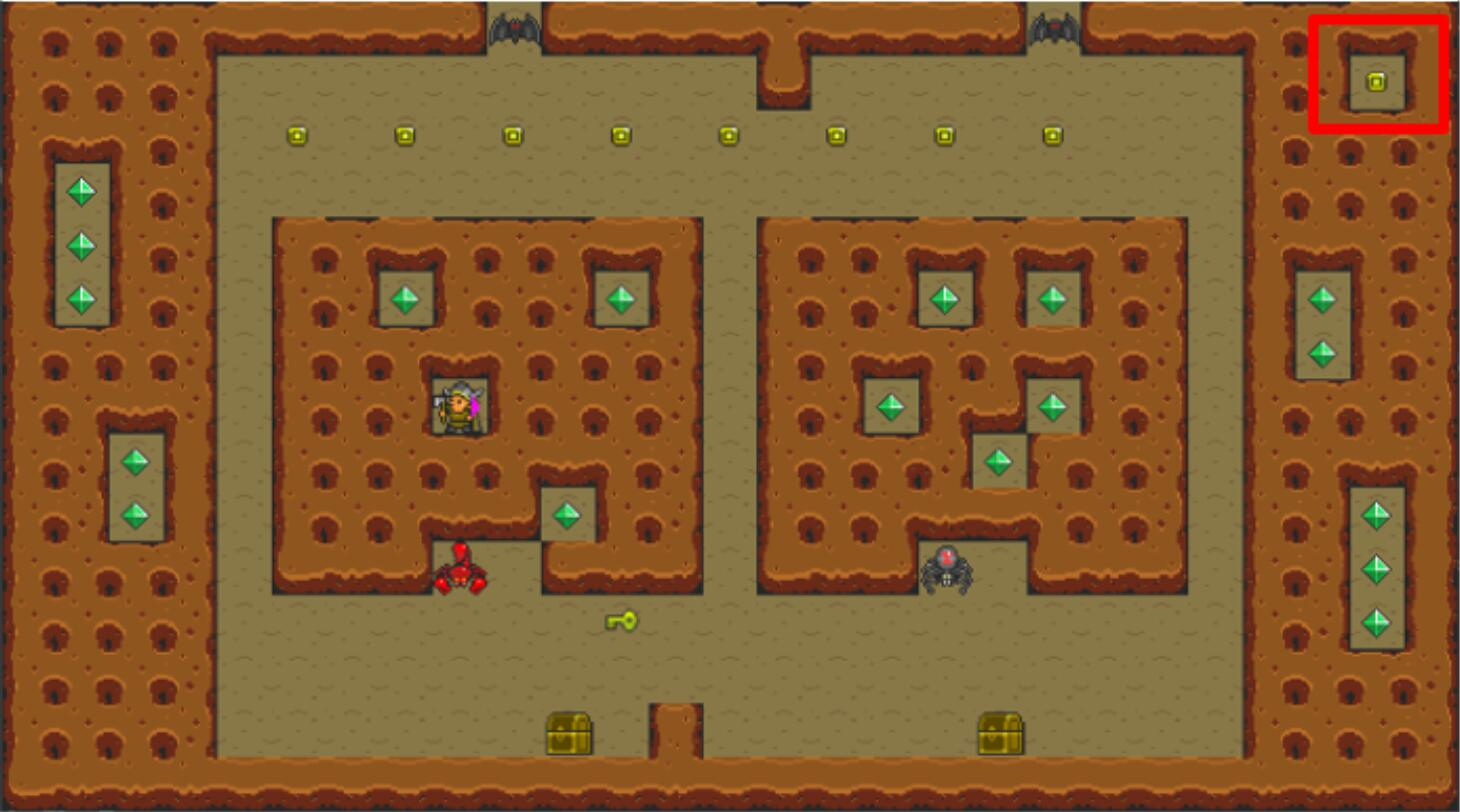}};}}
\caption{\label{fig:mutations}Design of \emph{GoldDigger} test levels.}
\end{figure}
    \begin{tikzpicture}[overlay,remember picture,-,>=latex,shorten >=1pt,auto,node distance=2cm, main node/.style={circle,draw}]
\draw[-latex,thick] (1AL) -- (2AL)
node[midway,left,text width=1.9cm]{Level 0 $\rightarrow$ 2}
node[midway,right,text width=1.5cm]{Multi-tile change}; 
\draw[-latex,thick] (1AR) -- (2AR)
node[midway,left,text width=1cm]{ 1 $\rightarrow$ 3}
node[midway,right,text width=1.5cm]{Single-tile change};
\draw[-latex,thick] (1AR) -- (C)
node[midway,left,text width=1.8cm]{~~~~~Level 4\\Combination};
\draw[-latex,thick] (1AL) -- (C);
\end{tikzpicture}
\paragraph{Training Level Set}
Screenshots of training levels are illustrated in Fig. \ref{fig:traininglevels}. The training levels for each game were designed to have different skill-depth. \emph{GoldDigger-1} is harder than \emph{GoldDigger-0}. On the contrary, more surrounding obstacles in \emph{TreasureKeeper-0} make it easier to protect the treasure chests from monsters. In \emph{WaterPuzzle}, the distances between sprites affect the difficulty of levels. Consequently, \emph{WaterPuzzle} levels not only use different mazes but also differ in sprite locations.

\paragraph{Test Level Set}\label{sec:testlevel}

For each game, a human designer generated three test levels based on training levels with one of the following operators separately.
\begin{itemize}
\item \emph{Single-tile change}:
\emph{Level-3} is designed by changing one single tile from \emph{Level-1}. 
For instance, the single-tile change occurs at the top right corner of \emph{GoldDigger-1}, where an obstacle is replaced by a jewel (Fig.
\ref{fig:mutations}). 
\item \emph{Multi-tile change}:
\emph{Level-2} is designed by changing multiple tiles from \emph{Level-0}. 
Taking \emph{GoldDigger} as an example, most of the tiles in the middle row of \emph{GoldDigger-0} are changed to generate \emph{GoldDigger-2}, a more challenging level due to the obstacles.
\item \emph{Map combination}:
To validate if the knowledge of all training levels has been transferred, we designed the final test level of each game by combining the two training levels. 
The last test level is simply a combination of the top half of \emph{level-0} and the bottom half of \emph{Level-1}. Notably, the avatar remains unique, and its location is changed at each level.
\end{itemize}
For example, Fig. \ref{fig:mutations} illustrates how the test levels are generated for \emph{GoldDigger} using the above operators.

\subsection{Competition Entries and Rankings}\label{sec:ranking}
Competition entries and rankings of the 2017 and 2018 editions have been reported in the work of \cite{perez2019general}. In this paper, we report the entries and rankings of the 2019--2021 editions.

\subsubsection{The 2019 Competition Editions}
Two editions of the GVGAI Learning Competition were held at the 2019 IEEE Congress on Evolutionary Computation (IEEE CEC2019)\footnote{\url{http://aingames.cn/gvgai/cec2019/}} and the 2019 IEEE Conference on Games (IEEE CoG2019)\footnote{\url{http://aingames.cn/gvgai/cog2019/}}, which received two and three entries, respectively.

The final results and rankings of those editions according to the rules described in Section \ref{sec:rules} are shown in Table \ref{tab:competitionres2019}\footnote{Unfortunately, description of the entries, including algorithms and training methods, was not mandatory when submitting an entry to the 2019 competition editions.}. ``Random'' refers to a baseline agent that plays actions uniformly at random.

\begin{table}[htpb]
\centering
\setlength{\tabcolsep}{3pt}
\caption{\label{tab:competitionres2019}Competition rankings of the 2019 editions.}
\subfloat[Competition rankings at IEEE CEC2019.\label{tab:competition2019cec}]{% 
    \begin{tabular}{>{\centering\arraybackslash}p{0.1\linewidth}|>{\centering\arraybackslash}p{0.25\linewidth}|>{\centering\arraybackslash}p{0.25\linewidth}|>{\centering\arraybackslash}p{0.25\linewidth}}
\toprule
Rank & Entry      & Points & \# Wins \\ \midrule
1 & mballa & 174 & 10/180    \\ \hline
2 & Random & 158 & 1/180    \\ \hline
3 & Ben & 151 & 14/180    \\
\bottomrule
\end{tabular}
}

\vspace{2mm}

\subfloat[Competition rankings at IEEE CoG2019.\label{tab:competition2019cog}]{% 

\begin{tabular}{>{\centering\arraybackslash}p{0.1\linewidth}|>{\centering\arraybackslash}p{0.25\linewidth}|>{\centering\arraybackslash}p{0.25\linewidth}|>{\centering\arraybackslash}p{0.25\linewidth}}
\toprule
Rank & Entry     & Points & \# Wins \\ \midrule
1 & Random & 158 & 9/180    \\ \hline
2 & mballa & 153 & 20/180    \\ \hline
3 & TNTBot & 124 & 1/180    \\ \hline
4 & UjiAgent & 120 & 0/180    \\
\bottomrule
\end{tabular}
}
\end{table}

\subsubsection{The 2020 Competition Edition}
The 2020 GVGAI Learning Competition was held at the Sixteenth International Conference on Parallel Problem Solving from Nature (PPSN-XVI) and the 2020 IEEE Conference on Games (IEEE CoG2020). The following five entries were received and ranked together (cf. Table \ref{tab:competitionres2020}).
\begin{itemize} 
    \item \emph{{Xybb}}: Three DQN models were trained for $200,000$ timesteps on each game separately using the two training levels and some new levels manually generated by the authors from the given training levels.
    \item \emph{{Elianentity}}: A DQN model from stable-baselines was trained for 2 million timesteps on \emph{GoldDigger-0} and used in test levels of \emph{GoldDigger}. Random actions were played in the other two games.
    \item \emph{{Visionpack}}: For each of the three games, a PPO2 model from stable-baselines3~\cite{stable-baselines3} was trained on level 0 of each game for 1 million timesteps. While testing on different games, the corresponding model, identified by the size of the game screen, was used to play the test levels.
    \item \emph{{Visionpack2}}: Two separate DQN models from stable-baselines3 were trained on \emph{GoldDigger-0} and \emph{GoldDigger-1}. An image classifier was created to help select a suitable model for \emph{GoldDigger}'s test levels. Random actions were played in the other two games.
    \item \emph{{Visionpack3}}: A PPO2 model from stable-baselines3 was trained on \emph{GoldDigger}'s training levels for 2 million timesteps using the concept of vectorized environments and an ACER model~\cite{wang2016sample} was trained on \emph{Waterpuzzle-0} only for 2 million timesteps. \emph{Treasurekeeper} was played randomly.
\end{itemize}
The detailed results on each game and the running log are presented on the competition website. According to Table \ref{tab:competitionres2020}, \emph{Xybb} performs significantly better than Random and all the other entries, which could not even beat Random.

\begin{table}[htpb]
\centering
        \setlength{\tabcolsep}{3pt}
        \caption{\label{tab:competitionres2020}Rankings of the 2020 competition edition.} 
\begin{tabular}{>{\centering\arraybackslash}p{0.1\linewidth}|>{\centering\arraybackslash}p{0.25\linewidth}|>{\centering\arraybackslash}p{0.25\linewidth}|>{\centering\arraybackslash}p{0.25\linewidth}}
\toprule
Rank & Entry      & Points & \# Wins \\ \midrule
1    & Xybb          & 189    & 10/180      \\ \hline
2    & Random        & 149    & 4/180       \\ \hline
3    & Visionpack2   & 129    & 2/180       \\ \hline
4    & Elianentity   & 109    & 0/180       \\ \hline
5    & Visionpack3   & 102    & 1/180       \\ \hline
6    & Visionpack    & 101    & 2/180       \\ \bottomrule
\end{tabular}

\end{table}

\subsubsection{The 2021 Competition Edition}\label{sec:edition2021}
The 2021 GVGAI Learning Competition was held at the 2021 IEEE Conference on Games (IEEE CoG2021). No entry was received in this edition. A PPO agent with dual-observation, named \emph{Arcane}, described later in Section \ref{sec:arcane}, was used as a baseline agent. Its performance in playing training and test levels of 2021 competition games is illustrated in Table \ref{tab:competitionres2021}.
Besides \emph{Arcane}, the performance of several planning agents, assuming the availability of forward model, are also reported for comparison. Open Loop Expectimax Tree Search (OLETS), the winner of the first GVGAI Single-Player Planning Competition~\cite{perez2014gvgpc}, performs the best on the three games.

Despite the our best attempts at making GVGAI learning platform easy to use, only a few papers have been published describing research done with the platfofm~\cite{kunanusont2017general,Apeldoorn2017AnAL,Dockhorn2018,torrado2018drlgvgai,justesen2018illuminating,ye2020rotation}. Notably, no agents performing much better than the baselines were received in the past competition editions~\cite{perez2019general}.

Another video game-like testbed for testing and enabling generalisation in reinforcement learning is Obstacle Tower, which is a single game but where levels are procedurally generated and multiple aspects of visual input, including lighting and art style, are systematically varied~\cite{juliani2019obstacle}.

\section{RL with Dual-observation}\label{sec:arcane}

As previously reported in Section \ref{sec:ranking}, no significantly effective entry has been reported in the GVGAI Learning Competition editions. The only well-performing ones are the baseline agents, \emph{xybb} and \emph{Arcane}, in the 2020 and 2021 editions, respectively. Those two agents are instantiations of our proposed RL agent with dual-observation and tile-vector encoding detailed in this section.

\begin{figure*}[htbp]
    \centering
    \includegraphics[width=1\textwidth]{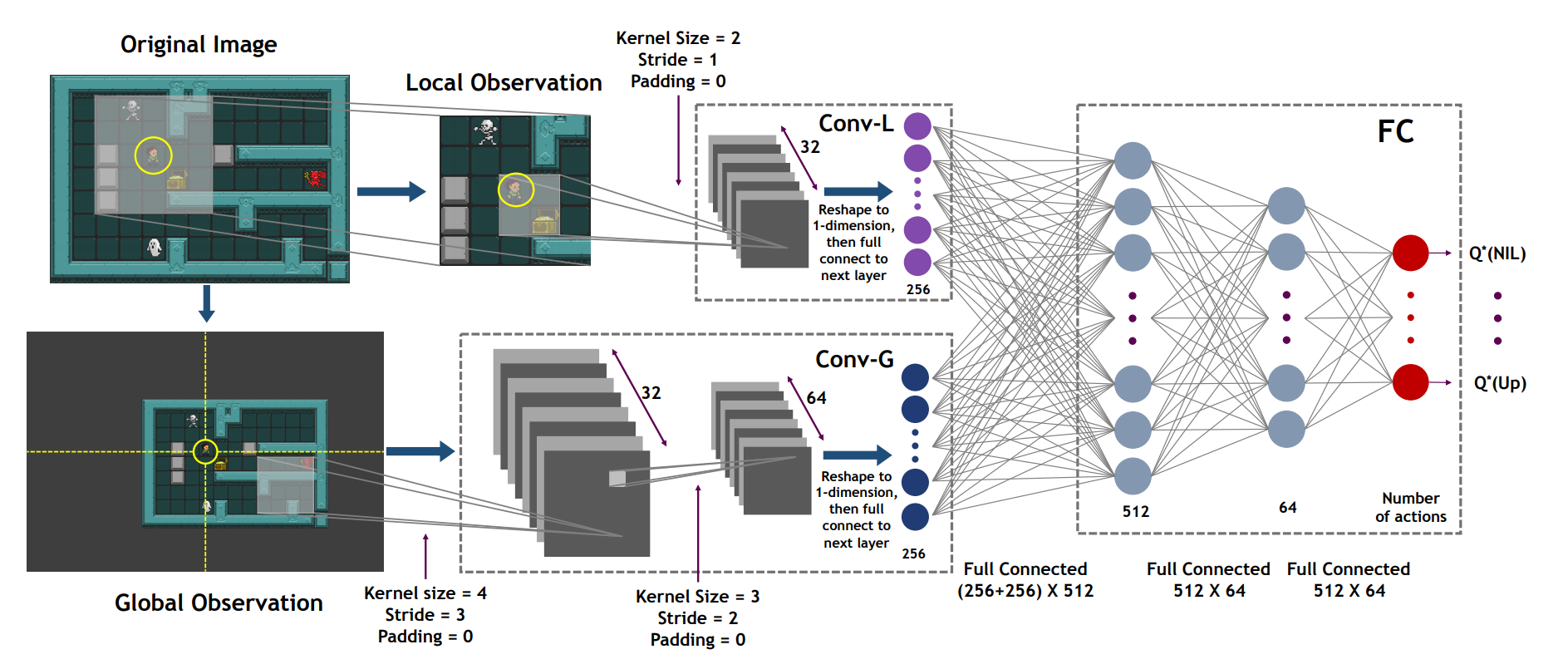}
    \caption{Illustration of reinforcement learning with dual-observation.}
	\label{fig:structure}
\end{figure*}
The agents submitted to the 2017--2020 competition editions used the raw screenshots of the whole game screen as a single input~\cite{perez2019general}. Under the assumption that local information has a higher chance of remaining invariant across different levels, we design a dual-observation as the input to RL agents (``DORL'' for short) that makes use of the local information during training and test. DORL is illustrated in Fig. \ref{fig:structure}. DORL processes the screenshot of the game screen into two observations (cf. Section \ref{sec:obs}), a global one and a local one, and then encodes them with tile-vector encoding and one-hot matrices (cf. Section \ref{sec:encoding}), which are later used as inputs of the learning model (cf. Section \ref{sec:network}). DORL can be implemented with various learning algorithms. When using DQN, either the deterministic policy or the stochastic one can be used by DORL for determining the action to play (cf. Section \ref{sec:decision}). Section \ref{sec:novelty} clarifies the novelty of this work.

\subsection{Transformed Game Observations}\label{sec:obs}

\begin{figure}[htbp]
    \centering
    \includegraphics[width=\linewidth]{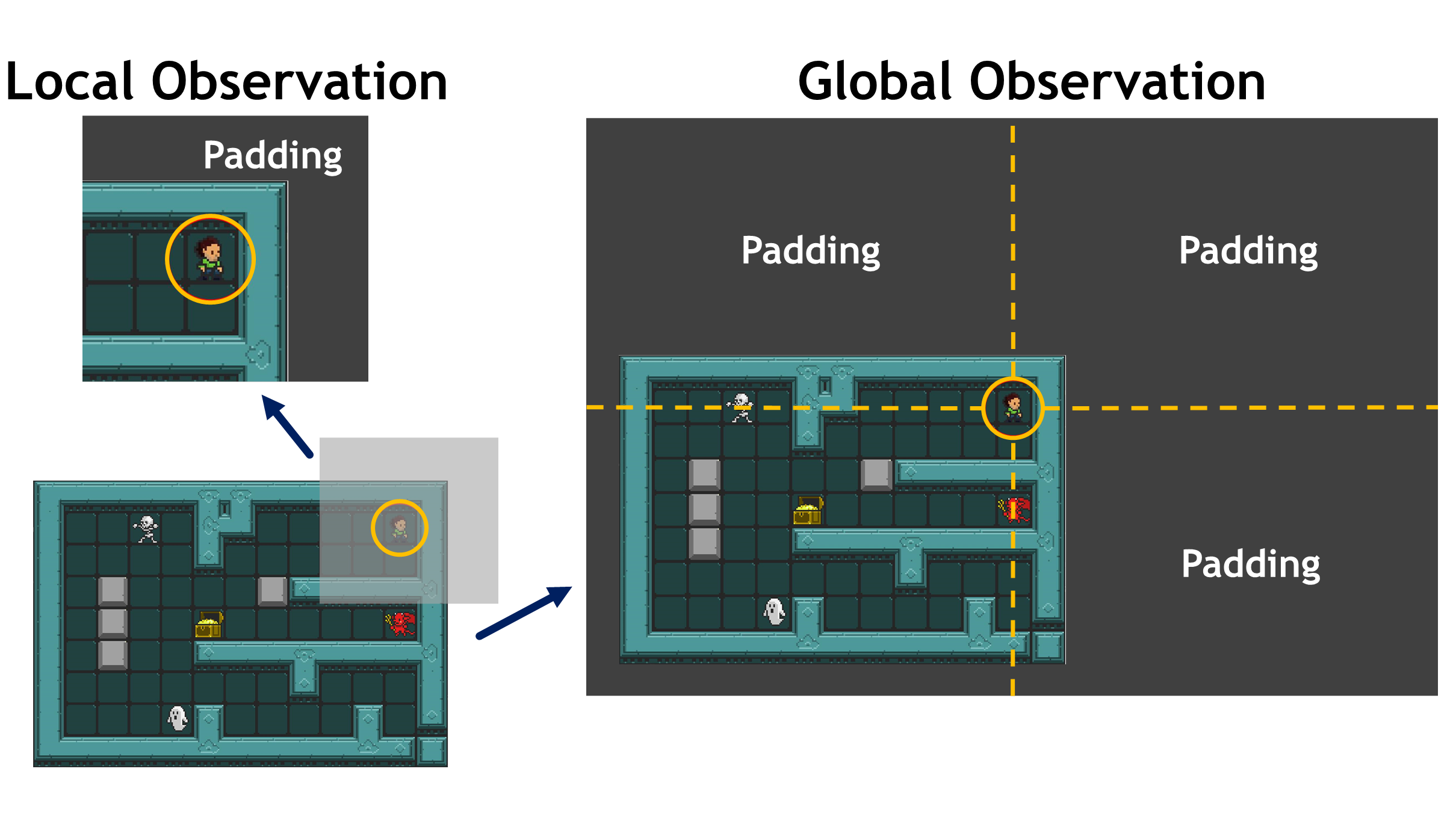}
    \caption{Illustration of transformed game observations.}
    \label{fig:tgo}
\end{figure}

Different areas of the game observation are of different importance. The avatar's surrounding area usually has a more immediate and greater impact than the distant areas on the action selection. Therefore, DORL is proposed to transform the received screenshot of the game screen into a global observation and a local one. The transformation procedure is described as follows and illustrated on the left part of Fig. \ref{fig:structure}.
\begin{itemize}
    \item A \emph{global observation (GO)} is not the direct use of the screenshot of the game screen. Instead, it is a $h_g \times w_g$ RGB image in which the avatar is placed at centre (cf. Fig. \ref{fig:tgo}), where $h_g = 2h-h_{tile}$ and $w_g = 2w-w_{tile}$, $w$ and $h$ refers to the width and height of the original screenshot, $w_{tile}$ and $h_{tile}$ refer to the width and height of one single tile, respectively. This guarantees that the original game screen is always contained in the global observation wherever the avatar moves. 
\item A \emph{local observation (LO)} is a $h_l \times w_l$ RGB image ($h_l<w$, $h_l<h$) in which the avatar is centred (cf. Fig. \ref{fig:tgo}). In our experiments, $h_l$ and $w_l$ are set as $50$, which means $5 \times 5$ surrounding tiles are used.
\end{itemize}
In the context of VGDL~\cite{schaul2013video,schaul2014extensible}, the avatar can be easily located with the tile dictionary described in the following subsection. 

\subsection{Tile-vector Encoded Inputs}\label{sec:encoding}
The GVGAI platform represents levels with tile-based maps~\cite{schaul2013video,schaul2014extensible}. Those tiles, either for different types of sprites or for accessible areas, have a predefined, fixed size of $10 \times 10$~\cite{perez2019general}. Hence, we convert the RGB images into tile-based matrices as input, on which one-hot encoding can be easily applied.

All distinct tiles appeared in the training levels have been collected to build a \textit{tile dictionary}, which is a set of $\langle$\emph{code}, \emph{reference vector}$\rangle$ tuples. The $code \in \mathbf{N}$ identifies a unique $10 \times 10$-pixel \emph{tile image}. For each tile, 5 pixels are selected from its $4^{th}$ row with a stride of $2$, and another 5 are selected from its $5^{th}$ column with a stride of $2$. For each selected $10$ pixels, its averaged RGB value is calculated as $\frac{1}{3}(v_R+v_G+v_B)$, where $v_R$, $v_G$ and $v_B$ refer to the red, green and blue values of the pixel, respectively. As a result, a \emph{reference vector} composed of $10$ averaged RGB values is obtained.

At every game tick during training or testing, the two RGB images of transformed global and local observations are considered as $\frac{h_g}{10} \times \frac{w_g}{10}$ and $\frac{h_l}{10} \times \frac{w_l}{10}$ grids of tiles, respectively. The blank spaces added during transformation, which are not part of the game screen, are replaced by zeros. For both grids, we can easily calculate the corresponding RGB-value vector for every tile inside and find its closest reference vector in the dictionary, measured by the Manhattan distance. Assuming that some unexpected tiles may appear in the test levels, we find the closest tile with distance between reference vectors instead of simply finding the code with the same reference vector. Then, the tile-based grids are converted to two matrices of tile codes, i.e., integers. Finally, one-hot encoding is applied. Since there are no more than $16$ tile types in each game, the size of final matrix is $\frac{h_g}{10} \times \frac{w_g}{10} \times 16$ for the transformed global observation and $\frac{h_l}{10} \times \frac{w_l}{10} \times 16$ for the local one.

\subsection{Network Architecture}\label{sec:network}

We use a convolutional neural network (CNN). The CNN consists of $3$ modules, namely Conv-G, Conv-L and FC. The Conv-G and the Conv-L modules are used to process the encoded global and local observations, respectively.

For agents using pixel input, the Conv-G module is composed of three convolutional layers with $\langle channel$, $kernel~ size$, $stride \rangle$ of $\langle 32, 8, 4 \rangle$, $\langle 64, 4, 3 \rangle$ and $\langle 64, 3, 1 \rangle$. The Conv-L module is composed of two convolutional layers with $\langle channel$, $kernel~ size$, $stride \rangle$ of $\langle 32, 6, 4 \rangle$, $\langle 64, 4, 2\rangle$. We consider agents using one-hot input and pixel input separately.

For agents using one-hot input, the Conv-G module is composed of two convolutional layers with $\langle channel$, $kernel~ size$, $stride \rangle$ of $\langle 32, 4, 3 \rangle$ and $\langle 64, 3, 2 \rangle$. The Conv-L module has only one convolutional layer with $\langle channel$, $kernel~ size$, $stride \rangle$ of $\langle 32, 3, 1 \rangle$.

The Conv-G module and the Conv-L module of agents using pixel input and one-hot input are different because the dimensionality of pixel input is larger than the one-hot input, so it requires much more down-sampling for the model of agents using pixel input.

All of those convolutional layers use ReLU function as activation. Both Conv-G's and Conv-L's outputs are reshaped to a vector and then fully connected to a linear layer with $256$ outputs. The FC module has two $512 \times 64 \rightarrow 64 \times \#actions$ linear layers with ReLU activation function and no activation function respectively, where $\#actions$ denotes the number of actions. For other agents with global observation only in this paper, the Conv-L module is removed and the first layer of FC just uses a $256 \times 64$ linear layer.

\subsection{Decision Policy}\label{sec:decision}
Action selection strategy is a core component of RL algorithms. It directly affects the experience transitions got by agents. When training DQN agents, we use the $\epsilon$-greedy strategy with linearly decreasing $\epsilon$ value. while testing, the following two decision policies have been considered.

The first one uses a greedy approach, i.e., a DQN agent takes action with the highest Q-value. It is referred to as the \emph{deterministic policy} in our paper. A greedy decision policy can clearly show the training effect of agents. However, sometimes it may be too strict and rigid for an agent. 

To encourage exploration, we also design a \emph{stochastic policy} to make DQN agents not always select the action with the highest Q-value but also the ``sub-optimal'' ones. However, the ordinary epsilon greedy that controls exploration probability proportional to Q-value is not suitable in our case since playing an action shown to be insufficient during training is more likely to be an impaired decision on test levels. Therefore, we design a ``scaled softmax'' that decreases the probability of selecting actions with very low Q-values, described as follows. First, a probability distribution is computed over actions $\mathcal{A}$ based on their Q-values $\mathbf{Q}^*$. For any action $a \in \mathcal{A}$, its probability of being selected is defined as $$p_a = \frac{ exp({k \cdot q'_a})}{\sum_{i \in \mathcal{A}} exp({k \cdot q'_i})},$$ 
where $q'$ denotes the Q-value shifted by $-(\max{\mathbf{Q}^*}+\min{\mathbf{Q}^*})/2$, and $k=\sigma/(\max{\mathbf{Q}^*}-\min{\mathbf{Q}^*})$ to avoid overflow. $\sigma$ is a control parameter. Then, an action randomly selected according to the resulting probability distribution is played. The parameter $\sigma$ is introduced to control how greedy the strategy is. The higher the $\sigma$ is, the higher selection probability the action with maximum Q-value has.

\subsection{Novelty of This Work}\label{sec:novelty}

It is worth mentioning that the work of \cite{ye2020rotation}, which has some similarities with ours, only used the raw image of either the global or the local observation as a single input. DORL not only uses both inputs simultaneously but also adapts the structure of the neural network to take tile-vector encoded observations as inputs. Moreover, only A2C was considered in \cite{ye2020rotation}, while DORL is a more general technique, and its instantiations using several different state-of-the-art RL algorithms, including A2C, DQN, and PPO, are implemented and compared in Section \ref{sec:xp}. The work of \cite{kostrikov2020image} and \cite{laskin2020reinforcement} used image transformation as a data augmentation technique during training, but their agents were trained and tested on identical game levels thus the test levels are also training levels. However, in our case, the test levels are unseen during training. Moreover, our proposed technique preprocesses the inputs with the novel tile-vector encoding method aiming at accelerating the process of encoding observations and handing  the possible appearance of new tiles in unseen test levels.

\section{Experimental Study}\label{sec:xp}
DORL is implemented using three different state-of-the-art RL algorithms (PPO, A2C and DQN) to validate its performance. For the purpose of examining the benefit brought by different components of DORL, for each RL algorithm, four groups of agents categorised by input format (one-hot or pixel input; GO alone or dual-observation), as summarised in Table \ref{tab:notations}, have been considered and compared on the 2020 GVGAI Learning Competition games (Sections \ref{sec:games2020} and \ref{sec:furtherdiscussion}). The PPO agent with DORL is used as a baseline agent in the 2021 GVGAI Learning Competition (Section \ref{sec:games2021}). Experimental details are provided in Section \ref{sec:setting}.

\begin{table}[htbp]
    \centering
    \caption{\label{tab:notations}Four groups of agents categorised by input.}
    \begin{tabular}{ccccc}
    \toprule
    \multirow{2}{*}{Notation} & \multicolumn{4}{c}{Input} \\
    & Pixel & One-hot & GO & LO \\
    \midrule
    G1 & $\surd$ & & $\surd$ & \\
    G2 & $\surd$ & & $\surd$ & $\surd$ \\
    G3 & & $\surd$ & $\surd$ &  \\
    DORL & & $\surd$ & $\surd$ & $\surd$ \\
    \bottomrule
    \end{tabular}
\end{table}

\subsection{Experimental Setting and Baseline}\label{sec:setting}

These four groups of agents (cf. Table \ref{tab:notations}) were trained separately with DQN, PPO, and A2C, adapted from stable-baselines3~\cite{stable-baselines3}. All DQN agents were tested with the deterministic policy and the stochastic policy, denoted as ``DQN'' and ``$\text{DQN}_\text{s}$'', respectively. 

As a baseline, the agent of Ye \emph{et al.} \cite{ye2020rotation} is applied with all its three operations, \emph{rotation}, \emph{transformation} and, \emph{crop}, and trained with its default parameters, network structure, and pixel input~\cite{ye2020rotation}. Additionally, the performance of four planning agents given forward models of games, including Monte Carlo Tree Search (MCTS)~\cite{liebana2019general}, Open Loop Expectimax Tree Search (OLETS)~\cite{liebana2019general}, Rolling Horizon Evolution (RHEA)~\cite{perez2013rolling}, and Random Search (RS), are also reported.

All learning agents were trained with the same procedure on the same machine with an Intel Xeon Gold 6240 CPU and four TITAN RTX GPUs. A learning agent is trained for 1,000,000 timesteps for each game on the two randomly alternated training levels. Three mono-agents are trained to build a meta-agent. When testing, the meta-agent first determines the game by its screen size and then deploys the corresponding mono-agent. All test experiments are repeated independently $20$ times. Experimental setting and network configuration are summarised in Table \ref{tab:hyperpara}.

\begin{table}[htbp]
\centering
\setlength{\tabcolsep}{10pt}
\caption{\label{tab:hyperpara}Experimental setting and network configuration. More detailed description of parameters can be found in \cite{stable-baselines3}.}

\begin{tabular}{lll}
\toprule
 & Hyper-parameter & Value\\ \midrule
\multirow{2}{*}{All} & Total training steps            & 1,000,000 \\
& Discount factor                 & 0.99\\
& $\sigma$ of scaled softmax & 3\\
\midrule
\multirow{6}{*}{DQN} 
& Initial exploration             & 1
 \\ 
& Final exploration               & 0.1
 \\ 
& Replay memory size              & 40,000
 \\ 
& Learning starts                 & 0
 \\ 
& Mini-batch size                 & 32
\\ 
& Learning rate                   & 0.001
 \\ 
\midrule
\multirow{3}{*}{A2C} 
& Rollout length             & 5\\
& Learning rate             & 0.0007\\
& Number of environments & 8\\
\midrule
\multirow{4}{*}{PPO} 
& Rollout length             & 2,048
 \\ 
& Learning rate             & 0.0003\\
& Number of environments & 8
\\ 
\bottomrule
\end{tabular}
\end{table}

\begin{figure}[htbp]
    \centering
    \begin{subfigure}[t]{.49\columnwidth}
		\centering
		\includegraphics[width=1\columnwidth]{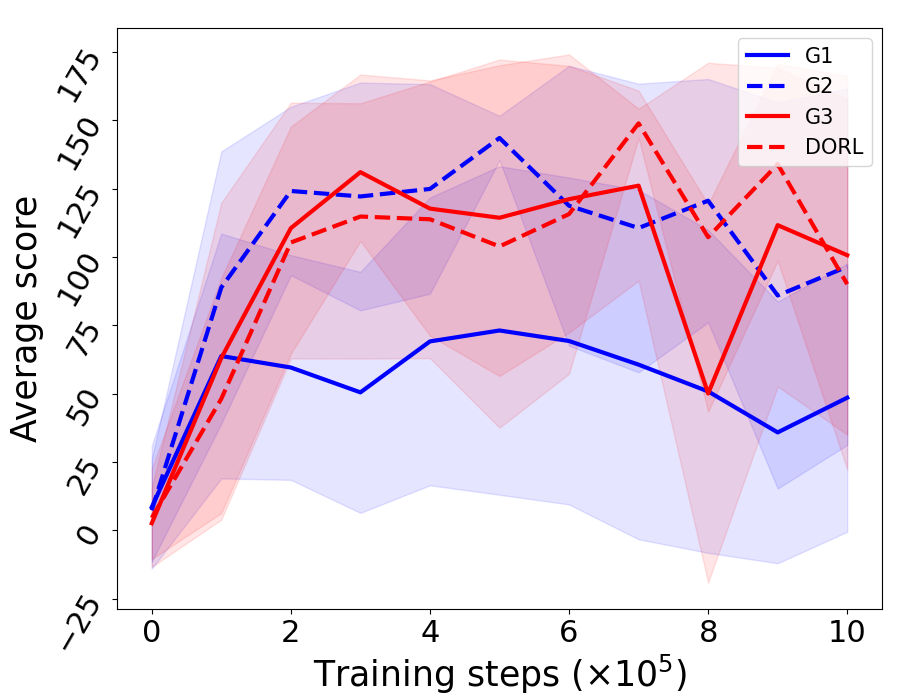}
		\caption{\emph{GoldDigger-0}.}\label{fig:GoldDigger0_ppo}		
	\end{subfigure}
	   \begin{subfigure}[t]{.49\columnwidth}
		\centering
		\includegraphics[width=1\columnwidth]{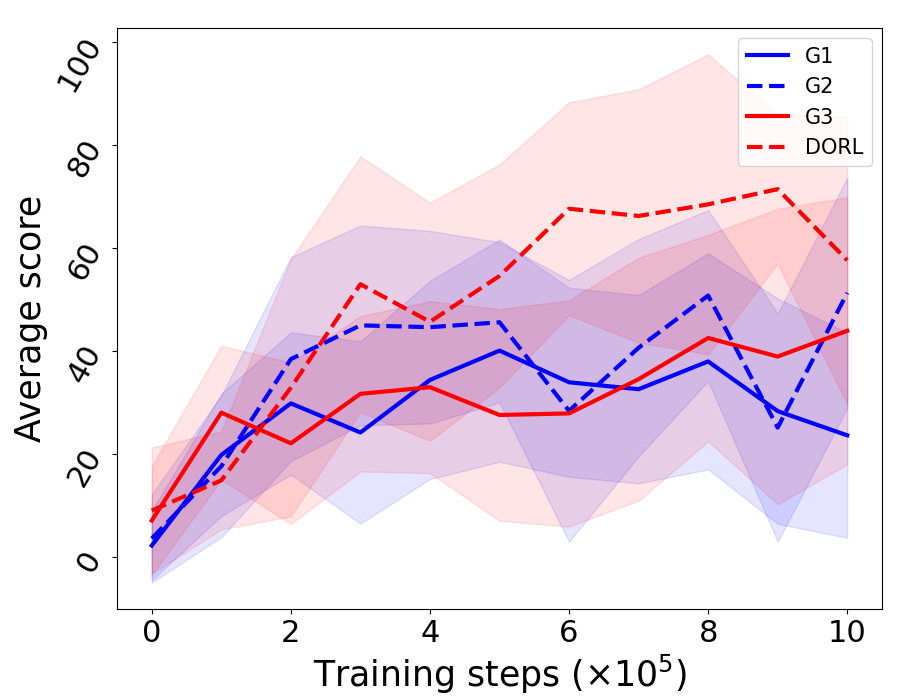}
		\caption{\emph{GoldDigger-1}.}\label{fig:GoldDigger1_ppo}
	\end{subfigure}
	\caption{\label{fig:ppo_train}Average scores obtained by PPO agents on \emph{GoldDigger}'s training levels over 20 independent trials each.}
\end{figure}

\subsection{Comparing Agents on The 2020 Competition Games}\label{sec:games2020}

The average score over $20$ independent runs, the highest score, the lowest score, and the number of wins obtained by each trained agent on the training levels and another 20 trials after training on the test levels of the 2020 competition games, \emph{GoldDigger}, \emph{TreasureKeeper}, and \emph{WaterPuzzle}, are provided in Tables \ref{tab:golddigger_result}, \ref{tab:treasurekeeper_result} and \ref{tab:waterpuzzle_result}, respectively. Wilcoxon rank-sum test has been performed on the game scores reported in Tables \ref{tab:golddigger_result}, \ref{tab:treasurekeeper_result} and \ref{tab:waterpuzzle_result}, and the results are provided in \emph{Appendix}. As an illustrative example, Fig. \ref{fig:ppo_train} shows the training curves of the four PPO agents with different input types (G1, G2, G3 and DORL as indicated in Table \ref{tab:notations}).

\begin{table*}[!ht]
\centering
\setlength{\tabcolsep}{0.45pt}
\caption{\emph{GoldDigger}: Average score, standard deviation and number of wins (in the brackets) over 20 independent trials. The bold numbers denotes the highest average scores among four input types or planning agents.}
\begin{tabular}{cc|cccc|cccccc}
\toprule
\multicolumn{2}{c|}{\multirow{2}{*}{Agent}}
&\multicolumn{2}{c}{\emph{Level-0 (167 / {-}20)}}& \multicolumn{2}{c|}{\emph{Level-1 (98 / {-}20)}} & \multicolumn{2}{c}{\emph{Level-2 (167 / {-}20)}} & \multicolumn{2}{c}{\emph{Level-3 (98 / {-}20)}} & \multicolumn{2}{c}{\emph{Level-4 (196 / {-}30)}} \\ 
&&Avg. $\pm$ Std &Max/Min(Wins)&Avg. $\pm$ Std &Max/Min(Wins)&Avg. $\pm$ Std &Max/Min(Wins)&Avg. $\pm$ Std &Max/Min(Wins)&Avg. $\pm$ Std &Max/Min(Wins)\\ 
\midrule 

\multirow{4}{*}{DQN}&\multirow{1}{*}{G1} &$-4.8\pm1.1$  &0.0/-5.0(0) &$-0.8\pm1.8$  &0.0/-5.0(0) &$-2.0\pm2.4$  &0.0/-5.0(0) &$-0.2\pm1.1$  &0.0/-5.0(0) &$-1.5\pm3.2$  &4.0/-5.0(0)\\
&\multirow{1}{*}{G2} &$40.6\pm41.3$  &78.0/-5.0(0) &$0.0\pm0.0$  &0.0/0.0(0) &$-1.5\pm2.3$  &0.0/-5.0(0) &$0.0\pm0.0$  &0.0/0.0(0) &$88.0\pm64.7$  &144.0/-5.0(0)\\
&\multirow{1}{*}{G3} &$22.8\pm2.5$  &29.0/17.0(0) &$4.7\pm1.5$  &6.0/-1.0(0) &$-2.5\pm2.5$  &0.0/-5.0(0) &$5.3\pm0.6$  &7.0/4.0(0) &$-0.5\pm3.7$  &5.0/-5.0(0)\\
&\multirow{1}{*}{DORL} &$\mathbf{129.1}\pm44.3$  & 159.0/23.0(0) &$\mathbf{41.3}\pm20.1$  & 81.0/19.0(0) &$\mathbf{-1.2}\pm2.2$  & 0.0/-5.0(0) &$\mathbf{27.9}\pm1.9$  & 31.0/23.0(0) &$\mathbf{99.4}\pm26.1$  & 116.0/26.0(0)\\
\midrule 
\multirow{4}{*}{$\text{DQN}_\text{s}$}&\multirow{1}{*}{G1} &$-3.2\pm2.4$  &1.0/-5.0(0) &$0.2\pm5.8$  &18.0/-5.0(0) &$-0.2\pm1.1$  &0.0/-5.0(0) &$3.2\pm9.1$  &23.0/-5.0(0) &$15.4\pm22.9$  &65.0/-5.0(0)\\
&\multirow{1}{*}{G2} &$99.8\pm47.0$  &136.0/-5.0(0) &$23.0\pm13.3$  &41.0/1.0(0) &$84.2\pm52.2$  &135.0/-5.0(0) &$24.2\pm12.3$  &48.0/4.0(0) &$97.4\pm56.7$  &146.0/-2.0(0)\\
&\multirow{1}{*}{G3} &$\mathbf{111.4}\pm50.9$  & 157.0/11.0(0) &$26.8\pm8.1$  &42.0/13.0(0) &$95.6\pm62.8$  &158.0/10.0(0) &$26.1\pm12.6$  &58.0/7.0(0) &$\mathbf{111.7}\pm48.3$  & 163.0/-5.0(0)\\
&\multirow{1}{*}{DORL} &$110.8\pm61.0$  &157.0/-5.0(0) &$\mathbf{53.4}\pm30.6$  & 97.0/5.0(2) &$\mathbf{105.8}\pm54.0$  & 159.0/-5.0(0) &$\mathbf{59.4}\pm33.6$  & 97.0/-5.0(1) &$92.4\pm73.6$  &189.0/-5.0(0)\\
\midrule

\multirow{4}{*}{PPO}&\multirow{1}{*}{G1} &$48.6\pm49.1$  &147.0/-5.0(0) &$23.7\pm19.9$  &50.0/-5.0(0) &$33.0\pm46.8$  &117.0/-5.0(0) &$31.1\pm15.2$  &59.0/7.0(0) &$7.2\pm19.1$  &61.0/-5.0(0)\\
&\multirow{1}{*}{G2} &$96.5\pm65.2$  &153.0/-5.0(0) &$51.4\pm22.4$  &86.0/-2.0(0) &$51.0\pm62.8$  &152.0/0.0(0) &$42.6\pm18.6$  &73.0/14.0(0) &$61.6\pm60.0$  &160.0/-5.0(0)\\
&\multirow{1}{*}{G3} &$\mathbf{100.7}\pm65.6$  & 157.0/-5.0(0) &$44.0\pm25.9$  &84.0/1.0(0) &$89.0\pm64.5$  &149.0/-5.0(0) &$37.3\pm27.6$  &75.0/-2.0(0) &$46.9\pm63.9$  &164.0/-5.0(0)\\
&\multirow{1}{*}{DORL} &$90.1\pm67.8$  &152.0/-5.0(0) &$\mathbf{57.7}\pm28.0$  & 87.0/7.0(0) &$\mathbf{95.6}\pm66.3$  & 157.0/-5.0(0) &$\mathbf{50.8}\pm26.8$  & 83.0/7.0(0) &$\mathbf{128.4}\pm59.8$  & 173.0/-5.0(0)\\
\midrule 
\multirow{4}{*}{$\text{PPO}_\text{D}$}&\multirow{1}{*}{G1} &$95.0\pm57.2$  &144.0/-5.0(0) &$9.7\pm12.2$  &37.0/-5.0(0) &$54.1\pm58.7$  &135.0/-5.0(0) &$10.1\pm13.2$  &39.0/-5.0(0) &$38.8\pm50.1$  &130.0/-5.0(0)\\
&\multirow{1}{*}{G2} &$\mathbf{110.0}\pm59.3$  & 157.0/-5.0(0) &$45.4\pm25.0$  &88.0/1.0(0) &$\mathbf{102.0}\pm47.2$  & 153.0/-5.0(0) &$\mathbf{54.9}\pm23.9$  & 84.0/11.0(0) &$\mathbf{112.4}\pm68.0$  & 179.0/-1.0(0)\\
&\multirow{1}{*}{G3} &$108.4\pm60.0$  &155.0/-5.0(0) &$33.6\pm22.2$  &78.0/-1.0(0) &$84.7\pm66.4$  &153.0/-5.0(0) &$34.3\pm19.9$  &67.0/-2.0(0) &$77.8\pm70.4$  &172.0/-5.0(0)\\
&\multirow{1}{*}{DORL} &$73.6\pm76.4$  &163.0/-5.0(5) &$\mathbf{51.5}\pm26.5$  & 82.0/2.0(0) &$58.5\pm71.7$  &162.0/-5.0(1) &$50.2\pm27.5$  &79.0/1.0(0) &$91.2\pm81.7$  &182.0/-5.0(0)\\
\midrule
\multirow{4}{*}{A2C}&\multirow{1}{*}{G1} &$-1.6\pm3.5$  &4.0/-5.0(0) &$5.6\pm4.7$  &13.0/-4.0(0) &$-1.2\pm3.6$  &7.0/-5.0(0) &$2.4\pm5.1$  &13.0/-5.0(0) &$0.1\pm4.2$  &9.0/-5.0(0)\\
&\multirow{1}{*}{G2} &$0.2\pm4.0$  &11.0/-5.0(0) &$29.2\pm11.0$  &48.0/2.0(0) &$17.7\pm43.4$  &147.0/0.0(0) &$27.6\pm11.4$  &55.0/5.0(0) &$3.8\pm9.8$  &37.0/-5.0(0)\\
&\multirow{1}{*}{G3} &$54.0\pm42.1$  &152.0/-2.0(0) &$\mathbf{62.7}\pm25.4$  & 81.0/4.0(0) &$\mathbf{23.4}\pm31.4$  & 64.0/-5.0(0) &$\mathbf{54.6}\pm25.4$  & 80.0/7.0(0) &$102.4\pm55.1$  &183.0/28.0(0)\\
&\multirow{1}{*}{DORL} &$\mathbf{149.6}\pm3.6$  & 159.0/145.0(0) &$55.1\pm38.6$  &93.0/-1.0(1) &$-1.0\pm2.0$  &0.0/-5.0(0) &$46.6\pm37.2$  &90.0/-1.0(0) &$\mathbf{167.7}\pm30.9$  & 188.0/70.0(0)\\
\midrule 
\multicolumn{2}{c|}{\multirow{1}{*}{Ye \emph{et al.} \cite{ye2020rotation}}} 
     & $22.8 \pm 19.3 $& $64 / {-}5(0) $    
     &  $1.5 \pm 4.9 $&$12 / {-}5(0)  $
     &  $11.2 \pm 11.4 $&$ 30 / {-}5(0) $
     &  $3.5 \pm 5.0 $&$ 13 / {-}5(0) $
     &  $4.2 \pm 12.4 $ &$ 48 / {-}5 (0)$ \\

\midrule
\multicolumn{2}{c|}{\multirow{1}{*}{Random }}  & $ 3.8 \pm 11.9$  & 42 / -5 (0) & $ 9.4 \pm 12.5$  & 35 / -5 (0) & $ 9.3 \pm 24.4$  & 86 / -5 (0) & $ 7.6 \pm 11.8$  & 38 / -5 (0) & $ 2.0 \pm 16.5$  & 69 / -5 (0)\\
\multicolumn{2}{c|}{\multirow{1}{*}{MCTS }}  & $ 155.1 \pm 8.2$  & 162 / 126 (8) & $ 67.4 \pm 12.2$  & 90 / 36 (0) & $ 129.4 \pm 36.6$  & 161 / 6 (1) & $ 67.7 \pm 11.4$  & 88 / 43 (0) & $ 156.5 \pm 31.1$  & 191 / 75 (1)\\
\multicolumn{2}{c|}{\multirow{1}{*}{OLETS}}  & $\mathbf{164.2} \pm 0.9$  & 166 / 162 (20) & $\mathbf{93.1} \pm 6.2$  & 98 / 75 (9) & $\mathbf{165.0} \pm 1.6$  & 167 / 161 (20) & $\mathbf{86.8} \pm 11.9$  & 98 / 63 (8) & $\mathbf{192.5} \pm 3.5$  & 196 / 184 (13)\\
\multicolumn{2}{c|}{\multirow{1}{*}{RHEA}}   & $ 146.2 \pm 25.4$  & 166 / 89 (4) & $ 61.8 \pm 18.0$  & 87 / 19 (0) & $ 124.3 \pm 51.0$  & 167 / 1 (3) & $ 57.9 \pm 15.3$  & 77 / 8 (0) & $ 156.5 \pm 30.3$  & 181 / 58 (0)\\
\multicolumn{2}{c|}{\multirow{1}{*}{RS}}   & $ 155.0 \pm 14.3$  & 167 / 110 (6) & $ 61.0 \pm 19.8$  & 84 / 22 (0) & $ 142.6 \pm 30.8$  & 166 / 56 (1) & $ 64.3 \pm 14.5$  & 85 / 36 (0) & $ 167.1 \pm 22.2$  & 193 / 97 (0)\\

\bottomrule
\end{tabular}

\label{tab:golddigger_result}
\end{table*}

By comparing the performance of instantiations of G1, G2, G3 and DORL, implemented with PPO, A2C, and DQN, we observe outstanding performance of DORL when playing all the test levels of \emph{GoldDigger} and \emph{TreasureKeeper}.
Highlights are as follows.
\begin{itemize}
\item Any algorithm with dual-observation and one-hot encoding (i.e., agents shown on rows or columns entitled with ``DORL'') obtains significantly higher score than or similar score to its version using other observation or pixel-based encoding (referred to as G1, G2 and G3) when playing all test levels of \emph{GoldDigger} and \emph{TreasureKeeper}. When playing \emph{WaterPuzzle}, using dual-observation is not always helpful.  
\item $\text{DQN}_\text{s}$-DORL and PPO agents perform significantly better than or similar to the agent of Ye \emph{et al.}~\cite{ye2020rotation} in terms of game score and number of wins in all the test levels of \emph{GoldDigger} and \emph{TreasureKeeper}, while this is not always the case for \emph{WaterPuzzle} (cf. Table \ref{tab:ranksum_waterpuzzle_all} in \emph{Appendix}). 
\item Using a stochastic policy for DQN does not always lead to an overall better performed agent. On one hand, $\text{DQN}_\text{s}$ obtains statistically higher or similar game score compared to DQN with the same input observation in 7 out of 9 test levels (i.e., 3 test levels for each game). On the other hand, DQN agent wins \emph{TreasureKeeper} more times than $\text{DQN}_\text{s}$ with the same input observation.
\end{itemize}

More discussions on the agents' performance on individual games are provided as follows.
\subsubsection{GoldDigger with Risky Reward}
According to Table \ref{tab:golddigger_result} and the results of statistical tests provided in \emph{Appendix}, all agents with dual-observation (G2 and DORL) perform significantly better than or similar to those using GO only (G1 and G3) in playing training levels and test levels. Among those four types of input, DORL agents implemented with DQN and PPO usually holds the first place. For example, DQN agent with DORL achieves the highest score among all five levels. The performance of A2C agents is not stable enough due to the high variance of gradients. Any agent with DORL obtains significantly higher or similar game score compared to its version using other observation or pixel-based encoding.

A notable observation of DQN agents is the disparate results of using the deterministic policy and the stochastic policy, especially in \emph{Level-2}. The former obtains $-1.2$ and the latter obtains $105.8$ as an average score. \emph{Level-2} is adapted from \emph{Level-0} by adding a wall (cf. Fig. \ref{fig:mutations}), making the avatar trap at a corner. The stochastic policy helps the avatar to explore the space and escape from the corner. All planning agents perform well in this game, but the random agent plays badly.

Despite the high average scores of DORLs in playing \emph{GoldDigger}, they sometimes get deficient scores which can be determined by the high variance shown in Table \ref{tab:golddigger_result}. PPO agent with DORL obtains 67.8 in \emph{Level-0} and 66.3 in \emph{Level-2} as standard deviations of scores. It may be explained by the fact that the agents sometimes attempt to approach a monster to attack it. However, due to the random behaviours of the monsters, the agent could be killed before attacking. DORLs rarely win since collecting all the jewels is hard. On the other hand, it may be blamed on the drawback of traditional RL algorithms~\cite{garcia2015comprehensive}. RL agents usually pay more attention to getting a higher long-term expected reward, but sometimes ignore risks caused by some uncertainties, e.g., agents bias more to gain score by attacking monsters but ignore the risk of being killed due to the direct use of the game score as the reward. Designing a suitable reward function for efficiently training a general agent is crucial.

\subsubsection{Deceptive TreasureKeeper}
As shown in Table \ref{tab:treasurekeeper_result}, most of the agents can obtain the maximum score ($35$) from \emph{Level-0} to \emph{Level-4}. According to the results of statistical tests provided in \emph{Appendix}, DQN agents with dual-observation (G2 and DORL) obtain significantly higher or similar scores than the ones with the global observation only (G1 and G3) when playing test levels. When using A2C as the base algorithm, the agent using tile-vector global observation only, thus G3, beats all the other agents. In particular, among all the agents, only A2C-G3 can win \emph{Level-1} and \emph{Level-3} with the maximum score at \emph{TreasureKeeper}.

Since \emph{TreasureKeeper} is a survival game, the game score increases along with the survival time of the avatar. Some core actions that make an avatar survive longer (e.g., pushing a box on purpose to block monsters), do not directly lead to any immediate reward. Random decisions can also result in such actions. Nevertheless, the game score occasionally happens to increase when the avatar is pushing a box. Such coincidences easily confuse a learning agent and encourage it to push boxes more frequently, which does not usually help guard the treasure and earn the reward. In such a situation, even if an agent has learnt how to avoid monsters, the treasure might still be attacked by monsters which leads to a failure of the game.
Planning agents can hardly win \emph{Level-1} and its variation \emph{Level-3}.

\subsubsection{Simple Yet Challenging WaterPuzzle}
All PPO agents can achieve acceptable average scores, especially G3 and DORL which get average scores of 15, over 20 trials of playing training levels (cf. Table \ref{tab:waterpuzzle_result}). However, their performance on test levels is disappointing. A2C and DQN agents rarely win the games. No agent can get any score in \emph{Level-3}, and a few agents, including PPO agents with G1/G2, and A2C agents with G2/G3 can get some points in \emph{Level-2} and \emph{Level-4}. \emph{WaterPuzzle} is challenging for planning agents. Most of the planning agents, except OLETS, rarely win the game.

The puzzle game \emph{WaterPuzzle} is extremely easy for human players but hard for RL agents due to the sparse reward. Some agents learned how to win the game at training levels but failed in test levels. It is probably because what agents actually learned is the winning path in a specific map.

\subsection{Further Discussion}\label{sec:furtherdiscussion}

Agents with dual-observation show promising results for playing 2020 competition games. Some remarkable phenomena are discussed as follows.

\subsubsection{Dual-Observation Improves Generalisation}
 Agents enhanced by dual-observation have an advantage over other agents, especially in playing \emph{GoldDigger}. Comparing their performance in playing training levels and unseen test levels, agents with dual-observation usually achieve superior results to others in all three games. Although those unseen levels have different layouts, they share the same game rules. Taking \emph{GoldDigger} as an example, an agent gains reward by approaching and digging diamonds. Agents need to pay more attention to the close elements, such as monsters, diamonds, and treasures. This playing strategy remains the same in training levels and unseen test levels. With the help of dual-observation, knowledge gained in playing training levels enhances the agents' ability to play unseen test levels.

The main benefit brought by using dual-observation is that an agent can receive the whole global information and at the same time collect possibly more important local information around it. Using the global observation alone treats all elements equally, while using the local observation alone is myopic and ignores the information in further area. An agent can consider different areas with different weights when using this dual-observation. In other words, it focuses on the core information that may affect the gained reward since the actions that an agent takes usually have limited effect on the global observation but are more meaningful from local view without losing the reception of remote areas that produce possible future rewards from global view. It can be attributed to the coordination of the global and local observations that the agents with dual-observation usually get better results than others at a new level. Even the layout changes in new levels, there is still some familiar local information in the view of the agents.

\begin{table*}[!ht]
\centering
\setlength{\tabcolsep}{0.5pt}
\caption{\emph{TreasureKeeper}: Average score, standard deviation and number of wins (in the brackets) over 20 independent trials. The bold numbers denotes the highest average scores among four input types or planning agents.}
\begin{tabular}{cc|cccc|cccccc}
\toprule
\multicolumn{2}{c|}{\multirow{2}{*}{Agent}}
& \multicolumn{2}{c}{\emph{Level-0 (35 / 5)}} & \multicolumn{2}{c}{\emph{Level-1 (35 / 5)}} & \multicolumn{2}{c}{\emph{Level-2 (35 / 5)}} & \multicolumn{2}{c}{\emph{Level-3 (35 / 5)}} & \multicolumn{2}{c}{\emph{Level-4 (35 / 5)}} \\ 
&&Avg. $\pm$ Std &Max/Min(Wins)&Avg. $\pm$ Std &Max/Min(Wins)&Avg. $\pm$ Std &Max/Min(Wins)&Avg. $\pm$ Std &Max/Min(Wins)&Avg. $\pm$ Std &Max/Min(Wins)\\ 
\midrule 

\multirow{4}{*}{DQN}&\multirow{1}{*}{G1} &$27.2\pm9.9$  &35.0/10.0(11) &$\mathbf{11.5}\pm10.0$  & 35.0/5.0(2) &$20.2\pm10.8$  &35.0/5.0(5) &$6.0\pm2.0$  &10.0/5.0(0) &$20.2\pm11.1$  &35.0/5.0(5)\\
&\multirow{1}{*}{G2} &$26.0\pm11.1$  &35.0/10.0(12) &$11.2\pm8.0$  &35.0/5.0(1) &$\mathbf{26.0}\pm10.2$  & 35.0/10.0(10) &$\mathbf{13.5}\pm10.4$  & 35.0/5.0(2) &$\mathbf{25.0}\pm10.1$  & 35.0/10.0(8)\\
&\multirow{1}{*}{G3} &$23.5\pm10.1$  &35.0/10.0(7) &$9.2\pm6.4$  &30.0/5.0(0) &$24.0\pm9.3$  &35.0/10.0(6) &$6.2\pm2.7$  &15.0/5.0(0) &$22.8\pm11.0$  &35.0/5.0(8)\\
&\multirow{1}{*}{DORL} &$\mathbf{28.2}\pm9.8$  & 35.0/10.0(13) &$9.2\pm7.1$  &35.0/5.0(1) &$25.2\pm10.9$  &35.0/5.0(9) &$10.8\pm9.1$  &35.0/5.0(2) &$23.2\pm12.5$  &35.0/5.0(10)\\
\midrule 
\multirow{4}{*}{$\text{DQN}_\text{s}$}&\multirow{1}{*}{G1} &$15.5\pm8.2$  &30.0/5.0(0) &$9.8\pm8.7$  &35.0/5.0(1) &$16.2\pm10.1$  &35.0/5.0(3) &$7.8\pm6.8$  &35.0/5.0(1) &$16.5\pm8.8$  &35.0/5.0(3)\\
&\multirow{1}{*}{G2} &$\mathbf{27.0}\pm8.4$  & 35.0/10.0(9) &$12.8\pm8.9$  &35.0/5.0(2) &$24.5\pm9.2$  &35.0/5.0(6) &$10.5\pm6.7$  &25.0/5.0(0) &$19.8\pm11.1$  &35.0/5.0(4)\\
&\multirow{1}{*}{G3} &$22.5\pm10.9$  &35.0/5.0(7) &$10.2\pm9.1$  &35.0/5.0(1) &$20.8\pm12.7$  &35.0/5.0(7) &$7.0\pm4.8$  &25.0/5.0(0) &$13.5\pm6.1$  &25.0/5.0(0)\\
&\multirow{1}{*}{DORL} &$24.0\pm11.9$  &35.0/5.0(9) &$\mathbf{14.5}\pm10.2$  & 35.0/5.0(3) &$\mathbf{25.0}\pm11.4$  & 35.0/5.0(9) &$\mathbf{10.8}\pm8.8$  & 35.0/5.0(1) &$\mathbf{22.5}\pm12.4$  & 35.0/5.0(9)\\
\midrule 

\multirow{5}{*}{PPO}&\multirow{1}{*}{G1} &$16.0\pm11.7$  &35.0/5.0(5) &$\mathbf{14.0}\pm10.6$  & 35.0/5.0(3) &$14.8\pm9.9$  &35.0/5.0(2) &$9.8\pm7.3$  &35.0/5.0(1) &$11.5\pm9.2$  &35.0/5.0(2)\\
&\multirow{1}{*}{G2} &$20.0\pm11.3$  &35.0/5.0(5) &$12.2\pm7.0$  &25.0/5.0(0) &$\mathbf{17.5}\pm10.2$  & 35.0/5.0(3) &$\mathbf{13.2}\pm9.4$  & 35.0/5.0(2) &$14.8\pm8.4$  &35.0/5.0(2)\\
&\multirow{1}{*}{G3} &$\mathbf{21.0}\pm12.0$  & 35.0/5.0(7) &$11.5\pm9.0$  &35.0/5.0(2) &$17.5\pm9.8$  &35.0/5.0(3) &$10.5\pm8.4$  &35.0/5.0(1) &$\mathbf{18.0}\pm11.2$  & 35.0/5.0(5)\\
&\multirow{1}{*}{DORL} &$20.0\pm10.7$  &35.0/5.0(5) &$12.2\pm10.3$  &35.0/5.0(3) &$16.2\pm11.3$  &35.0/5.0(3) &$9.2\pm5.8$  &25.0/5.0(0) &$11.8\pm8.1$  &35.0/5.0(1)\\
\midrule 
\multirow{4}{*}{$\text{PPO}_\text{D}$}&\multirow{1}{*}{G1} &$15.5\pm11.6$  &35.0/5.0(3) &$9.5\pm8.0$  &35.0/5.0(1) &$15.5\pm10.6$  &35.0/5.0(3) &$\mathbf{9.8}\pm8.9$  & 35.0/5.0(2) &$14.0\pm10.7$  &35.0/5.0(3)\\
&\multirow{1}{*}{G2} &$20.2\pm11.5$  &35.0/5.0(5) &$13.8\pm9.9$  &35.0/5.0(1) &$19.5\pm9.1$  &35.0/5.0(3) &$7.8\pm7.2$  &35.0/5.0(1) &$\mathbf{17.5}\pm11.6$  & 35.0/5.0(4)\\
&\multirow{1}{*}{G3} &$12.0\pm7.3$  &35.0/5.0(1) &$\mathbf{17.8}\pm12.3$  & 35.0/5.0(6) &$20.2\pm10.4$  &35.0/5.0(4) &$7.0\pm3.7$  &15.0/5.0(0) &$16.0\pm9.7$  &35.0/5.0(2)\\
&\multirow{1}{*}{DORL} &$\mathbf{22.8}\pm11.2$  & 35.0/5.0(6) &$6.2\pm3.1$  &15.0/5.0(0) &$\mathbf{22.8}\pm12.2$  & 35.0/5.0(9) &$8.8\pm6.9$  &30.0/5.0(0) &$15.5\pm12.1$  &35.0/5.0(4)\\
\midrule 
\multirow{4}{*}{A2C}&\multirow{1}{*}{G1} &$25.2\pm9.3$  &35.0/5.0(6) &$6.0\pm2.0$  &10.0/5.0(0) &$16.8\pm9.3$  &35.0/5.0(2) &$5.8\pm1.8$  &10.0/5.0(0) &$16.8\pm9.5$  &35.0/5.0(2)\\
&\multirow{1}{*}{G2} &$23.8\pm10.5$  &35.0/5.0(8) &$6.8\pm4.0$  &20.0/5.0(0) &$21.8\pm11.4$  &35.0/5.0(7) &$6.0\pm2.0$  &10.0/5.0(0) &$19.0\pm11.6$  &35.0/5.0(6)\\
&\multirow{1}{*}{G3} &$\mathbf{31.2}\pm8.0$  & 35.0/5.0(15) &$\mathbf{15.2}\pm10.8$  & 35.0/5.0(3) &$\mathbf{25.5}\pm11.4$  & 35.0/5.0(11) &$\mathbf{14.2}\pm9.4$  & 35.0/5.0(1) &$\mathbf{27.0}\pm10.3$  & 35.0/5.0(10)\\
&\multirow{1}{*}{DORL} &$26.5\pm10.4$  &35.0/5.0(11) &$6.5\pm2.3$  &10.0/5.0(0) &$23.2\pm10.8$  &35.0/5.0(7) &$6.0\pm2.5$  &15.0/5.0(0) &$20.0\pm9.5$  &35.0/5.0(4)\\
\midrule 
\multicolumn{2}{c|}{\multirow{1}{*}{Ye \emph{et al.} \cite{ye2020rotation}}} 
     & $7.5 \pm 5.1 $& $30 / 5(0) $
     &  $8.2 \pm 5.1 $&$25 / 5(0)  $
     &  $7.3 \pm 3.1 $&$15 / 5(0) $
     &  $7.3 \pm 5.4 $ &$30 / 5 (0)$
     &  $6.8 \pm 3.0 $&$15 / 5(0) $  \\
\midrule
    \multicolumn{2}{c|}{\multirow{1}{*}{Random}}&  $ 10.0 \pm 3.5$  & 20 / 5 (0) & $ 5.3 \pm 1.1$  & 10 / 5 (0) & $ 9.8 \pm 6.8$  & 30 / 5 (0) & $ 5.0 \pm 0.0$  & 5 / 5 (0) & $ 7.3 \pm 4.0$  & 20 / 5 (0)\\
\multicolumn{2}{c|}{\multirow{1}{*}{MCTS}} & $ 32.3 \pm 6.6$  & 35 / 15 (17) & $\mathbf{12.5} \pm 7.5$  & 20 / 5 (0) & $ 33.3 \pm 4.6$  & 35 / 15 (16) & $\mathbf{15.5} \pm 8.4$  & 35 / 5 (1) & $\mathbf{32.8} \pm 2.5$  & 35 / 30 (11)\\
  \multicolumn{2}{c|}{\multirow{1}{*}{OLETS}}& $ 24.3 \pm 9.5$  & 35 / 10 (8) & $ 5.0 \pm 0.0$  & 5 / 5 (0) & $ 24.8 \pm 7.5$  & 35 / 15 (5) & $ 5.0 \pm 0.0$  & 5 / 5 (0) & $ 26.0 \pm 6.6$  & 35 / 10 (4)\\
   \multicolumn{2}{c|}{\multirow{1}{*}{RHEA}}& $ 25.0 \pm 11.1$  & 35 / 10 (10) & $ 5.0 \pm 0.0$  & 5 / 5 (0) & $ 27.0 \pm 9.8$  & 35 / 10 (9) & $ 5.0 \pm 0.0$  & 5 / 5 (0) & $ 25.5 \pm 9.2$  & 35 / 10 (9)\\
  \multicolumn{2}{c|}{\multirow{1}{*}{RS}}& $\mathbf{33.3} \pm 5.3$  & 35 / 15 (18) & $ 11.8 \pm 10.4$  & 30 / 5 (0) & $\mathbf{34.8} \pm 1.1$  & 35 / 30 (19) & $ 10.0 \pm 10.3$  & 35 / 5 (2) & $ 27.0 \pm 10.4$  & 35 / 5 (12)\\

\bottomrule\end{tabular}
\label{tab:treasurekeeper_result}
\end{table*}

\subsubsection{Tile-vector Encoding helps}
Agents using pixel input (G1 and G2) usually perform inferior to those using tile-vector encoded input (cf. Tables \ref{tab:golddigger_result}, \ref{tab:treasurekeeper_result} and \ref{tab:waterpuzzle_result}). Taking \emph{GoldDigger} as an example, PPO agent with G3 beats G1 for all levels. A2C and $\text{DQN}_\text{s}$ meet the same situation. Using tile-vector encoding, each RGB image is converted into $\frac{h}{10} \times \frac{w}{10}\times 16$ one-hot encoded matrices using a pre-calculated tile dictionary, as described in Section \ref{sec:encoding}. It reduces the state space in representation. The compressed state space in representation helps our agents learn better within the same training budget. 

At the same time, we have to admit that one-hot has a restricted scope of use. Since the three games we use here all act with a speed of one tile per game tick, it is easy for us to convert the image to one-hot matrix. Nevertheless, if it is not the case, one-hot matrix may not be converted easily.

\subsubsection{Deterministic vs. Stochastic Decision-making}
Agents with the stochastic policy and the deterministic policy perform very differently, particularly in \emph{GoldDigger} and \emph{WaterPuzzle}. DQN agents with the deterministic policy play badly in test levels, while a much higher score is obtained with the stochastic policy for the same levels. The reason here is that the deterministic policy traps the agent in one state, such as moving forward to a wall which makes no sense. With the stochastic policy, the agent can get away from those states. However, the stochastic policy works worse than the deterministic one in \emph{TreasureKeeper}. Possibly, the stochastic policy makes agents move forward to the monsters mistakenly.

The deterministic policy and stochastic policy for deciding the action to play have their advantages. The stochastic policy can avoid the agent being trapped in one state, but if the penalty due to mistakes is too high, the deterministic one may be a better choice. For example, in \emph{GoldDigger-2}, the agent with the greedy strategy may trap in the wall if it does not choose an ``attack'' action due to a relatively low Q-value. By adding some randomness during action selection, the agent can possibly get rid of this situation.

\begin{table*}[!ht]
\centering
\setlength{\tabcolsep}{0.5pt}

\caption{\emph{WaterPuzzle}: Average score, standard deviation and number of wins (in the brackets) over 20 independent trials. The bold numbers denotes the highest average scores among four input types or planning agents.}
\begin{tabular}{cc|cccc|cccccc}
\toprule
\multicolumn{2}{c|}{\multirow{2}{*}{Agent}}
& \multicolumn{2}{c}{\emph{Level-0 (15 / 0)}} & \multicolumn{2}{c}{\emph{Level-1 (15 / 0)}} & \multicolumn{2}{c}{\emph{Level-2 (15 / 0)}} & \multicolumn{2}{c}{\emph{Level-3 (15 / 0)}} & \multicolumn{2}{c}{\emph{Level-4 (15 / 0)}} \\ 
&&Avg. $\pm$ Std & \ Max/Min(Wins) &\ Avg. $\pm$ Std & \ Max/Min(Wins)&\ Avg. $\pm$ Std &\ Max/Min(Wins)&\ Avg. $\pm$ Std &\ Max/Min(Wins)&\ Avg. $\pm$ Std &\ Max/Min(Wins)\\ 
\midrule 

\multirow{4}{*}{DQN}&\multirow{1}{*}{G1} &$\mathbf{0.0}\pm0.0$  & 0.0/0.0(0) &$0.0\pm0.0$  &0.0/0.0(0) &$\mathbf{0.0}\pm0.0$  & 0.0/0.0(0) &$\mathbf{0.0}\pm0.0$  & 0.0/0.0(0) &$\mathbf{0.0}\pm0.0$  & 0.0/0.0(0)\\
&\multirow{1}{*}{G2} &$0.0\pm0.0$  &0.0/0.0(0) &$0.0\pm0.0$  &0.0/0.0(0) &$0.0\pm0.0$  &0.0/0.0(0) &$0.0\pm0.0$  &0.0/0.0(0) &$0.0\pm0.0$  &0.0/0.0(0)\\
&\multirow{1}{*}{G3} &$0.0\pm0.0$  &0.0/0.0(0) &$0.0\pm0.0$  &0.0/0.0(0) &$0.0\pm0.0$  &0.0/0.0(0) &$0.0\pm0.0$  &0.0/0.0(0) &$0.0\pm0.0$  &0.0/0.0(0)\\
&\multirow{1}{*}{DORL} &$0.0\pm0.0$  &0.0/0.0(0) &$\mathbf{5.0}\pm0.0$  & 5.0/5.0(0) &$0.0\pm0.0$  &0.0/0.0(0) &$0.0\pm0.0$  &0.0/0.0(0) &$0.0\pm0.0$  &0.0/0.0(0)\\
\midrule 
\multirow{4}{*}{$\text{DQN}_\text{s}$}&\multirow{1}{*}{G1} &$0.0\pm0.0$  &0.0/0.0(0) &$2.0\pm2.4$  &5.0/0.0(0) &$0.0\pm0.0$  &0.0/0.0(0) &$\mathbf{0.0}\pm0.0$  & 0.0/0.0(0) &$0.0\pm0.0$  &0.0/0.0(0)\\
&\multirow{1}{*}{G2} &$0.0\pm0.0$  &0.0/0.0(0) &$0.0\pm0.0$  &0.0/0.0(0) &$\mathbf{0.5}\pm1.5$  & 5.0/0.0(0) &$0.0\pm0.0$  &0.0/0.0(0) &$0.0\pm0.0$  &0.0/0.0(0)\\
&\multirow{1}{*}{G3} &$\mathbf{15.0}\pm0.0$  & 15.0/15.0(20) &$4.2\pm1.8$  &5.0/0.0(0) &$0.0\pm0.0$  &0.0/0.0(0) &$0.0\pm0.0$  &0.0/0.0(0) &$\mathbf{1.5}\pm4.5$  & 15.0/0.0(2)\\
&\multirow{1}{*}{DORL} &$7.5\pm4.3$  &15.0/5.0(5) &$\mathbf{5.0}\pm0.0$  & 5.0/5.0(0) &$0.0\pm0.0$  &0.0/0.0(0) &$0.0\pm0.0$  &0.0/0.0(0) &$0.2\pm1.1$  &5.0/0.0(0)\\
\midrule 

\multirow{4}{*}{PPO}&\multirow{1}{*}{G1} &$0.0\pm0.0$  &0.0/0.0(0) &$\mathbf{15.0}\pm0.0$  & 15.0/15.0(20) &$1.5\pm2.3$  &5.0/0.0(0) &$\mathbf{0.0}\pm0.0$  & 0.0/0.0(0) &$0.0\pm0.0$  &0.0/0.0(0)\\
&\multirow{1}{*}{G2} &$4.8\pm1.1$  &5.0/0.0(0) &$15.0\pm0.0$  &15.0/15.0(20) &$\mathbf{8.5}\pm4.8$  & 15.0/5.0(7) &$0.0\pm0.0$  &0.0/0.0(0) &$\mathbf{4.5}\pm1.5$  & 5.0/0.0(0)\\
&\multirow{1}{*}{G3} &$\mathbf{15.0}\pm0.0$  & 15.0/15.0(20) &$15.0\pm0.0$  &15.0/15.0(20) &$0.0\pm0.0$  &0.0/0.0(0) &$0.0\pm0.0$  &0.0/0.0(0) &$0.0\pm0.0$  &0.0/0.0(0)\\
&\multirow{1}{*}{DORL} &$15.0\pm0.0$  &15.0/15.0(20) &$7.5\pm4.3$  &15.0/5.0(5) &$0.0\pm0.0$  &0.0/0.0(0) &$0.0\pm0.0$  &0.0/0.0(0) &$0.0\pm0.0$  &0.0/0.0(0)\\
\midrule 
\multirow{4}{*}{$\text{PPO}_\text{D}$}&\multirow{1}{*}{G1} &$0.0\pm0.0$  &0.0/0.0(0) &$5.0\pm0.0$  &5.0/5.0(0) &$\mathbf{0.0}\pm0.0$  & 0.0/0.0(0) &$\mathbf{0.0}\pm0.0$  & 0.0/0.0(0) &$\mathbf{0.0}\pm0.0$  & 0.0/0.0(0)\\
&\multirow{1}{*}{G2} &$\mathbf{15.0}\pm0.0$  & 15.0/15.0(20) &$12.5\pm4.3$  &15.0/5.0(15) &$0.0\pm0.0$  &0.0/0.0(0) &$0.0\pm0.0$  &0.0/0.0(0) &$0.0\pm0.0$  &0.0/0.0(0)\\
&\multirow{1}{*}{G3} &$15.0\pm0.0$  &15.0/15.0(20) &$\mathbf{15.0}\pm0.0$  & 15.0/15.0(20) &$0.0\pm0.0$  &0.0/0.0(0) &$0.0\pm0.0$  &0.0/0.0(0) &$0.0\pm0.0$  &0.0/0.0(0)\\
&\multirow{1}{*}{DORL} &$15.0\pm0.0$  &15.0/15.0(20) &$15.0\pm0.0$  &15.0/15.0(20) &$0.0\pm0.0$  &0.0/0.0(0) &$0.0\pm0.0$  &0.0/0.0(0) &$0.0\pm0.0$  &0.0/0.0(0)\\
\midrule
\multirow{4}{*}{A2C}&\multirow{1}{*}{G1} &$\mathbf{0.0}\pm0.0$  & 0.0/0.0(0) &$\mathbf{0.0}\pm0.0$  & 0.0/0.0(0) &$0.0\pm0.0$  &0.0/0.0(0) &$\mathbf{0.0}\pm0.0$  & 0.0/0.0(0) &$0.0\pm0.0$  &0.0/0.0(0)\\
&\multirow{1}{*}{G2} &$0.0\pm0.0$  &0.0/0.0(0) &$0.0\pm0.0$  &0.0/0.0(0) &$\mathbf{5.0}\pm0.0$  & 5.0/5.0(0) &$0.0\pm0.0$  &0.0/0.0(0) &$4.8\pm1.1$  &5.0/0.0(0)\\
&\multirow{1}{*}{G3} &$0.0\pm0.0$  &0.0/0.0(0) &$0.0\pm0.0$  &0.0/0.0(0) &$5.0\pm0.0$  &5.0/5.0(0) &$0.0\pm0.0$  &0.0/0.0(0) &$\mathbf{5.0}\pm0.0$  & 5.0/5.0(0)\\
&\multirow{1}{*}{DORL} &$0.0\pm0.0$  &0.0/0.0(0) &$0.0\pm0.0$  &0.0/0.0(0) &$0.0\pm0.0$  &0.0/0.0(0) &$0.0\pm0.0$  &0.0/0.0(0) &$0.0\pm0.0$  &0.0/0.0(0)\\
\midrule 
\multicolumn{2}{c|}{\multirow{1}{*}{Ye \emph{et al.} \cite{ye2020rotation}}} 
    & $1.0 \pm 3.0 $& $15 / 0(1)  $
    &  $0.7 \pm 1.7 $&$5 / 0(0) $
    &  $2.7 \pm 3.3 $&$15 / 0 (1)$
    &  $0.0 \pm 0.0 $&$0 / 0 (0) $
    &  $1.3 \pm 2.2 $ &$ 5 / 0(0) $ \\
\midrule
 \multicolumn{2}{c|}{\multirow{1}{*}{Random}}  & $ 4.8 \pm 6.2$  & 15 / 0 (5) & $ 1.0 \pm 3.4$  & 15 / 0 (1) & $ 2.8 \pm 3.7$  & 15 / 0 (1) & $ 0.3 \pm 1.1$  & 5 / 0 (0) & $ 3.0 \pm 4.6$  & 15 / 0 (2)\\
 \multicolumn{2}{c|}{\multirow{1}{*}{MCTS}} & $ 9.0 \pm 7.4$  & 15 / 0 (12) & $ 5.3 \pm 6.0$  & 15 / 0 (5) & $ 6.5 \pm 5.3$  & 15 / 0 (5) & $ 1.8 \pm 3.6$  & 15 / 0 (1) & $ 3.5 \pm 2.3$  & 5 / 0 (0)\\
 \multicolumn{2}{c|}{\multirow{1}{*}{OLETS}} & $\mathbf{15.0} \pm 0.0$  & 15 / 15 (20) & $\mathbf{11.3} \pm 5.2$  & 15 / 0 (13) & $\mathbf{15.0} \pm 0.0$  & 15 / 15 (20) & $\mathbf{9.8} \pm 5.4$  & 15 / 0 (10) & $\mathbf{15.0} \pm 0.0$  & 15 / 15 (20)\\
 \multicolumn{2}{c|}{\multirow{1}{*}{RHEA}} & $ 7.0 \pm 7.3$  & 15 / 0 (9) & $ 2.8 \pm 4.6$  & 15 / 0 (2) & $ 7.8 \pm 6.2$  & 15 / 0 (8) & $ 1.0 \pm 2.0$  & 5 / 0 (0) & $ 3.5 \pm 4.5$  & 15 / 0 (2)\\
 \multicolumn{2}{c|}{\multirow{1}{*}{RS}}  & $ 9.3 \pm 7.1$  & 15 / 0 (12) & $ 2.5 \pm 2.5$  & 5 / 0 (0) & $ 5.5 \pm 5.9$  & 15 / 0 (5) & $ 0.3 \pm 1.1$  & 5 / 0 (0) & $ 2.0 \pm 2.5$  & 5 / 0 (0)\\

\bottomrule\end{tabular}

\label{tab:waterpuzzle_result}
\end{table*}
\begin{table*}[!htbp]
\caption{\label{tab:competitionres2021}Comparing DORL with planning agents on the 2021 competition games. Average score and standard deviation over 20 independent runs on testing and training levels are reported, as well as the number of wins shown in the brackets. The bold numbers denotes the highest average scores among planning agents.}
\subfloat[\emph{GreedyMouse}\label{tab:greedymouse2021_table}]{% 
\resizebox{\textwidth}{!}{%
\setlength{\tabcolsep}{0.5pt}
\begin{tabular}{cc|cccc|cccccc}
\toprule
\multicolumn{2}{c|}{\multirow{2}{*}{Agent}}
&\multicolumn{2}{c}{\emph{Level-0 (98 / -40)}}& \multicolumn{2}{c}{\emph{Level-1 (67 / -60)}} & \multicolumn{2}{c}{\emph{Level-2 (98 / -50)}} & \multicolumn{2}{c}{\emph{Level-3 (67 / -60)}} & \multicolumn{2}{c}{\emph{Level-4 (77 / -70)}} \\ 
&&Avg. $\pm$ Std &\ Max/Min(Wins)&\ Avg. $\pm$ Std &\ Max/Min(Wins)&Avg. $\pm$ Std &\ Max/Min(Wins)&\ Avg. $\pm$ Std &\ Max/Min(Wins)&\ Avg. $\pm$ Std &\ Max/Min(Wins)\\ 
\midrule 
 \multicolumn{2}{c|}{\multirow{1}{*}{\emph{Arcane}}} &$6.0\pm2.6$  & 8.0/1.0(0) &$17.1\pm6.0$  & 33.0/9.0(0) &$-0.6\pm8.9$  & 35.0/-10.0(0) &$-0.8\pm4.5$  & 16.0/-5.0(0) &$4.7\pm12.6$  & 40.0/-5.0(0)\\
\midrule
 \multicolumn{2}{c|}{\multirow{1}{*}{Random}} & $ 0.8 \pm 1.8$  & 5 / 0 (0) & $ 0.0 \pm 0.0$  & 0 / 0 (0) & $ 2.8 \pm 2.5$  & 5 / 0 (0) & $ 0.0 \pm 0.0$  & 0 / 0 (0) & $ 0.3 \pm 1.1$  & 5 / 0 (0)\\
  \multicolumn{2}{c|}{\multirow{1}{*}{MCTS}} & $ 2.3 \pm 2.5$  & 5 / 0 (0) & $ 0.0 \pm 0.0$  & 0 / 0 (0) & $ 7.8 \pm 4.9$  & 15 / 0 (6) & $ 0.0 \pm 0.0$  & 0 / 0 (0) & $ 0.5 \pm 1.5$  & 5 / 0 (0)\\
 \multicolumn{2}{c|}{\multirow{1}{*}{OLETS}} & $\mathbf{12.3} \pm 5.6$  & 15 / 0 (16) & $\mathbf{11.0} \pm 6.2$  & 15 / 0 (14) & $\mathbf{11.3} \pm 5.9$  & 15 / 0 (14) & $\mathbf{10.0} \pm 6.9$  & 15 / 0 (13) & $\mathbf{9.5} \pm 5.0$  & 15 / 5 (9)\\
 \multicolumn{2}{c|}{\multirow{1}{*}{RHEA}} & $ 2.0 \pm 2.5$  & 5 / 0 (0) & $ 0.0 \pm 0.0$  & 0 / 0 (0) & $ 5.3 \pm 4.6$  & 15 / 0 (3) & $ 0.0 \pm 0.0$  & 0 / 0 (0) & $ 0.0 \pm 0.0$  & 0 / 0 (0)\\
 \multicolumn{2}{c|}{\multirow{1}{*}{RS}} & $ 2.3 \pm 2.5$  & 5 / 0 (0) & $ 0.0 \pm 0.0$  & 0 / 0 (0) & $ 6.8 \pm 4.3$  & 15 / 0 (4) & $ 0.0 \pm 0.0$  & 0 / 0 (0) & $ 0.5 \pm 1.5$  & 5 / 0 (0)\\
\bottomrule
\end{tabular}%
}}%
\vspace{2mm}

\subfloat[\emph{BraveKeeper}\label{tab:bravekeeper2021_table}]{% 
\resizebox{\textwidth}{!}{%
\setlength{\tabcolsep}{0.5pt}
\begin{tabular}{cc|cccc|cccccc}
\toprule
\multicolumn{2}{c|}{\multirow{2}{*}{Agent}}
&\multicolumn{2}{c}{\emph{Level-0 (100 / -40)}}& \multicolumn{2}{c}{\emph{Level-1 (90 / -30)}} & \multicolumn{2}{c}{\emph{Level-2 (100 / -40)}} & \multicolumn{2}{c}{\emph{Level-3 (90 / -30)}} & \multicolumn{2}{c}{\emph{Level-4 (90 / -30)}} \\ 
&&Avg. $\pm$ Std &\ Max/Min(Wins)&\ Avg. $\pm$ Std &\ Max/Min(Wins)&\ Avg. $\pm$ Std &\ Max/Min(Wins)&\ Avg. $\pm$\ Std &\ Max/Min(Wins)&\ Avg. $\pm$ Std &\ Max/Min(Wins)\\ 
\midrule 
 \multicolumn{2}{c|}{\multirow{1}{*}{\emph{Arcane}}} &$21.8\pm17.2$  & 50.0/0.0(4) &$32.0\pm27.36$  & 75.0/5.0(3) &$12.3\pm9.4$  & 35.0/0.0(1) &$41.3\pm26.4$  & 75.0/5.0(3) &$46.0\pm23.9$  & 75.0/5.0(12)\\
\midrule 
%\midrule 
\multicolumn{2}{c|}{\multirow{1}{*}{Random}}  & $ 7.0 \pm 5.6$  & 20 / 0 (0) & $ 5.0 \pm 0.0$  & 5 / 5 (0) & $ 8.5 \pm 6.9$  & 30 / 0 (0) & $ 5.3 \pm 1.1$  & 10 / 5 (0) & $ 33.5 \pm 16.7$  & 55 / 10 (7)\\
 \multicolumn{2}{c|}{\multirow{1}{*}{MCTS}}  & $ 49.0 \pm 4.4$  & 55 / 40 (20) & $\mathbf{64.5} \pm 10.7$  & 85 / 45 (20) & $\mathbf{48.5} \pm 3.6$  & 55 / 45 (20) & $ 36.3 \pm 20.1$  & 70 / 15 (11) & $ 62.0 \pm 4.0$  & 70 / 55 (20)\\
 \multicolumn{2}{c|}{\multirow{1}{*}{OLETS}} & $\mathbf{56.5} \pm 5.3$  & 65 / 45 (20) & $ 59.8 \pm 9.7$  & 80 / 45 (20) & $ 46.3 \pm 8.9$  & 65 / 35 (20) & $\mathbf{49.3} \pm 8.8$  & 65 / 40 (20) & $\mathbf{66.3} \pm 6.5$  & 80 / 55 (20)\\
 \multicolumn{2}{c|}{\multirow{1}{*}{RHEA}} & $ 34.0 \pm 10.6$  & 55 / 0 (17) & $ 33.5 \pm 17.5$  & 55 / 5 (10) & $ 43.5 \pm 5.7$  & 55 / 30 (19) & $ 27.5 \pm 18.8$  & 55 / 5 (7) & $ 56.8 \pm 12.3$  & 75 / 10 (18)\\
 \multicolumn{2}{c|}{\multirow{1}{*}{RS}} & $ 44.5 \pm 5.9$  & 60 / 35 (20) & $ 28.5 \pm 15.7$  & 55 / 5 (9) & $ 33.0 \pm 14.7$  & 45 / -5 (15) & $ 25.5 \pm 22.3$  & 65 / 5 (6) & $ 61.0 \pm 7.0$  & 75 / 50 (18)\\
\bottomrule
\end{tabular}%
}}%
\vspace{2mm}

\subfloat[\emph{TrappedHero}\label{tab:trappedhero2021_table}]{% 
\resizebox{\textwidth}{!}{%
\setlength{\tabcolsep}{0.5pt}
\begin{tabular}{cc|cccc|cccccc}
\toprule
\multicolumn{2}{c|}{\multirow{2}{*}{Agent}}
&\multicolumn{2}{c}{\emph{Level-0 (15 / 0)}}& \multicolumn{2}{c}{\emph{Level-1 (15 / 0)}} & \multicolumn{2}{c}{\emph{Level-2 (15 / 0)}} & \multicolumn{2}{c}{\emph{Level-3 (15 / 0)}} & \multicolumn{2}{c}{\emph{Level-4 (15 / 0)}} \\ 
&&Avg. $\pm$ Std &\ Max/Min(Wins)&\ Avg. $\pm$ Std &\ Max/Min(Wins)&\ Avg. $\pm$ Std &\ Max/Min(Wins)&\ Avg. $\pm$ Std &\ Max/Min(Wins)&\ Avg. $\pm$ Std &\ Max/Min(Wins)\\ 
\midrule 
 \multicolumn{2}{c|}{\multirow{1}{*}{\emph{Arcane}}} &$15.0\pm0.0$  & 15.0/15.0(20) &$0.0\pm0.0$  & 0.0/0.0(0) &$0.0\pm0.0$  & 0.0/0.0(0) &$0.0\pm0.0$  & 0.0/0.0(0) &$0.0\pm0.0$  & 0.0/0.0(0)\\
\midrule 
%\midrule 
 \multicolumn{2}{c|}{\multirow{1}{*}{Random}} & $ 0.8 \pm 1.8$  & 5 / 0 (0) & $ 0.0 \pm 0.0$  & 0 / 0 (0) & $ 2.8 \pm 2.5$  & 5 / 0 (0) & $ 0.0 \pm 0.0$  & 0 / 0 (0) & $ 0.3 \pm 1.1$  & 5 / 0 (0)\\
 \multicolumn{2}{c|}{\multirow{1}{*}{MCTS}} & $ 2.3 \pm 2.5$  & 5 / 0 (0) & $ 0.0 \pm 0.0$  & 0 / 0 (0) & $ 7.8 \pm 4.9$  & 15 / 0 (6) & $ 0.0 \pm 0.0$  & 0 / 0 (0) & $ 0.5 \pm 1.5$  & 5 / 0 (0)\\
 \multicolumn{2}{c|}{\multirow{1}{*}{OLETS}} & $\mathbf{12.3} \pm 5.6$  & 15 / 0 (16) & $\mathbf{11.0} \pm 6.2$  & 15 / 0 (14) & $\mathbf{11.3} \pm 5.9$  & 15 / 0 (14) & $\mathbf{10.0} \pm 6.9$  & 15 / 0 (13) & $\mathbf{9.5} \pm 5.0$  & 15 / 5 (9)\\
 \multicolumn{2}{c|}{\multirow{1}{*}{RHEA}} & $ 2.0 \pm 2.5$  & 5 / 0 (0) & $ 0.0 \pm 0.0$  & 0 / 0 (0) & $ 5.3 \pm 4.6$  & 15 / 0 (3) & $ 0.0 \pm 0.0$  & 0 / 0 (0) & $ 0.0 \pm 0.0$  & 0 / 0 (0)\\
 \multicolumn{2}{c|}{\multirow{1}{*}{RS}} & $ 2.3 \pm 2.5$  & 5 / 0 (0) & $ 0.0 \pm 0.0$  & 0 / 0 (0) & $ 6.8 \pm 4.3$  & 15 / 0 (4) & $ 0.0 \pm 0.0$  & 0 / 0 (0) & $ 0.5 \pm 1.5$  & 5 / 0 (0)\\
\bottomrule
\end{tabular}%
}}
\end{table*}
\subsubsection{Training with random starts}\label{sec:da}
To improve the generalisation ability of learning agents, we apply a simple data addition technique to PPO agent, denoted as $\text{PPO}_{\text{D}}$. We keep the layout of training level maps unchanged during training, but a new initial position is given to the avatar when launching a new episode. Thus, the agent keeps playing the same training level maps but with different initial positions.

According to Tables \ref{tab:golddigger_result}, \ref{tab:treasurekeeper_result} and \ref{tab:waterpuzzle_result}, generating more levels using this simple technique for training a PPO agent does not lead to a better performed agent. $\text{PPO}_\text{D}$ obtains statistically higher game score than PPO with the same input observation in playing two test levels of three games and statistically lower score than PPO with the same input observation in playing five test levels of three games.
PPO-DORL performs worse with the aforementioned random start technique, but dual-observation still helps. When playing test levels of \emph{GoldDigger}, $\text{PPO}_{\text{D}}$ with dual-observation still obtains better results than using global observation only, comparing G2 with G1, and DORL with G3. 

The $\text{PPO}_{\text{D}}$ agents show stable performance in playing training levels, especially for \emph{WaterPuzzle}, where they can get an average reward of 15. But those agents can not get any points in playing test levels.

\subsection{\emph{Arcane}: PPO-DORL Playing 2021 Competition Games}\label{sec:games2021}
PPO-DORL is used as a baseline agent in the 2021 competition edition, named as \emph{Arcane} and trained on the competition games, \emph{GreedyMouse}, \emph{BraveKeeper} and \emph{TrappedHero}, as described in Section \ref{sec:setting}. As no competition submission was received, \emph{Arcane} is also compared to planning agents in Table \ref{tab:competitionres2021}.

\emph{Arcane} achieves the higher max score in 4 levels of \emph{GreedyMouse} compared with OLETS and beats other planning agents regarding average score in \emph{Level-0}, \emph{Level-1}, and \emph{Level-4}. When playing \emph{BraveKeeper}, \emph{Arcane} is not stable enough, although its max scores are close to OLETS' scores in \emph{Level-1}, \emph{Level-3}, and \emph{Level-4}. \emph{Arcane} achieves a win rate of 100\% in playing \emph{TrappedHero-0}, higher than all the planning agents. However, it failed to earn any score in the other levels.

\section{Conclusion}\label{sec:conclusion}
In this paper, we first summarise the five years' GVGAI Learning Competition editions, in particular, the latest ones in 2019, 2020 and 2021. Viewing the poor performance of competition entries, we design a novel reinforcement learning technique with dual-observation (DORL) for general video game playing.

DORL simultaneously takes the tile-vector encoded, transformed global observation and local observation of the game screen as input, aiming at learning local information which may exist in unseen games or levels during training. As a general technique, DORL is applied to DQN, A2C, and PPO separately and compared in playing the games of the 2020 GVGAI Learning Competition. Two versions of DQN-DORL with the stochastic policy and deterministic policy for deciding actions to play during test, respectively, are performed. DORL shows its superior performance on the game set of the 2020 GVGAI Learning Competition, thanks to the introduction of local observation. Additionally, the training data set is enlarged with new levels generated by hanging the avatar's initial position. We observe that applying this simple data addition technique is not always helpful. DORL implemented with PPO (named as \emph{Arcane}), is used as a baseline agent in the 2021 competition.

In this paper, for each game, an agent is trained separately as a base agent. A meta-agent is then composed of those trained base agents without knowledge sharing for playing different games. As future work, we will investigate knowledge transfer across games.

%ack

\section*{Acknowledgement}
The authors would like to thank the authors of \cite{ye2020rotation} for helping with correctly executing their agent, the whole GVGAI team for implementing and maintaining the framework, Mr. Yang Tao for designing the 2020 competition games, and all the participants to the GVGAI Learning Competition editions for their participation.

%\section*{Author Contribution}
%Design of approach and agents: C. Hu, Z. Wang, T. Shu and J. Liu contributed to the design of agents; C. Hu and Z. Wang also contributed to all the experimental studies reported in this paper; H. Tong maintained the 2019, 2020 and 2021 GVGAI Learning Competition editions and tested the competition entries; all authors reviewed the final manuscript.

% \bibliographystyle{IEEEtran}
% \bibliography{main}

\newpage
\begin{IEEEbiography}
    [{\includegraphics[width=1in,height=1.25in,clip,keepaspectratio]{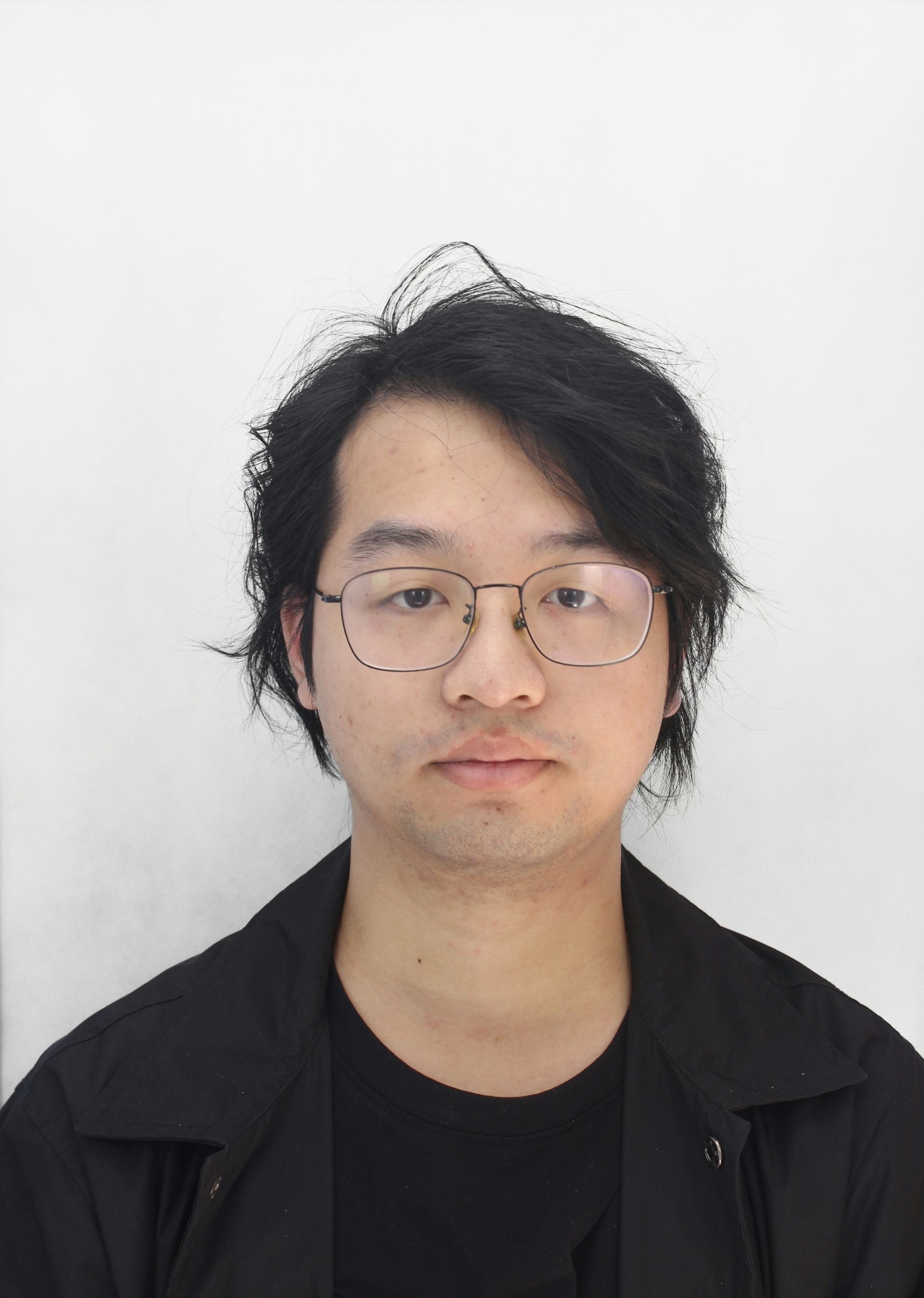}}]{Chengpeng Hu} received his B.E. degree in 2021 from the Southern University of Science and Technology (SUSTech), China. He is currently a postgraduate
student at the Department of Computer Science and Engineering of SUSTech. His research interests
include  AI in games, evolutionary computation and its applications in combinatorial optimisation. 
\end{IEEEbiography}

\begin{IEEEbiography}
    [{\includegraphics[width=1in,height=1.25in,clip,keepaspectratio]{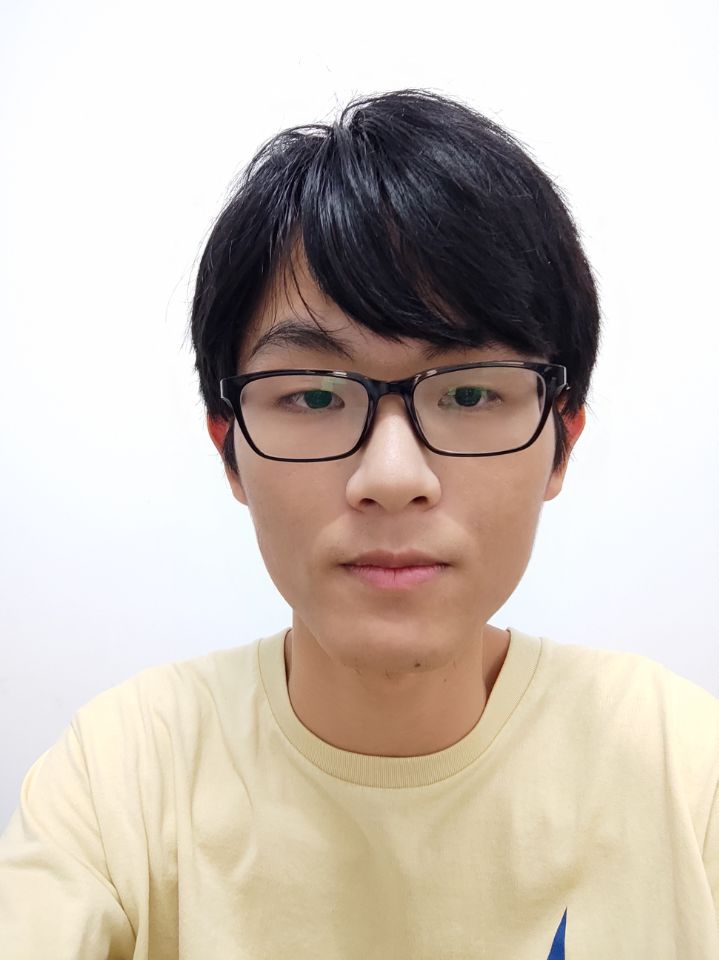}}]{Ziqi Wang} received his B.E. degree in 2021 from the Southern University of Science and Technology (SUSTech), China. He is currently a postgraduate student at the Department of Computer Science and Engineering of SUSTech. His research concentrates on procedural content generation through AI methods including deep reinforcement learning, generative adversarial networks and evolutionary computation. He has co-authored several papers on procedural content generation and has previously published with the IEEE Conference on Games and the IEEE Congress on Evolutionary Computation.
\end{IEEEbiography}

\begin{IEEEbiography}
    [{\includegraphics[width=1in,height=1.25in,clip,keepaspectratio]{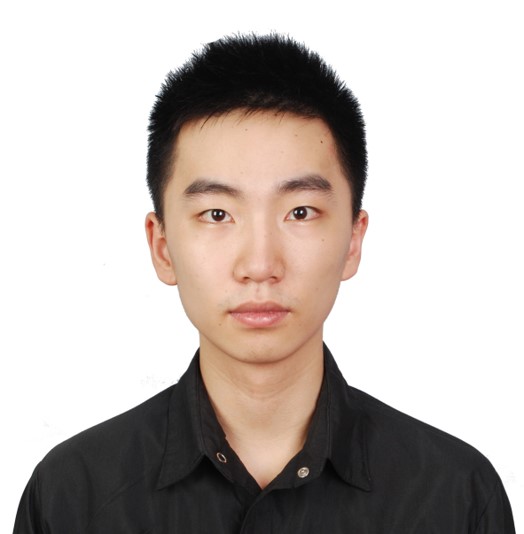}}]{Tianye Shu} received his B.E. degree in 2021 from the Southern University of Science and Technology (SUSTech), China. He is currently a postgraduate student at the Department of Computer Science and Engineering of SUSTech. His research interests include AI in games and evolutionary multi-objective optimisation. His recent paper on ``Experience-driven PCG via reinforcement learning: A Super Mario Bros study'' has been published at the 2022 IEEE Conference on Games.
\end{IEEEbiography}

\begin{IEEEbiography}
    [{\includegraphics[width=1in,height=1.25in,clip,keepaspectratio]{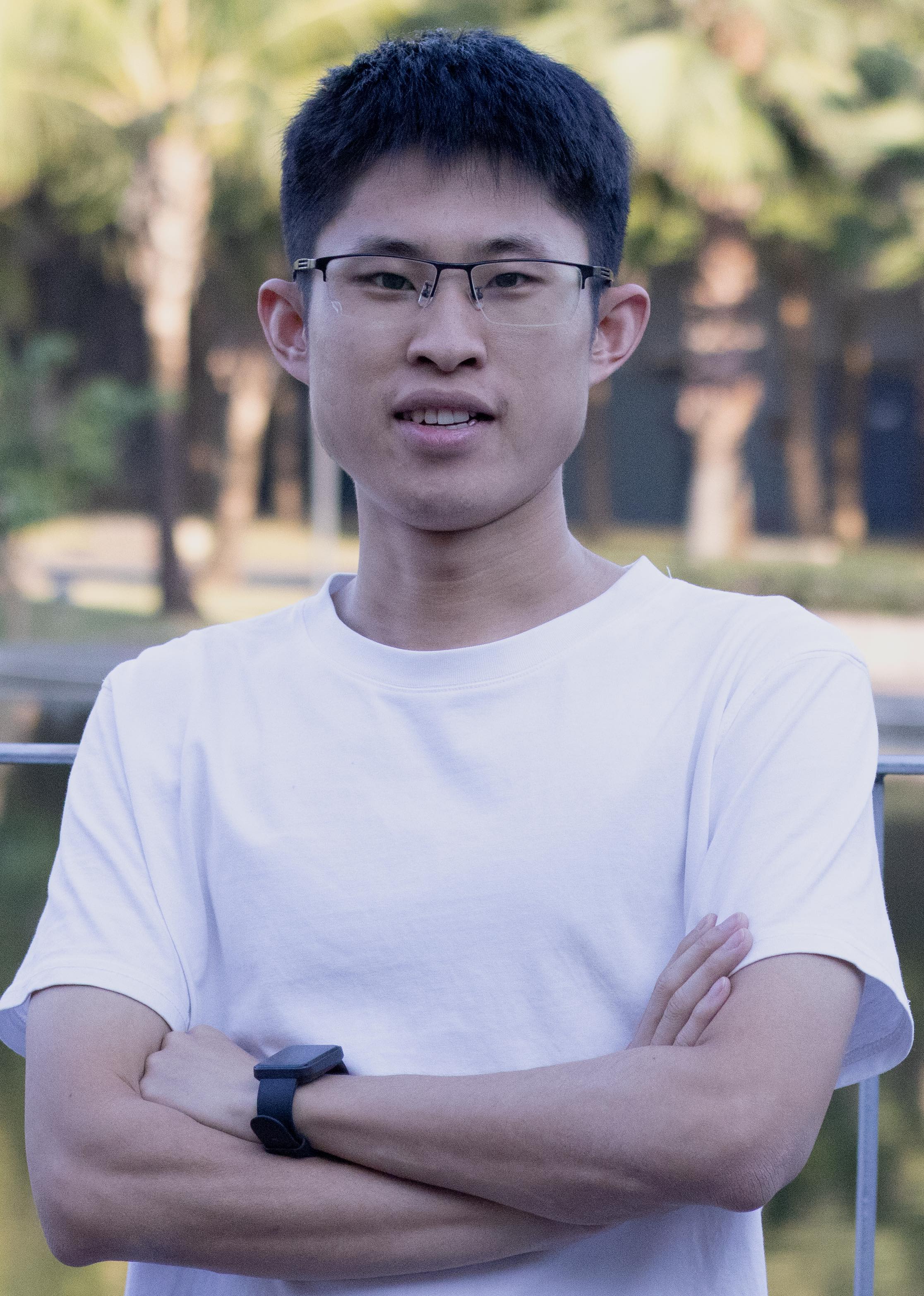}}]{Hao Tong} received his MA.Sc degree from the Joint Program of Southern University of science and Technology and Harbin Institute of Technology in 2019 and his B.E. degree from the China University of Mining and Technology in 2017. Now, he is a PhD student in University of Birmingham. His major research interests include evolutionary computation in vehicle routing problems and dynamic optimisation. He has previously published with CEC, GECCO, PPSN, TEVC and etc. In particular, he was also very interested in AI in games. He has organised a series of competitions in CoG, GECCO, PPSN about General Video Game AI. He was a member of the Games Technical Committee of the IEEE Computational Intelligence Society (2020-2021). 
\end{IEEEbiography}

\begin{IEEEbiography}
    [{\includegraphics[width=1in,height=1.25in,clip,keepaspectratio]{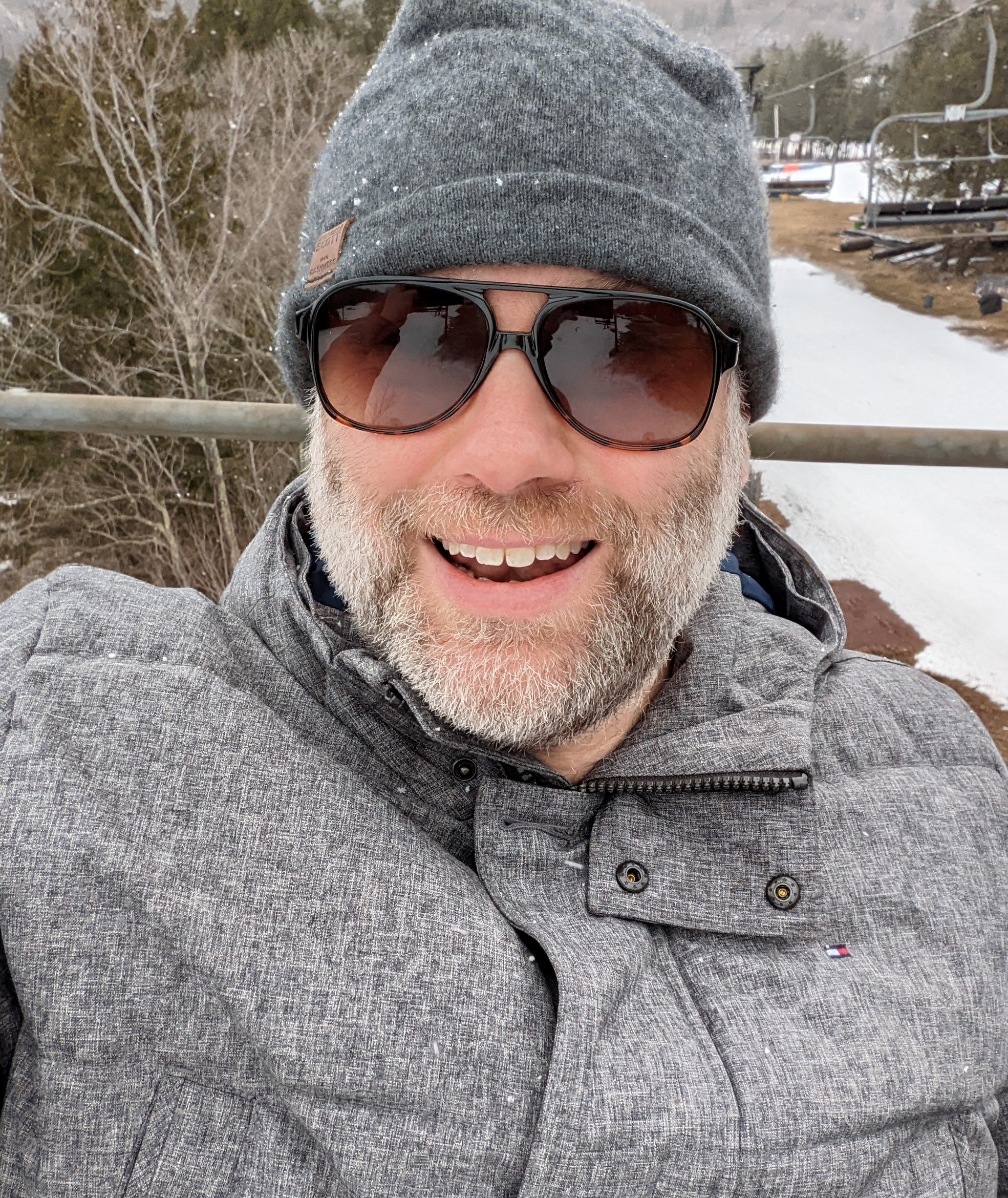}}]{Julian Togelius}  is an Associate Professor in the Department of Computer Science and Engineering, New York University, and a co-founder of \emph{modl.ai}. He works on artificial intelligence for games and on games for artificial intelligence. His current main research directions involve procedural content generation in games, general video game playing, player modelling, and fair and relevant benchmarking of AI through game-based competitions. Additionally, he works on topics in evolutionary computation, quality-diversity algorithms, and reinforcement learning. From 2018 to 2021, he was the Editor-in-Chief of the IEEE Transactions on Games. Togelius holds a BA from Lund University, an MSc from the University of Sussex, and a PhD from the University of Essex. He has previously worked at IDSIA in Lugano and at the IT University of Copenhagen.
\end{IEEEbiography}

\begin{IEEEbiography}[{\includegraphics[width=1in,height=1.25in,clip,keepaspectratio]{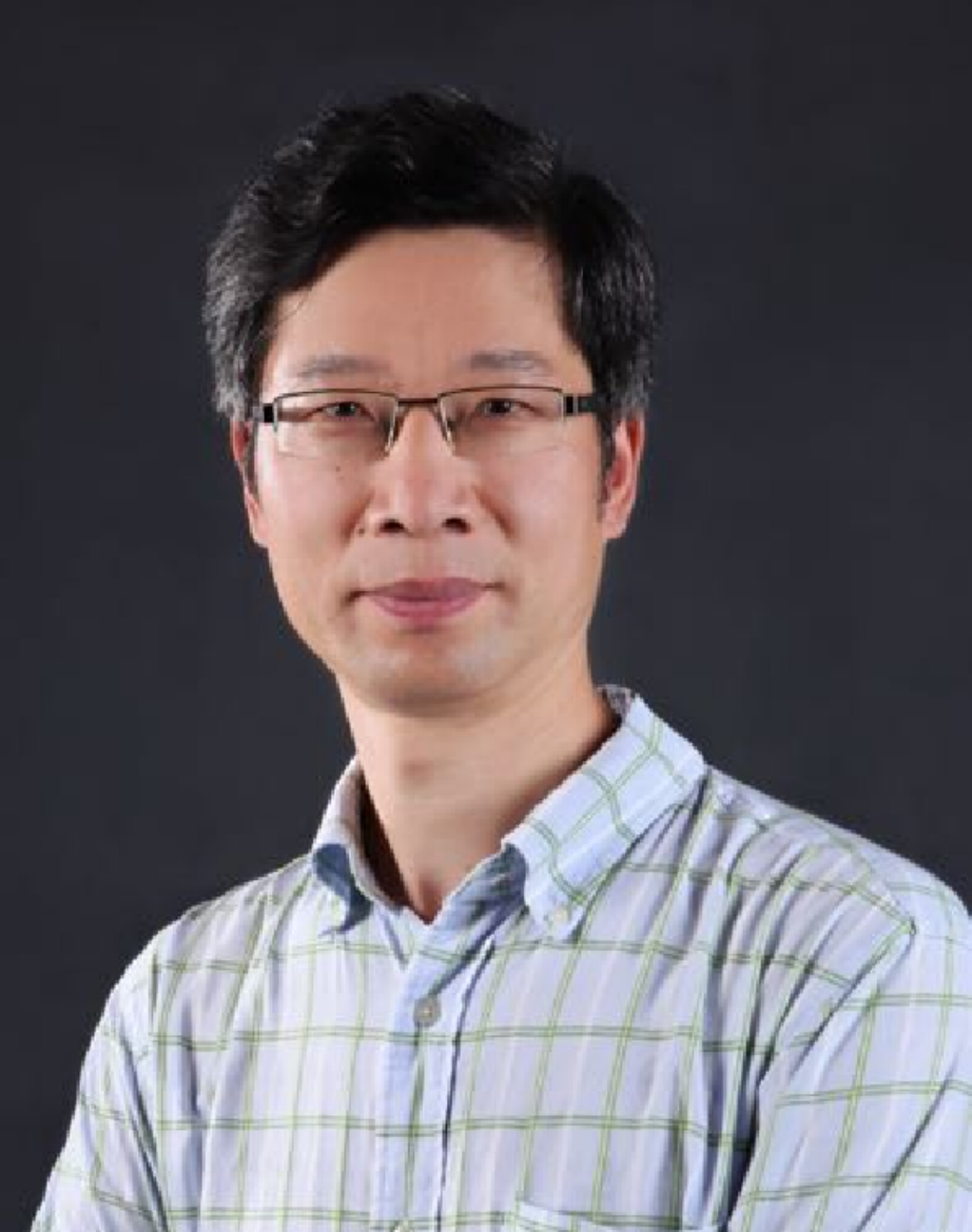}}]{Xin Yao} \textbf{(M'91--SM'96--F'03)} obtained his Ph.D. in 1990 from the University of Science and Technology of China (USTC) in Hefei, MSc in 1985 from North China Institute of Computing Technologies, Beijing, and BSc in 1982 from USTC. He is currently a Chair Professor and founding head of Department of Computer Science and Engineering at the Southern University of Science and Technology (SUSTech), Shenzhen, China, and a part-time Professor of Computer Science at the University of Birmingham, UK. He is an IEEE Fellow and was a Distinguished Lecturer of the IEEE Computational Intelligence Society (CIS). He served as the President (2014-15) of IEEE CIS and the Editor-in-Chief (2003-08) of IEEE Transactions on Evolutionary Computation. His major research interests include evolutionary computation and ensemble machine learning. In particular, he is very interested in iterated prisoner's dilemma games and co-evolutionary learning of game-playing strategies. His 2008 paper on ``Measuring Generalization Performance in Co-evolutionary Learning'' won the 2010 IEEE Transactions on Evolutionary Computation Outstanding Paper Award. He also won the 2001 IEEE Donald G. Fink Prize Paper Award; 2016 and 2017 IEEE Transactions on Evolutionary Computation Outstanding Paper Awards; 2011 IEEE Transactions on Neural Networks Outstanding Paper Award; 2010 BT Gordon Radley Award for Best Author of Innovation (Finalist); and several best paper awards at conferences. He received a 2012 Royal Society Wolfson Research Merit Award, the 2013 IEEE CIS Evolutionary Computation Pioneer Award and the 2020 IEEE Frank Rosenblatt Award.
\end{IEEEbiography}

\begin{IEEEbiography}[{\includegraphics[width=1in,height=1.25in,clip,keepaspectratio]{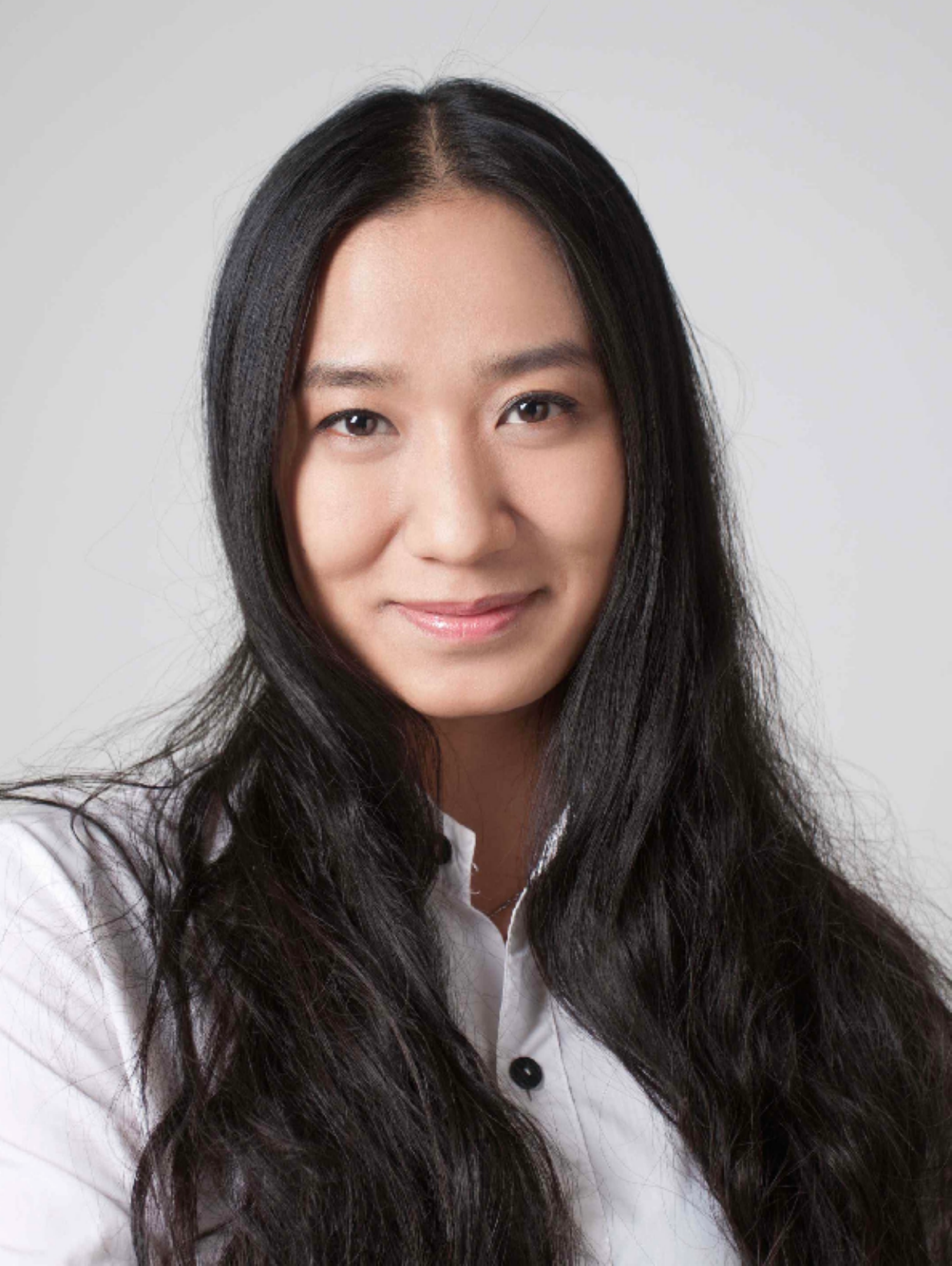}}]{Jialin Liu}
\textbf{(M'16--SM'20)}
received her Ph.D. in 2016 from Universit\'e Paris-Saclay, MSc in 2013 from the \'{E}cole Polytechnique \& Universit\'{e} Paris-Sud, France, Dipl\^ome d'Ing\'enieur in 2012 from the Polytech'Paris-Sud, France, and BSc in 2010 from the Huazhong University of Science and Technology (HUST), China. Currently, she is a Tenure-Track Assistant Professor at the Department of Computer Science and Engineering, Southern University of Science and Technology (SUSTech), China. Her research interests include AI in games, evolutionary computation, portfolio methods, and their applications to planning, scheduling and learning in uncertain environments. She is an Associate Editor of the IEEE Transactions on Games and was the Chair of the Games Technical Committee of the IEEE Computational Intelligence Society (2020-2021).
\end{IEEEbiography}

\newpage
\onecolumn
\appendix

Tables \ref{tab:ranksum_golddigger_all}, \ref{tab:ranksum_treasurekeeper_all} and \ref{tab:ranksum_waterpuzzle_all} provide the results of Wilcoxon rank-sum test performed on the game scores reported in Tables \ref{tab:golddigger_result}, \ref{tab:treasurekeeper_result} and \ref{tab:waterpuzzle_result} for comparing agents' performance in playing test levels of \emph{GoldDigger}, \emph{TreasureKeeper} and \emph{WaterPuzzle}. ``+/$\approx$/-'' indicates the number of times that the row agent obtains statistically higher/similar/lower game score compared to the column agent, respectively. The numbers of wins of the row agent and the column agent are repeated in the brackets.
Highlights are as follows.
\begin{itemize}
\item 
Any algorithm with dual-observation and one-hot encoding (i.e., agents shown on rows or columns entitled with ``DORL'') obtains significantly higher score than or similar score to its version using other observation or pixel-based encoding (referred to as G1, G2 and G3) when playing all test levels of \emph{GoldDigger} and \emph{TreasureKeeper}. When playing \emph{WaterPuzzle}, using dual-observation is not always helpful.  
\item $\text{DQN}_\text{s}$-DORL and PPO agents perform significantly better than or similar to the agent of Ye \emph{et al.}~\cite{ye2020rotation} in terms of game score and number of wins in all the test levels of \emph{GoldDigger} and \emph{TreasureKeeper}, while this is not always the case for \emph{WaterPuzzle} (cf. Table \ref{tab:ranksum_waterpuzzle_all}). 
\item Using a stochastic policy for DQN does not always lead to an overall better performed agent. On one hand, $\text{DQN}_\text{s}$ obtains statistically higher or similar game score compared to DQN with the same input observation in 7 out of 9 test levels (i.e., 3 test levels for each game). On the other hand, DQN agent wins \emph{TreasureKeeper} more times than $\text{DQN}_\text{s}$ with the same input observation.
\item Generating more levels for training a PPO agent does not lead to a better performed agent. $\text{PPO}_\text{D}$ obtains statistically higher game score than PPO with the same input observation in playing two test levels of three games and statistically lower score than PPO with the same input observation in playing five test levels of three games.
\end{itemize}

\definecolor{mygray}{gray}{.9}
\begin{table*}[htbp]

    \centering
     \caption{Results of Wilcoxon rank-sum test of agents' performance on \emph{GoldDigger} reported in Table \ref{tab:golddigger_result}. ``+/$\approx$/-'' indicates the number of times that the row agent performs statistically better/similar/worse to the column agent, respectively. The number of wins of the row agent and the column agent for three levels are shown in the brackets.}
 \resizebox{\textwidth}{!}{%
\setlength{\tabcolsep}{0.5pt}
    \begin{tabular}{cc|cccc|cccc|c}
    \toprule
\multicolumn{2}{c|}{\multirow{2}{*}{Agent}} &\multicolumn{4}{c|}{DQN} & \multicolumn{4}{c|}{$\text{DQN}_\text{s}$}& \multirow{2}{*}{Ye \emph{et al.} \cite{ye2020rotation}} \\ 
&& G1 &G2 & G3 & DORL & G1 &G2 & G3 & DORL &\\ \midrule

\multirow{4}{*}{DQN} & G1 & \  $--$ \  & \ $0/2/1$ $(0\backslash0)$ \  & \ $0/2/1$ $(0\backslash0)$ \  & \ $0/1/2$ $(0\backslash0)$ \  & \ \cellcolor[rgb]{.9,.9,.9}$0/2/1$ $(0\backslash0)$ \  & \ $0/0/3$ $(0\backslash0)$ \  & \ $0/0/3$ $(0\backslash0)$ \  & \ $0/0/3$ $(0\backslash1)$ \  & \ $0/1/2$ $(0\backslash0)$ \  \\ 
 & G2 & \ $1/2/0$ $(0\backslash0)$ \  & \  $--$ \  & \ $1/1/1$ $(0\backslash0)$ \  & \ $0/2/1$ $(0\backslash0)$ \  & \ $1/1/1$ $(0\backslash0)$ \  & \ \cellcolor[rgb]{.9,.9,.9}$0/1/2$ $(0\backslash0)$ \  & \ $0/1/2$ $(0\backslash0)$ \  & \ $0/1/2$ $(0\backslash1)$ \  & \ $1/0/2$ $(0\backslash0)$ \  \\ 
 & G3 & \ $1/2/0$ $(0\backslash0)$ \  & \ $1/1/1$ $(0\backslash0)$ \  & \  $--$ \  & \ $0/1/2$ $(0\backslash0)$ \  & \ $0/2/1$ $(0\backslash0)$ \  & \ $0/0/3$ $(0\backslash0)$ \  & \ \cellcolor[rgb]{.9,.9,.9}$0/0/3$ $(0\backslash0)$ \  & \ $0/0/3$ $(0\backslash1)$ \  & \ $1/1/1$ $(0\backslash0)$ \  \\ 
 & DORL & \ $2/1/0$ $(0\backslash0)$ \  & \ $1/2/0$ $(0\backslash0)$ \  & \ $2/1/0$ $(0\backslash0)$ \  & \  $--$ \  & \ $2/1/0$ $(0\backslash0)$ \  & \ $0/2/1$ $(0\backslash0)$ \  & \ $0/1/2$ $(0\backslash0)$ \  & \ \cellcolor[rgb]{.9,.9,.9}$0/1/2$ $(0\backslash1)$ \  & \ $2/0/1$ $(0\backslash0)$ \  \\ \midrule
\multirow{4}{*}{$\text{DQN}_\text{s}$} & G1 & \ \cellcolor[rgb]{.9,.9,.9}$1/2/0$ $(0\backslash0)$ \  & \ $1/1/1$ $(0\backslash0)$ \  & \ $1/2/0$ $(0\backslash0)$ \  & \ $0/1/2$ $(0\backslash0)$ \  & \  $--$ \  & \ $0/0/3$ $(0\backslash0)$ \  & \ $0/0/3$ $(0\backslash0)$ \  & \ $0/0/3$ $(0\backslash1)$ \  & \ $1/1/1$ $(0\backslash0)$ \  \\ 
 & G2 & \ $3/0/0$ $(0\backslash0)$ \  & \ \cellcolor[rgb]{.9,.9,.9}$2/1/0$ $(0\backslash0)$ \  & \ $3/0/0$ $(0\backslash0)$ \  & \ $1/2/0$ $(0\backslash0)$ \  & \ $3/0/0$ $(0\backslash0)$ \  & \  $--$ \  & \ $0/3/0$ $(0\backslash0)$ \  & \ $0/2/1$ $(0\backslash1)$ \  & \ $3/0/0$ $(0\backslash0)$ \  \\ 
 & G3 & \ $3/0/0$ $(0\backslash0)$ \  & \ $2/1/0$ $(0\backslash0)$ \  & \ \cellcolor[rgb]{.9,.9,.9}$3/0/0$ $(0\backslash0)$ \  & \ $2/1/0$ $(0\backslash0)$ \  & \ $3/0/0$ $(0\backslash0)$ \  & \ $0/3/0$ $(0\backslash0)$ \  & \  $--$ \  & \ $0/2/1$ $(0\backslash1)$ \  & \ $3/0/0$ $(0\backslash0)$ \  \\ 
 & DORL & \ $3/0/0$ $(1\backslash0)$ \  & \ $2/1/0$ $(1\backslash0)$ \  & \ $3/0/0$ $(1\backslash0)$ \  & \ \cellcolor[rgb]{.9,.9,.9}$2/1/0$ $(1\backslash0)$ \  & \ $3/0/0$ $(1\backslash0)$ \  & \ $1/2/0$ $(1\backslash0)$ \  & \ $1/2/0$ $(1\backslash0)$ \  & \  $--$ \  & \ $3/0/0$ $(1\backslash0)$ \  \\ \midrule
\multicolumn{2}{c|}{\multirow{1}{*}{Ye \emph{et al.} \cite{ye2020rotation}}} & \ $2/1/0$ $(0\backslash0)$ \  & \ $2/0/1$ $(0\backslash0)$ \  & \ $1/1/1$ $(0\backslash0)$ \  & \ $1/0/2$ $(0\backslash0)$ \  & \ $1/1/1$ $(0\backslash0)$ \  & \ $0/0/3$ $(0\backslash0)$ \  & \ $0/0/3$ $(0\backslash0)$ \  & \ $0/0/3$ $(0\backslash1)$ \  & \  $--$ \  \\

\bottomrule
 
\end{tabular}
}
\vspace{4mm}   

 \resizebox{\textwidth}{!}{%
\setlength{\tabcolsep}{0.5pt}
\begin{tabular}{cc|cccc|cccc|c}
    \toprule
\multicolumn{2}{c|}{\multirow{2}{*}{Agent}} &\multicolumn{4}{c|}{PPO} & \multicolumn{4}{c|}{$\text{PPO}_\text{D}$}& \multirow{2}{*}{Ye \emph{et al.} \cite{ye2020rotation}} \\ 
&& G1 &G2 & G3 & DORL & G1 &G2 & G3 & DORL &\\ \midrule

\multirow{4}{*}{PPO} & G1 & \  $--$ \  & \ $0/1/2$ $(0\backslash0)$ \  & \ $0/2/1$ $(0\backslash0)$ \  & \ $0/0/3$ $(0\backslash0)$ \  & \ \cellcolor[rgb]{.9,.9,.9}$1/2/0$ $(0\backslash0)$ \  & \ $0/0/3$ $(0\backslash0)$ \  & \ $0/1/2$ $(0\backslash0)$ \  & \ $0/1/2$ $(0\backslash1)$ \  & \ $1/2/0$ $(0\backslash0)$ \  \\ 
 & G2 & \ $2/1/0$ $(0\backslash0)$ \  & \  $--$ \  & \ $0/3/0$ $(0\backslash0)$ \  & \ $0/2/1$ $(0\backslash0)$ \  & \ $1/2/0$ $(0\backslash0)$ \  & \ \cellcolor[rgb]{.9,.9,.9}$0/1/2$ $(0\backslash0)$ \  & \ $0/3/0$ $(0\backslash0)$ \  & \ $0/3/0$ $(0\backslash1)$ \  & \ $2/1/0$ $(0\backslash0)$ \  \\ 
 & G3 & \ $1/2/0$ $(0\backslash0)$ \  & \ $0/3/0$ $(0\backslash0)$ \  & \  $--$ \  & \ $0/2/1$ $(0\backslash0)$ \  & \ $2/1/0$ $(0\backslash0)$ \  & \ $0/1/2$ $(0\backslash0)$ \  & \ \cellcolor[rgb]{.9,.9,.9}$0/3/0$ $(0\backslash0)$ \  & \ $0/3/0$ $(0\backslash1)$ \  & \ $2/1/0$ $(0\backslash0)$ \  \\ 
 & DORL & \ $3/0/0$ $(0\backslash0)$ \  & \ $1/2/0$ $(0\backslash0)$ \  & \ $1/2/0$ $(0\backslash0)$ \  & \  $--$ \  & \ $3/0/0$ $(0\backslash0)$ \  & \ $0/3/0$ $(0\backslash0)$ \  & \ $2/1/0$ $(0\backslash0)$ \  & \ \cellcolor[rgb]{.9,.9,.9}$0/3/0$ $(0\backslash1)$ \  & \ $3/0/0$ $(0\backslash0)$ \  \\ \midrule
\multirow{4}{*}{$\text{PPO}_\text{D}$} & G1 & \ \cellcolor[rgb]{.9,.9,.9}$0/2/1$ $(0\backslash0)$ \  & \ $0/2/1$ $(0\backslash0)$ \  & \ $0/1/2$ $(0\backslash0)$ \  & \ $0/0/3$ $(0\backslash0)$ \  & \  $--$ \  & \ $0/0/3$ $(0\backslash0)$ \  & \ $0/2/1$ $(0\backslash0)$ \  & \ $0/2/1$ $(0\backslash1)$ \  & \ $1/2/0$ $(0\backslash0)$ \  \\ 
 & G2 & \ $3/0/0$ $(0\backslash0)$ \  & \ \cellcolor[rgb]{.9,.9,.9}$2/1/0$ $(0\backslash0)$ \  & \ $2/1/0$ $(0\backslash0)$ \  & \ $0/3/0$ $(0\backslash0)$ \  & \ $3/0/0$ $(0\backslash0)$ \  & \  $--$ \  & \ $2/1/0$ $(0\backslash0)$ \  & \ $0/3/0$ $(0\backslash1)$ \  & \ $3/0/0$ $(0\backslash0)$ \  \\ 
 & G3 & \ $2/1/0$ $(0\backslash0)$ \  & \ $0/3/0$ $(0\backslash0)$ \  & \ \cellcolor[rgb]{.9,.9,.9}$0/3/0$ $(0\backslash0)$ \  & \ $0/1/2$ $(0\backslash0)$ \  & \ $1/2/0$ $(0\backslash0)$ \  & \ $0/1/2$ $(0\backslash0)$ \  & \  $--$ \  & \ $0/3/0$ $(0\backslash1)$ \  & \ $3/0/0$ $(0\backslash0)$ \  \\ 
 & DORL & \ $2/1/0$ $(1\backslash0)$ \  & \ $0/3/0$ $(1\backslash0)$ \  & \ $0/3/0$ $(1\backslash0)$ \  & \ \cellcolor[rgb]{.9,.9,.9}$0/3/0$ $(1\backslash0)$ \  & \ $1/2/0$ $(1\backslash0)$ \  & \ $0/3/0$ $(1\backslash0)$ \  & \ $0/3/0$ $(1\backslash0)$ \  & \  $--$ \  & \ $2/1/0$ $(1\backslash0)$ \  \\ \midrule
\multicolumn{2}{c|}{\multirow{1}{*}{Ye \emph{et al.} \cite{ye2020rotation}}} & \ $0/2/1$ $(0\backslash0)$ \  & \ $0/1/2$ $(0\backslash0)$ \  & \ $0/1/2$ $(0\backslash0)$ \  & \ $0/0/3$ $(0\backslash0)$ \  & \ $0/2/1$ $(0\backslash0)$ \  & \ $0/0/3$ $(0\backslash0)$ \  & \ $0/0/3$ $(0\backslash0)$ \  & \ $0/1/2$ $(0\backslash1)$ \  & \  $--$ \  \\ 
 \bottomrule
    \end{tabular}
    }
 \vspace{4mm}
 
        \begin{tabular}{cc|cccc|c}
    \toprule
\multicolumn{2}{c|}{\multirow{2}{*}{Agent}} &\multicolumn{4}{c|}{A2C}& \multirow{2}{*}{Ye \emph{et al.} \cite{ye2020rotation}} \\ 
&& G1 &G2 & G3 & DORL &\\ \midrule

\multirow{4}{*}{A2C} & G1 &  $--$\  & \ $0/1/2$ $(0\backslash0)$ \  & \ $0/1/2$ $(0\backslash0)$ \  & \ $0/1/2$ $(0\backslash0)$ \  & \ $0/2/1$ $(0\backslash0)$ \  \\ 
 & G2 & \ $2/1/0$ $(0\backslash0)$ \  &  $--$\  & \ $0/1/2$ $(0\backslash0)$ \  & \ $1/1/1$ $(0\backslash0)$ \  & \ $1/2/0$ $(0\backslash0)$ \  \\ 
 & G3 & \ $2/1/0$ $(0\backslash0)$ \  & \ $2/1/0$ $(0\backslash0)$ \  &  $--$\  & \ $0/2/1$ $(0\backslash0)$ \  & \ $2/1/0$ $(0\backslash0)$ \  \\ 
 & DORL & \ $2/1/0$ $(0\backslash0)$ \  & \ $1/1/1$ $(0\backslash0)$ \  & \ $1/2/0$ $(0\backslash0)$ \  &  $--$\  & \ $2/0/1$ $(0\backslash0)$ \  \\ \midrule
\multicolumn{2}{c|}{\multirow{1}{*}{Ye \emph{et al.} \cite{ye2020rotation}}} & \ $1/2/0$ $(0\backslash0)$ \  & \ $0/2/1$ $(0\backslash0)$ \  & \ $0/1/2$ $(0\backslash0)$ \  & \ $1/0/2$ $(0\backslash0)$ \  &  $--$\  \\ 

\bottomrule
    \end{tabular}
    \label{tab:ranksum_golddigger_all}
\end{table*}

\begin{table*}[htbp]
    \centering
            \caption{Results of Wilcoxon rank-sum test of agents' performance on \emph{TreasureKeeper} reported in Table \ref{tab:treasurekeeper_result}. 
            ``+/$\approx$/-'' indicates the number of times that the row agent performs statistically better/similar/worse to the column agent, respectively. The number of wins of the row agent and the column agent for three levels are shown in the brackets.}
\resizebox{\textwidth}{!}{%
\setlength{\tabcolsep}{0.5pt}
\scriptsize
    \begin{tabular}{cc|cccc|cccc|c}
    \toprule
\multicolumn{2}{c|}{\multirow{2}{*}{Agent}} &\multicolumn{4}{c|}{DQN} & \multicolumn{4}{c|}{$\text{DQN}_\text{s}$}& \multirow{2}{*}{Ye \emph{et al.} \cite{ye2020rotation}} \\ 
&& G1 &G2 & G3 & DORL & G1 &G2 & G3 & DORL &\\ \midrule

\multirow{4}{*}{DQN} & G1 & \  $--$ \  & \ $0/2/1$ $(10\backslash20)$ \  & \ $0/3/0$ $(10\backslash14)$ \  & \ $0/3/0$ $(10\backslash21)$ \  & \ \cellcolor[rgb]{.9,.9,.9}$0/3/0$ $(10\backslash7)$ \  & \ $0/2/1$ $(10\backslash10)$ \  & \ $0/3/0$ $(10\backslash7)$ \  & \ $0/3/0$ $(10\backslash19)$ \  & \ $2/1/0$ $(10\backslash0)$ \  \\ 
 & G2 & \ $1/2/0$ $(20\backslash10)$ \  & \  $--$ \  & \ $1/2/0$ $(20\backslash14)$ \  & \ $0/3/0$ $(20\backslash21)$ \  & \ $3/0/0$ $(20\backslash7)$ \  & \ \cellcolor[rgb]{.9,.9,.9}$0/3/0$ $(20\backslash10)$ \  & \ $2/1/0$ $(20\backslash7)$ \  & \ $0/3/0$ $(20\backslash19)$ \  & \ $3/0/0$ $(20\backslash0)$ \  \\ 
 & G3 & \ $0/3/0$ $(14\backslash10)$ \  & \ $0/2/1$ $(14\backslash20)$ \  & \  $--$ \  & \ $0/3/0$ $(14\backslash21)$ \  & \ $1/2/0$ $(14\backslash7)$ \  & \ $0/2/1$ $(14\backslash10)$ \  & \ \cellcolor[rgb]{.9,.9,.9}$1/2/0$ $(14\backslash7)$ \  & \ $0/3/0$ $(14\backslash19)$ \  & \ $2/1/0$ $(14\backslash0)$ \  \\ 
 & DORL & \ $0/3/0$ $(21\backslash10)$ \  & \ $0/3/0$ $(21\backslash20)$ \  & \ $0/3/0$ $(21\backslash14)$ \  & \  $--$ \  & \ $1/2/0$ $(21\backslash7)$ \  & \ $0/3/0$ $(21\backslash10)$ \  & \ $1/2/0$ $(21\backslash7)$ \  & \ \cellcolor[rgb]{.9,.9,.9}$0/3/0$ $(21\backslash19)$ \  & \ $2/1/0$ $(21\backslash0)$ \  \\ \midrule
\multirow{4}{*}{$\text{DQN}_\text{s}$} & G1 & \ \cellcolor[rgb]{.9,.9,.9}$0/3/0$ $(7\backslash10)$ \  & \ $0/0/3$ $(7\backslash20)$ \  & \ $0/2/1$ $(7\backslash14)$ \  & \ $0/2/1$ $(7\backslash21)$ \  & \  $--$ \  & \ $0/2/1$ $(7\backslash10)$ \  & \ $0/3/0$ $(7\backslash7)$ \  & \ $0/2/1$ $(7\backslash19)$ \  & \ $2/1/0$ $(7\backslash0)$ \  \\ 
 & G2 & \ $1/2/0$ $(10\backslash10)$ \  & \ \cellcolor[rgb]{.9,.9,.9}$0/3/0$ $(10\backslash20)$ \  & \ $1/2/0$ $(10\backslash14)$ \  & \ $0/3/0$ $(10\backslash21)$ \  & \ $1/2/0$ $(10\backslash7)$ \  & \  $--$ \  & \ $1/2/0$ $(10\backslash7)$ \  & \ $0/3/0$ $(10\backslash19)$ \  & \ $3/0/0$ $(10\backslash0)$ \  \\ 
 & G3 & \ $0/3/0$ $(7\backslash10)$ \  & \ $0/1/2$ $(7\backslash20)$ \  & \ \cellcolor[rgb]{.9,.9,.9}$0/2/1$ $(7\backslash14)$ \  & \ $0/2/1$ $(7\backslash21)$ \  & \ $0/3/0$ $(7\backslash7)$ \  & \ $0/2/1$ $(7\backslash10)$ \  & \  $--$ \  & \ $0/2/1$ $(7\backslash19)$ \  & \ $2/1/0$ $(7\backslash0)$ \  \\ 
 & DORL & \ $0/3/0$ $(19\backslash10)$ \  & \ $0/3/0$ $(19\backslash20)$ \  & \ $0/3/0$ $(19\backslash14)$ \  & \ \cellcolor[rgb]{.9,.9,.9}$0/3/0$ $(19\backslash21)$ \  & \ $1/2/0$ $(19\backslash7)$ \  & \ $0/3/0$ $(19\backslash10)$ \  & \ $1/2/0$ $(19\backslash7)$ \  & \  $--$ \  & \ $2/1/0$ $(19\backslash0)$ \  \\ \midrule
\multicolumn{2}{c|}{\multirow{1}{*}{Ye \emph{et al.} \cite{ye2020rotation}}} & \ $0/1/2$ $(0\backslash10)$ \  & \ $0/0/3$ $(0\backslash20)$ \  & \ $0/1/2$ $(0\backslash14)$ \  & \ $0/1/2$ $(0\backslash21)$ \  & \ $0/1/2$ $(0\backslash7)$ \  & \ $0/0/3$ $(0\backslash10)$ \  & \ $0/1/2$ $(0\backslash7)$ \  & \ $0/1/2$ $(0\backslash19)$ \  & \  $--$ \  \\

 \bottomrule
    \end{tabular}
    }
    \vspace{4mm}
    
    \resizebox{\textwidth}{!}{%
\setlength{\tabcolsep}{0.5pt}
        \begin{tabular}{cc|cccc|cccc|c}
    \toprule
\multicolumn{2}{c|}{\multirow{2}{*}{Agent}} &\multicolumn{4}{c|}{PPO} & \multicolumn{4}{c|}{$\text{PPO}_\text{D}$}& \multirow{2}{*}{Ye \emph{et al.} \cite{ye2020rotation}} \\ 
&& G1 &G2 & G3 & DORL & G1 &G2 & G3 & DORL &\\ \midrule

\multirow{4}{*}{PPO} & G1 & \  $--$ \  & \ $0/3/0$ $(5\backslash7)$ \  & \ $0/2/1$ $(5\backslash9)$ \  & \ $0/3/0$ $(5\backslash4)$ \  & \ \cellcolor[rgb]{.9,.9,.9}$0/3/0$ $(5\backslash8)$ \  & \ $0/3/0$ $(5\backslash8)$ \  & \ $0/3/0$ $(5\backslash6)$ \  & \ $0/2/1$ $(5\backslash13)$ \  & \ $1/2/0$ $(5\backslash0)$ \  \\ 
 & G2 & \ $0/3/0$ $(7\backslash5)$ \  & \  $--$ \  & \ $0/3/0$ $(7\backslash9)$ \  & \ $0/3/0$ $(7\backslash4)$ \  & \ $0/3/0$ $(7\backslash8)$ \  & \ \cellcolor[rgb]{.9,.9,.9}$1/2/0$ $(7\backslash8)$ \  & \ $1/2/0$ $(7\backslash6)$ \  & \ $0/3/0$ $(7\backslash13)$ \  & \ $3/0/0$ $(7\backslash0)$ \  \\ 
 & G3 & \ $1/2/0$ $(9\backslash5)$ \  & \ $0/3/0$ $(9\backslash7)$ \  & \  $--$ \  & \ $0/3/0$ $(9\backslash4)$ \  & \ $0/3/0$ $(9\backslash8)$ \  & \ $0/3/0$ $(9\backslash8)$ \  & \ \cellcolor[rgb]{.9,.9,.9}$0/3/0$ $(9\backslash6)$ \  & \ $0/3/0$ $(9\backslash13)$ \  & \ $2/1/0$ $(9\backslash0)$ \  \\ 
 & DORL & \ $0/3/0$ $(4\backslash5)$ \  & \ $0/3/0$ $(4\backslash7)$ \  & \ $0/3/0$ $(4\backslash9)$ \  & \  $--$ \  & \ $0/3/0$ $(4\backslash8)$ \  & \ $0/3/0$ $(4\backslash8)$ \  & \ $0/3/0$ $(4\backslash6)$ \  & \ \cellcolor[rgb]{.9,.9,.9}$0/3/0$ $(4\backslash13)$ \  & \ $2/1/0$ $(4\backslash0)$ \  \\ \midrule
\multirow{4}{*}{$\text{PPO}_\text{D}$} & G1 & \ \cellcolor[rgb]{.9,.9,.9}$0/3/0$ $(8\backslash5)$ \  & \ $0/3/0$ $(8\backslash7)$ \  & \ $0/3/0$ $(8\backslash9)$ \  & \ $0/3/0$ $(8\backslash4)$ \  & \  $--$ \  & \ $0/3/0$ $(8\backslash8)$ \  & \ $0/3/0$ $(8\backslash6)$ \  & \ $0/3/0$ $(8\backslash13)$ \  & \ $2/1/0$ $(8\backslash0)$ \  \\ 
 & G2 & \ $0/3/0$ $(8\backslash5)$ \  & \ \cellcolor[rgb]{.9,.9,.9}$0/2/1$ $(8\backslash7)$ \  & \ $0/3/0$ $(8\backslash9)$ \  & \ $0/3/0$ $(8\backslash4)$ \  & \ $0/3/0$ $(8\backslash8)$ \  & \  $--$ \  & \ $0/3/0$ $(8\backslash6)$ \  & \ $0/3/0$ $(8\backslash13)$ \  & \ $2/1/0$ $(8\backslash0)$ \  \\ 
 & G3 & \ $0/3/0$ $(6\backslash5)$ \  & \ $0/2/1$ $(6\backslash7)$ \  & \ \cellcolor[rgb]{.9,.9,.9}$0/3/0$ $(6\backslash9)$ \  & \ $0/3/0$ $(6\backslash4)$ \  & \ $0/3/0$ $(6\backslash8)$ \  & \ $0/3/0$ $(6\backslash8)$ \  & \  $--$ \  & \ $0/3/0$ $(6\backslash13)$ \  & \ $2/1/0$ $(6\backslash0)$ \  \\ 
 & DORL & \ $1/2/0$ $(13\backslash5)$ \  & \ $0/3/0$ $(13\backslash7)$ \  & \ $0/3/0$ $(13\backslash9)$ \  & \ \cellcolor[rgb]{.9,.9,.9}$0/3/0$ $(13\backslash4)$ \  & \ $0/3/0$ $(13\backslash8)$ \  & \ $0/3/0$ $(13\backslash8)$ \  & \ $0/3/0$ $(13\backslash6)$ \  & \  $--$ \  & \ $2/1/0$ $(13\backslash0)$ \  \\ \midrule
\multicolumn{2}{c|}{\multirow{1}{*}{Ye \emph{et al.} \cite{ye2020rotation}}} & \ $0/2/1$ $(0\backslash5)$ \  & \ $0/0/3$ $(0\backslash7)$ \  & \ $0/1/2$ $(0\backslash9)$ \  & \ $0/1/2$ $(0\backslash4)$ \  & \ $0/1/2$ $(0\backslash8)$ \  & \ $0/1/2$ $(0\backslash8)$ \  & \ $0/1/2$ $(0\backslash6)$ \  & \ $0/1/2$ $(0\backslash13)$ \  & \  $--$ \  \\

\bottomrule
    \end{tabular}}
    
\vspace{1em}
    \begin{tabular}{cc|cccc|c}
    \toprule
\multicolumn{2}{c|}{\multirow{2}{*}{Agent}} &\multicolumn{4}{c|}{A2C}& \multirow{2}{*}{Ye \emph{et al.} \cite{ye2020rotation}} \\ 
&& G1 &G2 & G3 & DORL &\\ \midrule

\multirow{4}{*}{A2C} & G1 &  $--$\  & \ $0/3/0$ $(4\backslash13)$ \  & \ $0/0/3$ $(4\backslash22)$ \  & \ $0/3/0$ $(4\backslash11)$ \  & \ $2/1/0$ $(4\backslash0)$ \  \\ 
 & G2 & \ $0/3/0$ $(13\backslash4)$ \  &  $--$\  & \ $0/2/1$ $(13\backslash22)$ \  & \ $0/3\backslash0$ $(13\backslash11)$ \  & \ $2/1/0$ $(13\backslash0)$ \  \\ 
 & G3 & \ $3/0/0$ $(22\backslash4)$ \  & \ $1/2/0$ $(22\backslash13)$ \  &  $--$\  & \ $2/1\backslash0$ $(22\backslash11)$ \  & \ $3/0/0$ $(22\backslash0)$ \  \\ 
 & DORL & \ $0/3/0$ $(11\backslash4)$ \  & \ $0/3/0$ $(11\backslash13)$ \  & \ $0/1/2$ $(11\backslash22)$ \  &  $--$\  & \ $2/1/0$ $(11\backslash0)$ \  \\ \midrule
\multicolumn{2}{c|}{\multirow{1}{*}{Ye \emph{et al.} \cite{ye2020rotation}}} & \ $0/1/2$ $(0\backslash4)$ \  & \ $0/1/2$ $(0\backslash13)$ \  & \ $0/0/3$ $(0\backslash22)$ \  & \ $0/1/2$ $(0\backslash11)$ \  &  $--$\  \\ 

\bottomrule
    \end{tabular}
    \label{tab:ranksum_treasurekeeper_all}
\end{table*}

\begin{table*}[htbp]
    \centering
     \caption{Results of Wilcoxon rank-sum test of agents' performance on \emph{WaterPuzzle} reported in Table \ref{tab:waterpuzzle_result}. ``+/$\approx$/-'' indicates the number of times that the row agent performs statistically better/similar/worse to the column agent, respectively. The number of wins of the row agent and the column agent for three levels are shown in the brackets.}
            \resizebox{\textwidth}{!}{%
\setlength{\tabcolsep}{0.5pt}
    \begin{tabular}{cc|cccc|cccc|c}
    \toprule
\multicolumn{2}{c|}{\multirow{2}{*}{Agent}} &\multicolumn{4}{c|}{DQN} & \multicolumn{4}{c|}{$\text{DQN}_\text{s}$}& \multirow{2}{*}{Ye \emph{et al.} \cite{ye2020rotation}} \\ 
&& G1 &G2 & G3 & DORL & G1 &G2 & G3 & DORL &\\ \midrule

\multirow{4}{*}{DQN} & G1 & \  $--$ \  & \ $0/3/0$ $(0\backslash0)$ \  & \ $0/3/0$ $(0\backslash0)$ \  & \ $0/3/0$ $(0\backslash0)$ \  & \ \cellcolor[rgb]{.9,.9,.9}$0/3/0$ $(0\backslash0)$ \  & \ $0/3/0$ $(0\backslash0)$ \  & \ $0/3/0$ $(0\backslash2)$ \  & \ $0/3/0$ $(0\backslash0)$ \  & \ $0/1/2$ $(0\backslash1)$ \  \\ 
 & G2 & \ $0/3/0$ $(0\backslash0)$ \  & \  $--$ \  & \ $0/3/0$ $(0\backslash0)$ \  & \ $0/3/0$ $(0\backslash0)$ \  & \ $0/3/0$ $(0\backslash0)$ \  & \ \cellcolor[rgb]{.9,.9,.9}$0/3/0$ $(0\backslash0)$ \  & \ $0/3/0$ $(0\backslash2)$ \  & \ $0/3/0$ $(0\backslash0)$ \  & \ $0/1/2$ $(0\backslash1)$ \  \\ 
 & G3 & \ $0/3/0$ $(0\backslash0)$ \  & \ $0/3/0$ $(0\backslash0)$ \  & \  $--$ \  & \ $0/3/0$ $(0\backslash0)$ \  & \ $0/3/0$ $(0\backslash0)$ \  & \ $0/3/0$ $(0\backslash0)$ \  & \ \cellcolor[rgb]{.9,.9,.9}$0/3/0$ $(0\backslash2)$ \  & \ $0/3/0$ $(0\backslash0)$ \  & \ $0/1/2$ $(0\backslash1)$ \  \\ 
 & DORL & \ $0/3/0$ $(0\backslash0)$ \  & \ $0/3/0$ $(0\backslash0)$ \  & \ $0/3/0$ $(0\backslash0)$ \  & \  $--$ \  & \ $0/3/0$ $(0\backslash0)$ \  & \ $0/3/0$ $(0\backslash0)$ \  & \ $0/3/0$ $(0\backslash2)$ \  & \ \cellcolor[rgb]{.9,.9,.9}$0/3/0$ $(0\backslash0)$ \  & \ $0/1/2$ $(0\backslash1)$ \  \\ \midrule
\multirow{4}{*}{$\text{DQN}_\text{s}$} & G1 & \ \cellcolor[rgb]{.9,.9,.9}$0/3/0$ $(0\backslash0)$ \  & \ $0/3/0$ $(0\backslash0)$ \  & \ $0/3/0$ $(0\backslash0)$ \  & \ $0/3/0$ $(0\backslash0)$ \  & \  $--$ \  & \ $0/3/0$ $(0\backslash0)$ \  & \ $0/3/0$ $(0\backslash2)$ \  & \ $0/3/0$ $(0\backslash0)$ \  & \ $0/1/2$ $(0\backslash1)$ \  \\ 
 & G2 & \ $0/3/0$ $(0\backslash0)$ \  & \ \cellcolor[rgb]{.9,.9,.9}$0/3/0$ $(0\backslash0)$ \  & \ $0/3/0$ $(0\backslash0)$ \  & \ $0/3/0$ $(0\backslash0)$ \  & \ $0/3/0$ $(0\backslash0)$ \  & \  $--$ \  & \ $0/3/0$ $(0\backslash2)$ \  & \ $0/3/0$ $(0\backslash0)$ \  & \ $0/1/2$ $(0\backslash1)$ \  \\ 
 & G3 & \ $0/3/0$ $(2\backslash0)$ \  & \ $0/3/0$ $(2\backslash0)$ \  & \ \cellcolor[rgb]{.9,.9,.9}$0/3/0$ $(2\backslash0)$ \  & \ $0/3/0$ $(2\backslash0)$ \  & \ $0/3/0$ $(2\backslash0)$ \  & \ $0/3/0$ $(2\backslash0)$ \  & \  $--$ \  & \ $0/3/0$ $(2\backslash0)$ \  & \ $0/2/1$ $(2\backslash1)$ \  \\ 
 & DORL & \ $0/3/0$ $(0\backslash0)$ \  & \ $0/3/0$ $(0\backslash0)$ \  & \ $0/3/0$ $(0\backslash0)$ \  & \ \cellcolor[rgb]{.9,.9,.9}$0/3/0$ $(0\backslash0)$ \  & \ $0/3/0$ $(0\backslash0)$ \  & \ $0/3/0$ $(0\backslash0)$ \  & \ $0/3/0$ $(0\backslash2)$ \  & \  $--$ \  & \ $0/2/1$ $(0\backslash1)$ \  \\ \midrule
\multicolumn{2}{c|}{\multirow{1}{*}{Ye \emph{et al.} \cite{ye2020rotation}}} & \ $2/1/0$ $(1\backslash0)$ \  & \ $2/1/0$ $(1\backslash0)$ \  & \ $2/1/0$ $(1\backslash0)$ \  & \ $2/1/0$ $(1\backslash0)$ \  & \ $2/1/0$ $(1\backslash0)$ \  & \ $2/1/0$ $(1\backslash0)$ \  & \ $1/2/0$ $(1\backslash2)$ \  & \ $1/2/0$ $(1\backslash0)$ \  & \  $--$ \  \\ 

 \bottomrule
 
\end{tabular}
}

\vspace{4mm}    

\resizebox{\textwidth}{!}{%
\setlength{\tabcolsep}{0.5pt}
\begin{tabular}{cc|cccc|cccc|c}
    \toprule
\multicolumn{2}{c|}{\multirow{2}{*}{Agent}} &\multicolumn{4}{c|}{PPO} & \multicolumn{4}{c|}{$\text{PPO}_\text{D}$}& \multirow{2}{*}{Ye \emph{et al.} \cite{ye2020rotation}} \\ 
&& G1 &G2 & G3 & DORL & G1 &G2 & G3 & DORL &\\ \midrule

\multirow{4}{*}{PPO} & G1 & \  $--$ \  & \ $0/1/2$ $(0\backslash7)$ \  & \ $1/2/0$ $(0\backslash0)$ \  & \ $1/2/0$ $(0\backslash0)$ \  & \ \cellcolor[rgb]{.9,.9,.9}$1/2/0$ $(0\backslash0)$ \  & \ $1/2/0$ $(0\backslash0)$ \  & \ $1/2/0$ $(0\backslash0)$ \  & \ $1/2/0$ $(0\backslash0)$ \  & \ $0/2/1$ $(0\backslash1)$ \  \\ 
 & G2 & \ $2/1/0$ $(7\backslash0)$ \  & \  $--$ \  & \ $2/1/0$ $(7\backslash0)$ \  & \ $2/1/0$ $(7\backslash0)$ \  & \ $2/1/0$ $(7\backslash0)$ \  & \ \cellcolor[rgb]{.9,.9,.9}$2/1/0$ $(7\backslash0)$ \  & \ $2/1/0$ $(7\backslash0)$ \  & \ $2/1/0$ $(7\backslash0)$ \  & \ $2/1/0$ $(7\backslash1)$ \  \\ 
 & G3 & \ $0/2/1$ $(0\backslash0)$ \  & \ $0/1/2$ $(0\backslash7)$ \  & \  $--$ \  & \ $0/3/0$ $(0\backslash0)$ \  & \ $0/3/0$ $(0\backslash0)$ \  & \ $0/3/0$ $(0\backslash0)$ \  & \ \cellcolor[rgb]{.9,.9,.9}$0/3/0$ $(0\backslash0)$ \  & \ $0/3/0$ $(0\backslash0)$ \  & \ $0/1/2$ $(0\backslash1)$ \  \\ 
 & DORL & \ $0/2/1$ $(0\backslash0)$ \  & \ $0/1/2$ $(0\backslash7)$ \  & \ $0/3/0$ $(0\backslash0)$ \  & \  $--$ \  & \ $0/3/0$ $(0\backslash0)$ \  & \ $0/3/0$ $(0\backslash0)$ \  & \ $0/3/0$ $(0\backslash0)$ \  & \ \cellcolor[rgb]{.9,.9,.9}$0/3/0$ $(0\backslash0)$ \  & \ $0/1/2$ $(0\backslash1)$ \  \\ \midrule
\multirow{4}{*}{$\text{PPO}_\text{D}$} & G1 & \ \cellcolor[rgb]{.9,.9,.9}$0/2/1$ $(0\backslash0)$ \  & \ $0/1/2$ $(0\backslash7)$ \  & \ $0/3/0$ $(0\backslash0)$ \  & \ $0/3/0$ $(0\backslash0)$ \  & \  $--$ \  & \ $0/3/0$ $(0\backslash0)$ \  & \ $0/3/0$ $(0\backslash0)$ \  & \ $0/3/0$ $(0\backslash0)$ \  & \ $0/1/2$ $(0\backslash1)$ \  \\ 
 & G2 & \ $0/2/1$ $(0\backslash0)$ \  & \ \cellcolor[rgb]{.9,.9,.9}$0/1/2$ $(0\backslash7)$ \  & \ $0/3/0$ $(0\backslash0)$ \  & \ $0/3/0$ $(0\backslash0)$ \  & \ $0/3/0$ $(0\backslash0)$ \  & \  $--$ \  & \ $0/3/0$ $(0\backslash0)$ \  & \ $0/3/0$ $(0\backslash0)$ \  & \ $0/1/2$ $(0\backslash1)$ \  \\ 
 & G3 & \ $0/2/1$ $(0\backslash0)$ \  & \ $0/1/2$ $(0\backslash7)$ \  & \ \cellcolor[rgb]{.9,.9,.9}$0/3/0$ $(0\backslash0)$ \  & \ $0/3/0$ $(0\backslash0)$ \  & \ $0/3/0$ $(0\backslash0)$ \  & \ $0/3/0$ $(0\backslash0)$ \  & \  $--$ \  & \ $0/3/0$ $(0\backslash0)$ \  & \ $0/1/2$ $(0\backslash1)$ \  \\ 
 & DORL & \ $0/2/1$ $(0\backslash0)$ \  & \ $0/1/2$ $(0\backslash7)$ \  & \ $0/3/0$ $(0\backslash0)$ \  & \ \cellcolor[rgb]{.9,.9,.9}$0/3/0$ $(0\backslash0)$ \  & \ $0/3/0$ $(0\backslash0)$ \  & \ $0/3/0$ $(0\backslash0)$ \  & \ $0/3/0$ $(0\backslash0)$ \  & \  $--$ \  & \ $0/1/2$ $(0\backslash1)$ \  \\ \midrule
\multicolumn{2}{c|}{\multirow{1}{*}{Ye \emph{et al.} \cite{ye2020rotation}}} & \ $1/2/0$ $(1\backslash0)$ \  & \ $0/1/2$ $(1\backslash7)$ \  & \ $2/1/0$ $(1\backslash0)$ \  & \ $2/1/0$ $(1\backslash0)$ \  & \ $2/1/0$ $(1\backslash0)$ \  & \ $2/1/0$ $(1\backslash0)$ \  & \ $2/1/0$ $(1\backslash0)$ \  & \ $2/1/0$ $(1\backslash0)$ \  & \  $--$ \  \\

 \bottomrule
    \end{tabular}
    }
 \vspace{4mm}
 
        \begin{tabular}{cc|cccc|c}
    \toprule
\multicolumn{2}{c|}{\multirow{2}{*}{Agent}} &\multicolumn{4}{c|}{A2C}& \multirow{2}{*}{Ye \emph{et al.} \cite{ye2020rotation}} \\ 
&& G1 &G2 & G3 & DORL &\\ \midrule

\multirow{4}{*}{A2C} & G1 &  $--$\  & \ $0/1/2$ $(0\backslash0)$ \  & \ $0/1/2$ $(0\backslash0)$ \  & \ $0/3/0$ $(0\backslash0)$ \  & \ $0/1/2$ $(0\backslash1)$ \  \\ 
 & G2 & \ $2/1/0$ $(0\backslash0)$ \  &  $--$\  & \ $0/3/0$ $(0\backslash0)$ \  & \ $2/1/0$ $(0\backslash0)$ \  & \ $2/1/0$ $(0\backslash1)$ \  \\ 
 & G3 & \ $2/1/0$ $(0\backslash0)$ \  & \ $0/3/0$ $(0\backslash0)$ \  &  $--$\  & \ $2/1/0$ $(0\backslash0)$ \  & \ $2/1/0$ $(0\backslash1)$ \  \\ 
 & DORL & \ $0/3/0$ $(0\backslash0)$ \  & \ $0/1/2$ $(0\backslash0)$ \  & \ $0/1/2$ $(0\backslash0)$ \  &  $--$\  & \ $0/1/2$ $(0\backslash1)$ \  \\ \midrule
\multicolumn{2}{c|}{\multirow{1}{*}{Ye \emph{et al.} \cite{ye2020rotation}}} & \ $2/1/0$ $(1\backslash0)$ \  & \ $0/1/2$ $(1\backslash0)$ \  & \ $0/1/2$ $(1\backslash0)$ \  & \ $2/1/0$ $(1\backslash0)$ \  &  $--$\  \\

\bottomrule
 
    \end{tabular}
    \label{tab:ranksum_waterpuzzle_all}
\end{table*}         

\end{document}